\documentclass[preprint,12pt]{elsarticle}

\usepackage{microtype}
\usepackage{amsmath,amssymb,amsthm}
\usepackage{mathtools}
\usepackage{booktabs}
\usepackage{fancyhdr}
\usepackage{graphicx}
\usepackage{subcaption}
\usepackage{tikz-cd}
\usepackage{multicol}
\usepackage{xcolor}
\usepackage{url}
\usepackage{hyperref}

\hypersetup{
  colorlinks=true,
  linkcolor=blue!50!black,
  citecolor=blue!50!black,
  urlcolor=blue!60!black
}

\journal{Journal Of Computational Physics}

\begin{document}

\begin{frontmatter}



\title{Iterative Refinement Diffusion for Super-Resolved Data Assimilation of Multiscale Physical Systems}

\author[utk]{Mrigank Dhingra\corref{cor1}}
\ead{mdhingra@vols.utk.edu}

\author[uchicago]{Ramchandran Muthukumar}
\ead{ramchandran@uchicago.edu}

\author[uchicago]{Rebecca Willett}
\ead{willett@uchicago.edu}

\author[utk]{Omer San}
\ead{osan@utk.edu}

\cortext[cor1]{Corresponding author}

\affiliation[utk]{
    organization={Department of Mechanical and Aerospace  Engineering, University of Tennessee},
    city={Knoxville},
    state={TN},
    postcode={37996},
    country={USA}
}

\affiliation[uchicago]{
    organization={UChicago Data Science Institute, University of Chicago},
    city={Chicago},
    state={IL},
    postcode={60637},
    country={USA}
}

\begin{abstract}
Recovering dynamically coherent high-resolution states from sparse, low-resolution observations is a central challenge in scientific machine learning and data assimilation. Classical data assimilation methods exploit temporal information through forecast--update cycles, but typically require repeated access to computationally expensive high-resolution forecast models. In contrast, modern generative super-resolution methods can reconstruct unresolved fine-scale structure from coarse observations, yet they often operate as one-shot mappings that do not exploit the dynamical constraints carried by past states. We introduce an iterative refinement framework for learned data assimilation that combines these two perspectives. Rather than performing a single coarse-to-fine state reconstruction, we propose decomposing the task into a sequence of resolution-wise forecast--analysis operations across a multiresolution  hierarchy. At each stage, a shared neural operator with resolution-dependent spectral mode slicing provides a dynamical prior, and a shared conditional diffusion corrector uses the current coarser-resolution state to produce a refined posterior at the next finer resolution.

We evaluate our proposed method, termed \textit{Iterative Refinement} (IR), on two canonical multiscale benchmarks: the one-dimensional stochastically forced Burgers equation and the two-dimensional Kraichnan turbulence. On the demanding Kraichnan Turbulence benchmark at \(256\times256\), IR achieves an RMSE of $0.184$ and an SSIM of $0.836$, outperforming baselines such as spectral interpolation, one-shot diffusion super-resolution, enhanced deep super-resolution (EDSR) and an autoregressive forecaster. 
On the Burgers testbed where the coarse-to-fine inverse problem is more constrained, IR achieves an RMSE of \(0.00436\) and remains competitive with the best method, the one-shot diffusion super-resolution, which achieves the lowest RMSE of \(0.00278\). These results indicate that although one-shot generative reconstruction approaches are effective for sufficiently constrained data assimilation, a hierarchical forecast--analysis refinement becomes advantageous for strongly multiscale and underdetermined regimes. Overall, the proposed framework provides a learned data assimilation strategy for combining temporal priors, generative spatial models, and multiresolution reconstruction in complex physical systems.
\end{abstract}


\begin{highlights}
\item Introduces a multiresolution forecast--analysis diffusion framework for super-resolved data assimilation.
\item Uses a shared FNO forecaster with spectral mode slicing and a shared conditional diffusion corrector.
\item Demonstrates improved recovery of unresolved high-wavenumber energy in 2D Kraichnan turbulence.
\item Outperforms one-shot super-resolution and learned EnKF baselines while avoiding online full-solver forecasts.
\end{highlights}

\begin{keyword}
 data assimilation \sep diffusion models \sep super-resolution \sep neural operators\sep multiscale turbulence \sep ensemble Kalman filtering


\end{keyword}

\end{frontmatter}



\section{Introduction}
\subsection{Background and Motivation}


Recovering high-resolution states of complex physical systems from sparse, noisy, or low-resolution observations is a central objective in data assimilation, inverse problems, and scientific machine learning~\cite{kalnay2003atmospheric,asch2016data,stuart2010inverse,brunton2019datadriven}. In applications such as numerical weather prediction~\cite{kalnay2003atmospheric,evensen1994sequential}, geophysical fluid dynamics~\cite{Vallis_2017}, and turbulence modeling~\cite{pope2000turbulent,frisch1995turbulence}, the underlying dynamics evolve across a wide range of interacting spatial and temporal scales. Classical data assimilation methods, such as Kalman filtering and ensemble Kalman filtering~\cite{kalman1960new,evensen1994sequential}, address this estimation problem sequentially: past states are propagated forward by a forecast model and then corrected against current observations in an analysis step~\cite{Sanz-Alonso_Stuart_Taeb_2023}. This forecast--update cycle is attractive because it explicitly exploits temporal structure and maintains dynamical coherence across successive states. Physics-based forecast models derived from the governing equations of a physical system, while accurate, typically incur a prohibitive computational cost when used repeatedly at the target resolution. As a consequence, they may be infeasible in high-resolution operational or ensemble-based assimilation settings. This exposes a tension between reconstruction fidelity, which requires resolving fine-scale structure, and computational tractability, which constrains how many high-resolution forecasts can be propagated online.

A rapidly growing body of work alleviates the cost of physics-based forecast models with data-driven surrogates learned from offline state trajectories \cite{Chen_2023, lam2023graphcast, pathak2022fourcastnet}. Once trained, the surrogate forecast models can be used within a data assimilation pipeline to advance the state at a fraction of the cost of their physics-based counterparts. The promise of data-driven assimilation pipelines has motivated further improvements based on the underlying filtering mechanisms \cite{rozet2023score, huang2024diffda, hodyss2025using}. Independently, generative super-resolution has emerged as a powerful tool for reconstructing fine-scale content directly from coarse observations ~\cite{ho2020ddpm,saharia2023image,lopezgomez2025dynamical}, with early success in fluid dynamics~\cite{fukami2019superresolution}, weather forecasting~\cite{lopezgomez2025dynamical}, and more. Similar to data-driven surrogate forecast models, generative super-resolution methods amortize the cost of reconstruction using an offline training stage based on a curated, paired dataset of high-resolution states and low-resolution observations. However, generative super-resolution methods are often formulated as one-shot coarse-to-fine mappings that treat each low-resolution observation as an independent input. As a result, they fail to take advantage of the strong dynamical constraints carried by previous states, especially in systems where fine-scale structures evolve coherently over time. These two lines of work provide complementary halves of a learned assimilation pipeline. A learned surrogate forecaster provides the high-resolution dynamical prior but leaves assimilation of low-resolution current observations to the classical Kalman-gain based update. On the other hand, a generative super-resolution model reconstructs high-resolution states directly from low-resolution current observations but ignores the dynamical information encoded in past state estimates. 

In this article, we propose a single, fully learned forecast-correct pipeline in which a learned forecast provides the dynamical prior for state reconstruction, while a generative model enhances the classical analysis step based on current observations. Further, we introduce an \textit{iterative-refinement} (IR) strategy that decomposes the state reconstruction task into a sequence of forecast--analysis steps across a multiscale hierarchy, each operating over a modest refinement ratio. Such a formulation, converts a single severely ill-conditioned coarse-to-fine inverse problem into a sequence of better-conditioned local refinements. IR is motivated by the guiding principle of classical multigrid methods \cite{TrotMult2001} that solve multiscale problems not on a single fine grid but by decomposing them across a hierarchy of resolutions. 






\subsection{Challenges in Super-Resolved Data Assimilation}

Despite its appeal, super-resolved data assimilation is a fundamentally difficult problem. The first challenge is that the inverse mapping from low-resolution observations to high-resolution states is severely ill-conditioned. Coarse measurements discard a large fraction of the physically relevant degrees of freedom, as a consequence, many distinct fine-resolution states may be compatible with a particular coarse observation~\cite{stuart2010inverse}. This ambiguity is especially prevalent in turbulent or shock-dominated systems, where small-scale features carry substantial energy and can strongly influence subsequent evolution~\cite{frisch1995turbulence,kraichnan1967inertial}. A single coarse-to-fine step may result in oversmoothing and under-recovery of high-wavenumber energy, a limitation well-documented in super-resolution at large upscaling factors \cite{ledig2017photo, fukami2019superresolution}.
Although fine-scale structures are not uniquely determined by coarse measurements, they are still constrained by the interaction of transport, dissipation, forcing, and nonlinear couplings across scales ~\cite{kraichnan1967inertial,fukami2019superresolution}. 
This motivates regularizing the reconstruction task with additional structure, and where possible, decomposition into sub problems that are individually better conditioned.  

A second challenge is temporal consistency. A reconstruction method must do more than generate plausible fine-scale snapshots at isolated time steps; it must produce a sequence of states whose evolution remains dynamically coherent. One-shot generative super-resolution models may generate visually convincing details, yet without explicit temporal conditioning they may introduce frame-to-frame jitter, spurious small-scale artifacts, and physically implausible transitions. Similar issues are well recognized in video and spatio-temporal super-resolution, where recurrent or temporally coupled designs are often introduced specifically to improve consistency across frames~\cite{sajjadi2018framerecurrent,fukami2021spatiotemporal}. In sequential scientific settings, these errors can accumulate rapidly and destabilize downstream forecasting or filtering. 

A third challenge concerns the deployment of learned super-resolution correctors inside a sequential assimilation loop. A corrector trained only on clean inputs or idealized priors may not see the same conditioning distribution at inference time. In closed-loop inference, the forecast prior is generated autoregressively from previous corrected states, and each refinement level receives conditioning information produced by earlier stages of the cascade. Thus, errors can accumulate across both time and resolution levels. This train--test discrepancy is closely related to exposure bias in sequence models, where errors can accumulate when a model is rolled out on its own predictions rather than ground-truth inputs~\cite{bengio2015scheduled,lamb2016professor}. We address this issue by training the diffusion corrector on teacher-forced outputs of the learned FNO forecaster, so that the corrector is exposed to the characteristic error distribution of the forecast model before deployment.


These considerations suggest that super-resolved data assimilation should not be treated as either a pure forecasting problem or a pure image-style super-resolution problem. Instead, it requires a framework that combines dynamical priors, observational correction, multiscale structure, and probabilistic reconstruction within a single sequential pipeline. This motivates the iterative refinement strategy developed in the present work, in which the reconstruction is performed progressively across a hierarchy of resolutions through repeated forecast--analysis operations.

\subsection{Unresolved scales recovery task}

\begin{figure}[htbp]
  \centering
  \includegraphics[width=\textwidth]{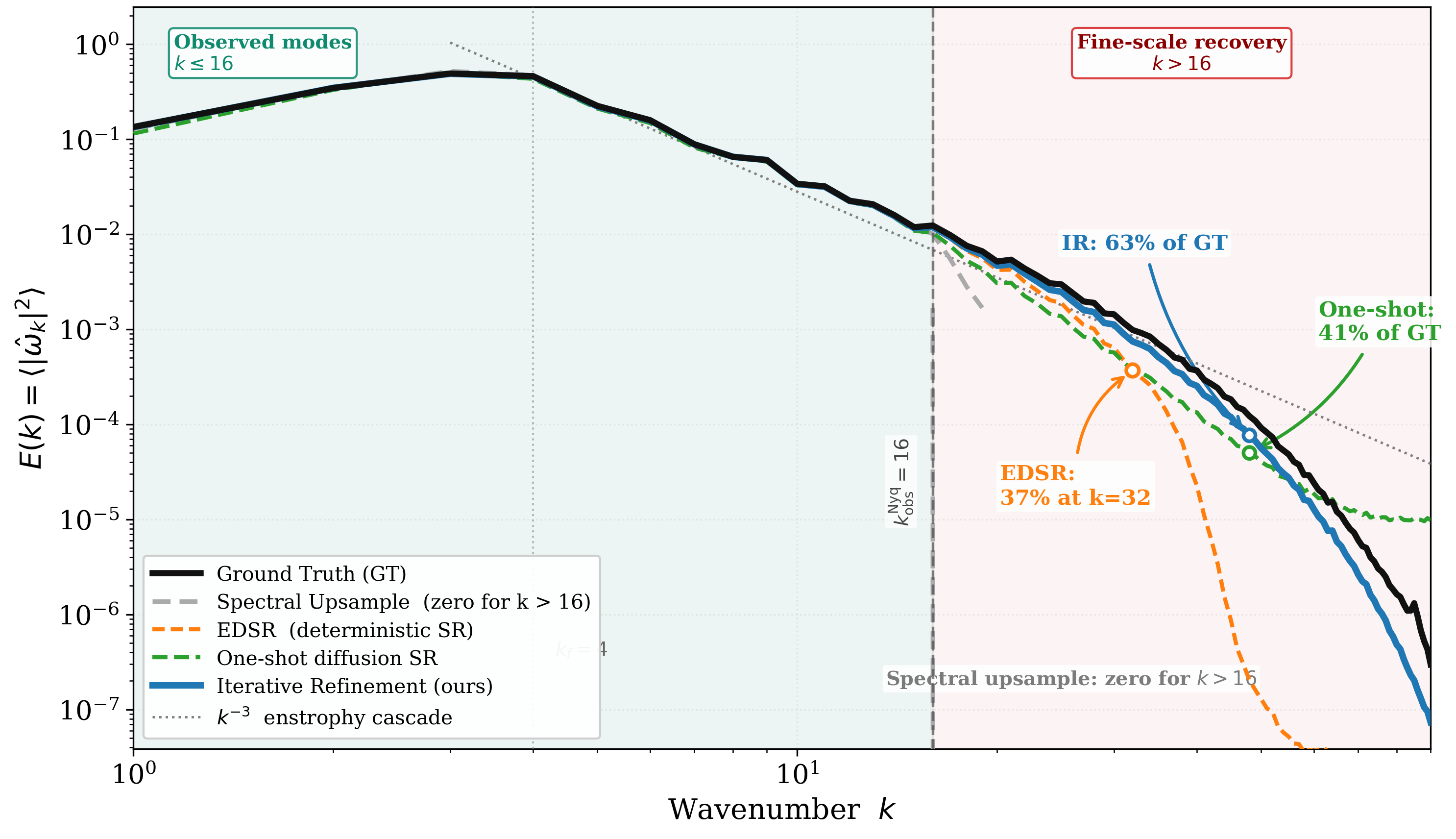}
  \caption{Preview of the main spectral recovery result on the 2D Kraichnan turbulence benchmark. The shaded blue region denotes the observed low-wavenumber modes resolved by the coarse \(32\times32\) input, while the shaded red region denotes unresolved modes with \(k>16\) that must be reconstructed. Spectral upsampling loses energy beyond the coarse Nyquist limit, EDSR is overly dissipative, and one-shot diffusion recovers only part of the high-wavenumber tail. Iterative refinement most closely follows the ground-truth spectrum in the unresolved range, demonstrating improved recovery of fine-scale turbulent structure.}
  \label{fig:spectrum_schematic_intro}
\end{figure}

A central question in super-resolved data assimilation is whether a method recovers the physically relevant unresolved scales or merely reconstructs visually plausible fields that under-represent high-wavenumber energy. Figure~\ref{fig:spectrum_schematic_intro} previews our central empirical finding on the 2D Kraichnan turbulence benchmark. The coarse \(32\times32\) observation resolves only low-wavenumber content up to its Nyquist cutoff, \(k=16\); all modes beyond this cutoff must be inferred by the reconstruction method. In this unresolved range, spectral upsampling contains no meaningful energy, deterministic enhanced deep super-resolution (EDSR) network~\cite{lim2017edsr} decays too rapidly, and one-shot diffusion recovers only part of the high-wavenumber tail. By contrast, iterative refinement follows the ground-truth spectrum most closely, indicating substantially better recovery of fine-scale turbulent structure.

\begin{figure}[htbp]
  \centering
  \includegraphics[width=0.86\textwidth]{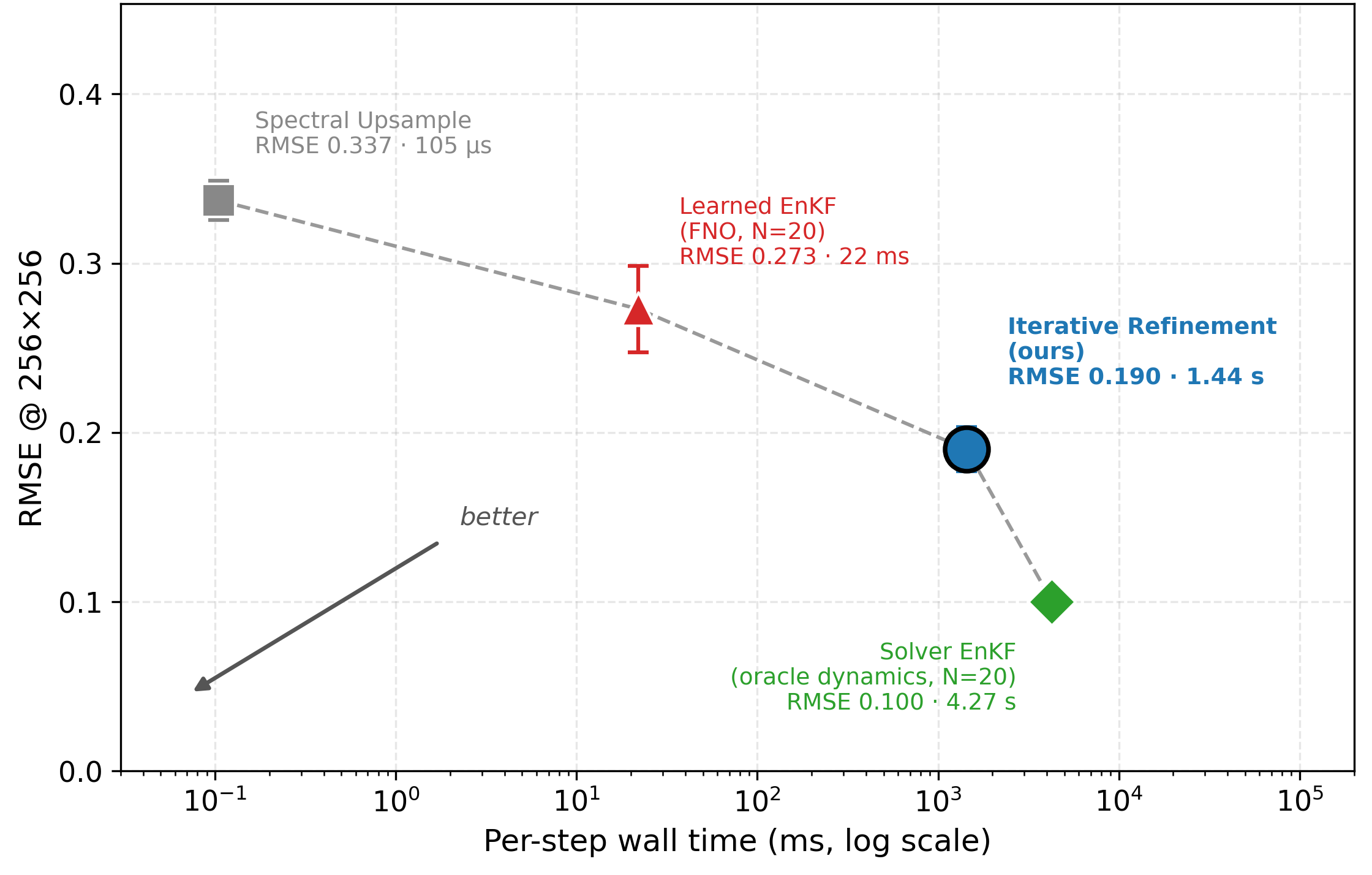}
  \caption{Preview of the accuracy--cost trade-off for super-resolved data assimilation on the 2D Kraichnan benchmark. Each point reports the overall RMSE at \(256\times256\) against the per-step wall time on a logarithmic scale. Lower and further left is better. The solver EnKF, labeled as ``Solver EnKF (oracle dynamics knowledge),'' achieves the lowest RMSE because it advances an ensemble using the high-resolution pseudospectral solver. The learned EnKF is much faster but less accurate. Iterative refinement provides an intermediate learned assimilation strategy, substantially improving over the learned EnKF while avoiding repeated online full-solver ensemble forecasts.}
  \label{fig:da_cost_accuracy_intro}
\end{figure}

Because the proposed method is also intended as a learned data-assimilation strategy, we additionally preview its accuracy--cost trade-off relative to ensemble Kalman filtering baselines in Fig.~\ref{fig:da_cost_accuracy_intro}. A solver-based EnKF with oracle access to the high-resolution Kraichnan dynamics achieves the lowest RMSE, but requires repeated full-solver ensemble forecasts. A learned EnKF based on an FNO forecast is much cheaper, but substantially less accurate. Iterative refinement occupies an intermediate regime: it avoids online solver integration while achieving much lower error than the learned EnKF. This comparison motivates the use of a learned forecast--analysis cascade rather than either a purely linear learned filtering update or an expensive solver-based ensemble forecast.

Together, these two previews motivate the hierarchical forecast--analysis formulation developed in the following sections. Rather than attempting to infer all missing scales in a single step or relying on repeated high-resolution ensemble forecasts, the proposed method reconstructs the high-resolution state through a sequence of resolution-wise learned analyses, allowing fine-scale information to be progressively recovered across the cascade.

\subsection{Related Work}

Our work lies at the intersection of data assimilation, generative super-resolution, and neural operator learning. We therefore review related work along four directions: diffusion-based data assimilation, generative downscaling and super-resolution, neural operators for dynamical forecasting, and diffusion model foundations.

\paragraph{Diffusion-based data assimilation}
Recent work has begun to explore diffusion and score-based generative models as alternatives to classical Gaussian or ensemble-based data assimilation updates. Score-based data assimilation (SDA) learns a generative prior over state trajectories and performs inference by guiding samples toward observations, providing a probabilistic route to trajectory-level assimilation~\cite{rozet2023score}. DiffDA scales this idea to weather-scale data assimilation by adapting a pretrained GraphCast-style weather model as the denoising backbone and conditioning on forecast information and sparse observations~\cite{huang2024diffda}. Hodyss and Morzfeld further clarify the probabilistic foundations of diffusion-based data assimilation, showing that different diffusion-DA formulations may correspond to different posterior distributions depending on the prior and likelihood assumptions~\cite{hodyss2025using}. Recent work has investigated diffusion-based nonlinear ensemble filtering~\cite{bao2025nonlinear}, state-observation augmented diffusion models for nonlinear assimilation with unknown dynamics~\cite{li2025soad}, and large-scale latent diffusion models for global atmospheric data assimilation~\cite{andry2025appa} and autoregressive diffusion-control approaches that reduce forecast drift under sparse observations~\cite{srivastava2025cada}.
Our method differs in emphasis: rather than learning a global trajectory prior or a single end-to-end assimilation operator, we formulate super-resolved data assimilation as a multiresolution forecast--analysis cascade. 

\paragraph{Generative downscaling and super-resolution}
A related line of work uses generative models for downscaling and super-resolution in images, fluids, weather, and climate. In computer vision, SR3 introduced diffusion-based image super-resolution through iterative denoising conditioned on low-resolution inputs~\cite{saharia2023image}, while cascaded diffusion models extended this idea to high-fidelity image generation by composing a low-resolution generative model with one or more diffusion-based super-resolution stages~\cite{ho2021cascadeddiffusionmodelshigh}. Deterministic architectures such as the enhanced deep super-resolution network (EDSR) remain strong baselines for single-image super-resolution~\cite{lim2017edsr}. In fluid dynamics, machine-learning-based super-resolution has been used to reconstruct turbulent flow fields from severely under-resolved data~\cite{fukami2019superresolution}. In weather and climate, recent diffusion-based methods have shown strong promise for stochastic downscaling. CorrDiff uses a residual corrective diffusion model to downscale coarse weather states to kilometer-scale fields while recovering realistic spectra and distributions~\cite{mardani2025corrdiff}. \citet{lopezgomez2025dynamical} combine dynamical downscaling with generative refinement to reduce the cost of producing large downscaled climate ensembles. These approaches show that generative models are powerful tools for recovering unresolved fine-scale structure. However, many super-resolution and downscaling methods are formulated as one-shot coarse-to-fine mappings or purely spatial generative cascades, rather than sequential forecast--analysis procedures that repeatedly combine a dynamical prior with current coarse observations. 

\paragraph{Neural operators and learned dynamical priors}
The forecast component of our method is related to neural operator learning, which seeks to learn mappings between function spaces rather than finite-dimensional vectors. DeepONet introduced an operator-learning architecture based on branch and trunk networks~\cite{lu2021deeponet}, while the broader neural-operator framework formalized discretization-invariant operator learning for PDE solution maps~\cite{kovachki2023neural}. The Fourier Neural Operator (FNO) parameterizes integral operators in Fourier space and has become a widely used architecture for learning PDE dynamics and surrogate models~\cite{li2021fourier}. FNO-style models have also influenced large-scale learned weather forecasting systems, including FourCastNet~\cite{pathak2022fourcastnet}, while GraphCast demonstrates the broader potential of learned global weather forecasting from reanalysis data~\cite{lam2023graphcast}. 

\paragraph{Diffusion model foundations}
The correction stage builds on the denoising diffusion framework. Denoising Diffusion Probabilistic Models (DDPMs) define a forward noising process and learn the reverse denoising dynamics through noise prediction~\cite{ho2020ddpm}. Score-based generative modeling provides a continuous-time view of diffusion processes through stochastic differential equations~\cite{song2021scorebased}. Denoising Diffusion Implicit Models (DDIMs) show that the same trained model can be sampled using faster non-Markovian reverse processes~\cite{song2021ddim}, and improved DDPMs demonstrate practical modifications for better likelihoods and more efficient sampling~\cite{nichol2021improved}. These developments provide the algorithmic foundation for our conditional diffusion corrector and DDIM-based inference procedure. Relative to prior work, the key distinction of our approach is the combination of (i) a learned dynamical prior at each refinement level, (ii) a shared diffusion-based correction mechanism, and (iii) a hierarchical resolution cascade that decomposes coarse-to-fine reconstruction into sequential local refinement steps. In this sense, our method is closer to a learned multiscale filtering procedure than to either one-shot super-resolution or conventional autoregressive neural forecasting.


\section{Problem Formulation}

\begin{figure}[htbp]
    \centering
    \includegraphics[width=0.95\textwidth]{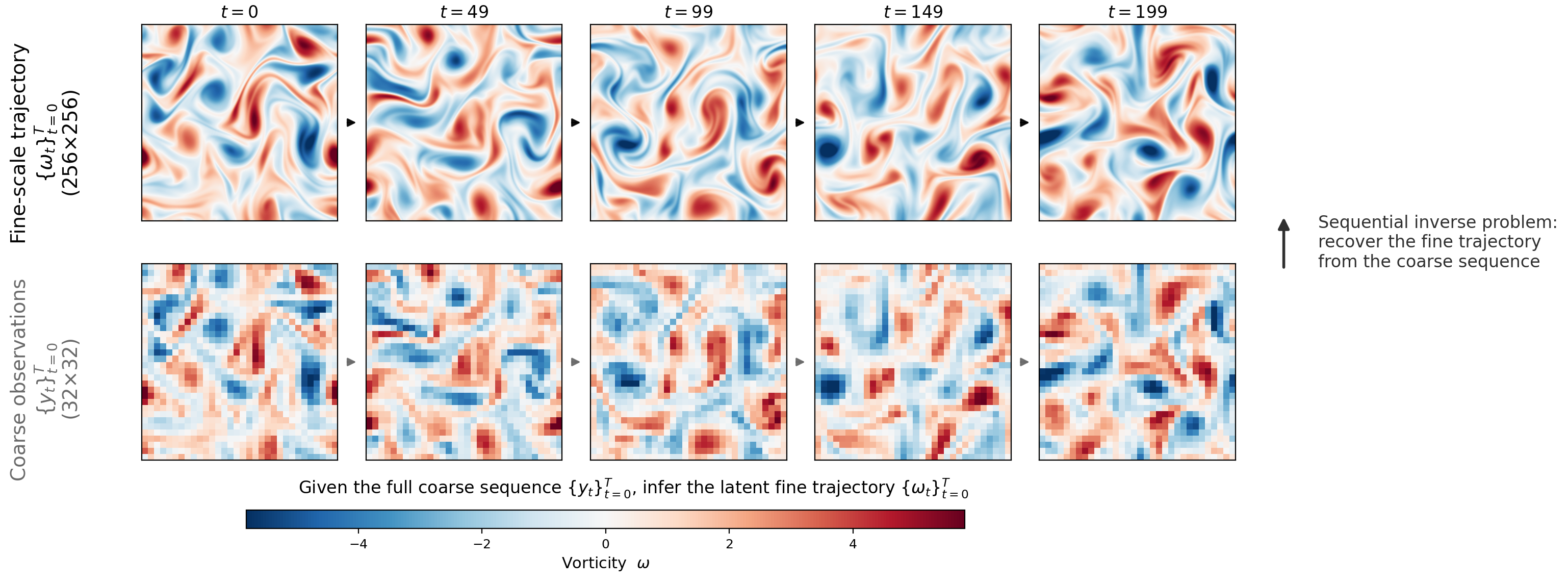}
    \caption{Illustration of the sequential super-resolution inverse problem. The finest-resolution trajectory contains coherent small-scale structures that are only partially visible after coarse observation. Given only the low-resolution sequence, many high-resolution trajectories may be compatible with the same observations, making the reconstruction problem ill-posed. Temporal information from previous high-resolution estimates is therefore essential for narrowing the space of admissible fine-scale states.}
    \label{fig:inverse_problem}
\end{figure}

Consider studying turbulent flow via low-resolution observations as depicted in Figure \ref{fig:inverse_problem}. At each instant we only see a coarse, blurred snapshot, while the underlying flow carries sharp fine-scale structure, including shocks, filaments and vortices. Reconstruction of the high-resolution flow from a single low-resolution observation is ill-conditioned, however fine-scale structure at higher-resolution arises from the continuous evolution of fluid flow along a trajectory. Thus, the reconstruction task is a sequential estimation problem. The remainder of this section formalizes the learning task. 

\subsection{Sequential Super-Resolution as Data Assimilation}

Let $x_t \in \mathbb{R}^{n_R}$ denote the latent state of a physical system at time step $t$, represented on the finest spatial grid with $n_R$ grid points. We assume that the state evolves according to an unknown dynamical process $x_t = \mathcal{M}(x_{t-1}) + \xi_t$, where $\mathcal{M}$ denotes the underlying evolution operator and $\xi_t$ represents model error, forcing, or unresolved stochastic effects. In many scientific applications, direct access to $x_t$ is unavailable, and instead we observe only a coarse, noisy version of the state, $y_t = \mathcal{H}(x_t) + \varepsilon_t$ where $\mathcal{H}: \mathbb{R}^{n_R} \to \mathbb{R}^{n_0}$ is an observation operator mapping the fine state to the coarsest observable resolution with $n_0 \ll n_R$, and $\varepsilon_t \in \mathbb{R}^{n_0}$ denotes the observational noise. Given a low-resolution observation trajectory $\{y_t\}_{t=0}^T$, the goal is to recover the high-resolution states $\{{x}_t\}_{t=0}^T$. This sequential state-estimation viewpoint is standard in filtering and data assimilation~\cite{kalman1960new,jazwinski1970stochastic,asch2016data}.

This problem may be viewed as a form of sequential super-resolution. Unlike standard single-image super-resolution, the target is not to reconstruct each high-resolution state independently from the current coarse observation alone, but rather to infer a temporally coherent sequence of fine states consistent with both the observations and the underlying dynamics. In this sense, the task lies at the intersection of super-resolution and data assimilation: the observations constrain the resolved large scales, while the temporal history provides a dynamical prior that restricts the admissible fine-scale reconstructions~\cite{tian2011survey,asch2016data}.

Formulated this way, super-resolved reconstruction becomes a filtering problem over a hierarchy of partially observed states. The key challenge is that the observation operator removes substantial fine-scale information, rendering the inverse mapping from $y_t$ to $x_t$ non-unique (Fig.~\ref{fig:inverse_problem}). This non-uniqueness is characteristic of ill-posed inverse problems, where additional prior, dynamical, or statistical structure is required to regularize the solution~\cite{stuart2010inverse, Sanz-Alonso_Stuart_Taeb_2023}. The temporal dependence across successive states is therefore essential for narrowing the space of plausible reconstructions.

\subsection{Notation and Resolution Hierarchy}

\begin{figure}[htbp]
    \centering
    \includegraphics[width=0.8\textwidth]{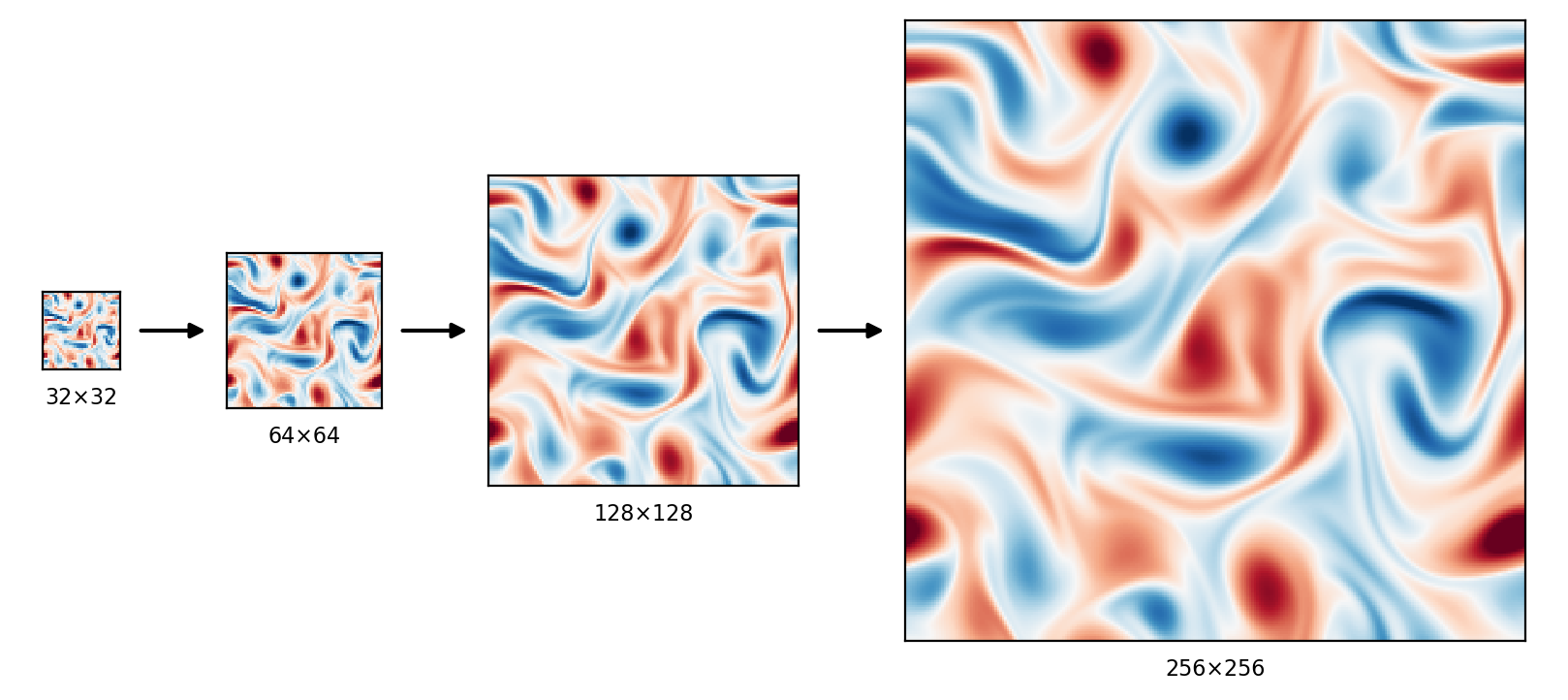}
    \caption{Multiresolution state hierarchy used in the proposed formulation. The coarsest level $r=0$ corresponds to the observed state, while the finest level $r=R$ is the target reconstruction. Downsampling operators $\mathcal{D}_{r+1\to r}$ define the coarse representations, and upsampling operators $\mathcal{U}_{r\to r+1}$ transfer information upward through the hierarchy. This decomposition converts a single large coarse-to-fine inverse problem into a sequence of smaller refinement tasks.}
    \label{fig:res_pyr}
\end{figure}

To exploit the multiscale structure of the problem, we represent the state on a hierarchy of grids (see Fig.~\ref{fig:res_pyr}) indexed by resolution levels $r \in \{0,1,\dots,R\}$ where $r=0$ denotes the coarsest level and $r=R$ denotes the target finest level with resolution dimensions $n_0 < n_1 < \cdots < n_R$. Let $x_t^{(r)} \in \mathbb{R}^{n_r}$ denote the state at time $t$ represented on resolution level $r$, in particular, $x_t^{(R)} = x_t$ is the finest-resolution state of interest.

We assume that the multiresolution states are linked by fixed resolution- transfer operators. Let $\mathcal{D}_{r+1 \to r}: \mathbb{R}^{n_{r+1}} \to \mathbb{R}^{n_r}$ denote a downsampling operator from level $r+1$ to level $r$, and let $\mathcal{U}_{r \to r+1}: \mathbb{R}^{n_r} \to \mathbb{R}^{n_{r+1}}$ denote a corresponding upsampling operator. Such hierarchical representations are closely related to classical multiresolution and pyramid constructions~\cite{burt1983laplacian,mallat1989multiresolution}. In practice, these operators may be implemented using spectral restriction and zero-padding, but for the present formulation we treat them abstractly.
Given a finest-resolution trajectory $\{x_t^{(R)}\}_{t=0}^T$, the associated lower-resolution states may be defined recursively by
\begin{equation}
x_t^{(r)} = \mathcal{D}_{R \to r}(x_t^{(R)}), \qquad r=0,\dots,R-1,
\end{equation}
where $\mathcal{D}_{R \to r}$ denotes the appropriate composition of downsampling maps from level $R$ to level $r$. Under this notation, the observation sequence may be identified with the coarsest state sequence, possibly corrupted by noise:
\begin{equation}
y_t = x_t^{(0)} + \varepsilon_t.
\end{equation}

The introduction of the resolution hierarchy allows the original coarse-to-fine reconstruction problem to be decomposed into a sequence of local refinement problems. Rather than directly recovering $x_t^{(R)}$ from $y_t$, one may instead infer $x_t^{(1)}, x_t^{(2)}, \dots, x_t^{(R)}$ in turn such that each stage increases resolution by a modest factor. This multiresolution viewpoint is central to the iterative refinement approach developed later.

\subsection{Observation Model and Reconstruction Objective}

With the multiresolution hierarchy in place, we now state the target objective of reconstruction. Given the observation sequence $\{y_t\}_{t=0}^T$, we seek an estimate of the finest-resolution trajectory
$\hat{x}_{0:T}^{(R)} = \{\hat{x}_t^{(R)}\}_{t=0}^T$ such that it is both observationally consistent and dynamically coherent. At a minimum, the reconstructed states should match the available coarse observations under the observation operator,
\begin{equation}
\mathcal{H}(\hat{x}_t^{(R)}) \approx y_t,
\end{equation}
while also evolving smoothly and plausibly in time. Cast as sequential Bayesian estimation, the ideal target is the filtering distribution of the finest state given all observations up to the current time,
\begin{equation}
\hat{x}_t^{(R)} \sim p\!\left(x_t^{(R)} \middle| y_{0:t}\right),
\label{eq:standard-filtering}
\end{equation}
where $y_{0:t} = \{y_0,\dots,y_t\}$
~\cite{jazwinski1970stochastic,asch2016data}. Directly characterizing~\eqref{eq:standard-filtering} is difficult as it couples the full range of scales in a single distribution. The multiresolution hierarchy enables us to factor the task across scales instead, replacing the single filtering distribution with a chain of local condition models that estimate the state at intermediate resolutions, 
\begin{equation}
\hat{x}_t^{(r+1)} \sim p\!\left(x_t^{(r+1)} \mid x_{t-1}^{(r+1)}, x_t^{(r)}\right),
\qquad r=0,\dots,R-1.
\label{eq:multi-res-conditioning}
\end{equation}
This formulation makes explicit that the current finer-scale state $\hat{x}_t^{(r+1)}$ should depend on both the previous state $x_{t-1}^{(r+1)}$ at the same resolution and the current state at $x_t^{(r)}$ available at the next coarser level. 

The central premise of this work is that this multilevel conditional structure is more tractable than a direct one-shot mapping from $y_t$ to $x_t^{(R)}$. A single coarse observation typically admits many plausible fine-scale reconstructions, especially in multiscale systems with shocks, filaments, or turbulent eddies. By contrast, the conditional distribution of $x_t^{(r+1)}$ given a coarse current state $x_t^{(r)}$ and a previous finer state $x_{t-1}^{(r+1)}$ is substantially more concentrated. This motivates learning the reconstruction process as a sequence of refinement steps across both time and scale. In the next section, we introduce a practical realization of this idea in which each refinement stage is implemented via a learned forecast-analysis mechanism that combines a dynamical prior at the target resolution with a conditional generative correction informed by the current coarser-resolution state.

\section{The Iterative Refinement Framework}

\begin{figure}[htbp]
    \centering
    \includegraphics[width=1.0\textwidth]{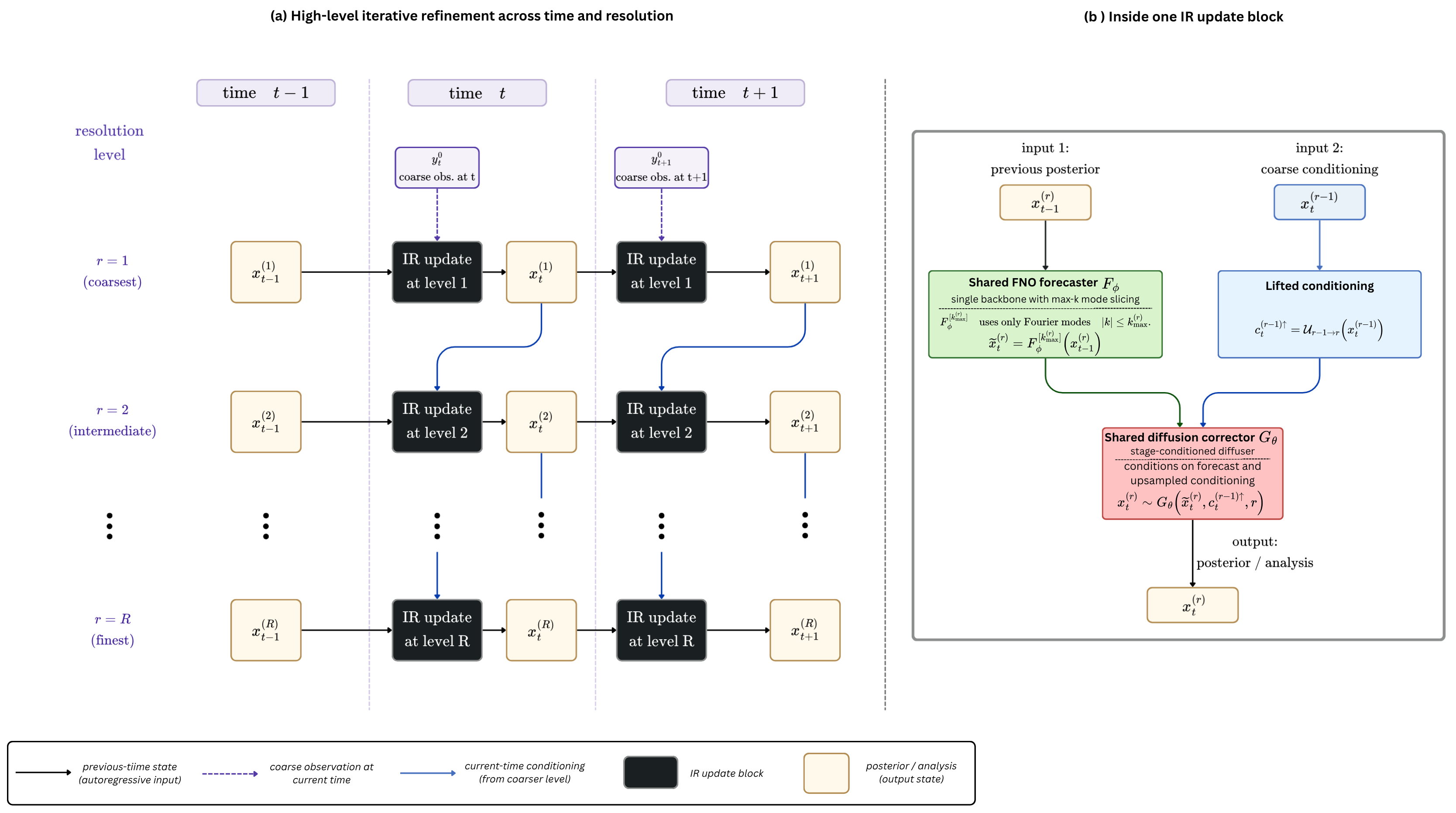}
    \caption{High-level view of iterative refinement across time and resolution. 
At each physical time step, an IR update block receives the previous posterior 
at the same resolution and the current conditioning signal from the coarser 
level, then produces the current posterior analysis at that resolution. 
Horizontal arrows denote temporal autoregressive propagation, while vertical 
arrows denote coarse-to-fine conditioning within the same time step. The right 
panel expands a single IR update block into its three components: a shared 
mode-sliced FNO forecaster, a conditioned upsampler, and a shared diffusion 
corrector.}
\label{fig:ir_time_resolution_overview}
    
\end{figure}

The iterative refinement (IR) framework reconstructs high-resolution states through a sequence of resolution-wise forecast--analysis operations across both time and scale, as summarized in Fig.~\ref{fig:ir_time_resolution_overview}. Rather than learning a single mapping from coarse observations to fine states, the method applies a learned IR update at each resolution level and physical time step. 

Let $\{x_t^{(r)}\}_{r=0}^R$ denote the multiresolution representation of the system state at time $t$, where $r=0$ corresponds to the coarsest observable level and $r=R$ to the target finest resolution. At each time step $t$, the reconstruction proceeds sequentially across resolution levels from coarse to fine:
\begin{equation}
x_t^{(0)} \rightarrow x_t^{(1)} \rightarrow \cdots \rightarrow x_t^{(R)}.
\end{equation}
Each refinement stage $r \to r+1$ combines two sources of information:
(i) a dynamical prior obtained by advancing the state at previous time step at resolution $r+1$, and (ii) a coarse conditioning signal through the state at current time step at the coarser resolution $r$.

This structure mirrors the classical data assimilation paradigm, in which a forecast is corrected using new observations. However, instead of relying on a known forecast model and a linear update rule, both components are learned from data. Specifically, we employ a shared FNO forecaster evaluated with a resolution-dependent spectral mode slice to propagate the state forward in time, and a shared conditional generative model to perform the correction. The reconstruction at each time step is thus a cascade of local refinement operations, each resolving a modest scale gap. This hierarchical decomposition transforms a highly ill-posed coarse-to-fine reconstruction problem into a sequence of tractable conditional inference tasks.

\subsection{Resolution-Wise Forecast--Analysis Decomposition}
\label{sec:forecast_analysis_decomposition}

At each target resolution level \(r=1,\dots,R\), we decompose the update from time \(t-1\) to time \(t\) into two stages: a forecast step and an analysis step. The forecast step provides a dynamical prior at resolution \(r\), while the analysis step assimilates current information from the previous coarser level \(r-1\) to produce the posterior state.

\paragraph{Forecast step}
We first construct a dynamical prior at resolution \(r\) by advancing the previous posterior state using a shared Fourier Neural Operator forecaster \(F_{\phi}\)~\cite{li2021fourier}, whose spectral weights are evaluated through a resolution-dependent low-frequency slice. The forecast prior is written as
\begin{equation}
\widetilde{x}_t^{(r)}
=
F_{\phi}^{[k_{\max}^{(r)}]}
\!\left(
x_{t-1}^{(r)}
\right),
\qquad r=1,\dots,R,
\end{equation}
where \(F_{\phi}^{[k_{\max}^{(r)}]}\) denotes the shared FNO evaluated using only the Fourier modes active at resolution level \(r\). The quantity \(k_{\max}^{(r)}\) specifies the maximum retained spectral mode at that level. This mode-slicing strategy allows the same forecasting model to be shared across the hierarchy while respecting the bandwidth available at each resolution. The forecast step plays the role of the dynamical model in classical data assimilation, providing a prior estimate of the current state before observational analysis.

\paragraph{Analysis step}
The forecast prior is then refined using information from the previous coarser level at the current time step. Let \(c_t^{(r-1)}\) denote the conditioning signal available at level \(r-1\). For the first refinement stage, this conditioning signal is the observed coarse measurement. For all subsequent stages, it is the corrected posterior from the previous refinement level:
\begin{equation}
c_t^{(r-1)}
=
\begin{cases}
y_t^{(0)}, & r=1, \\
x_t^{(r-1)}, & r=2,\dots,R.
\end{cases}
\end{equation}
We lift this conditioning signal to the target resolution using the upsampling operator \(\mathcal{U}_{r-1\to r}\):
\begin{equation}
c_t^{(r-1)\uparrow}
=
\mathcal{U}_{r-1\to r}
\!\left(
c_t^{(r-1)}
\right),
\qquad r=1,\dots,R.
\end{equation}
The posterior at resolution \(r\) is then obtained from the shared conditional generative analysis model:
\begin{equation}
x_t^{(r)}
\sim
p_\theta\!\left(
x_t^{(r)}
\,\middle|\,
\widetilde{x}_t^{(r)}, \;
c_t^{(r-1)\uparrow}, \;
r
\right),
\qquad r=1,\dots,R.
\end{equation}
Here, \(p_\theta\) is the diffusion-based corrector shared across all refinement stages. The explicit resolution index \(r\) allows the corrector to adapt its denoising behavior to the current scale transition.

Two properties of the analysis step are worth emphasizing. First, the analysis depends jointly on the dynamical prior \(\widetilde{x}_t^{(r)}\) and the current coarser-level information \(c_t^{(r-1)\uparrow}\), analogous to the innovation-based update in classical filtering. Second, the conditional distribution spans only a modest refinement gap from level \(r-1\) to level \(r\), and is therefore typically more concentrated than the distribution of finest-level states \(x_t^{(R)}\) conditioned directly on the coarsest observation \(y_t^{(0)}\).

\paragraph{Shared models across scales}
Both components of the forecast--analysis update are shared across resolution levels. The forecaster shares a single set of FNO parameters \(F_{\phi}\), with resolution adaptation achieved by slicing the learned spectral weights according to \(k_{\max}^{(r)}\). The corrector shares a single diffusion model \(p_\theta\), with scale adaptation achieved by conditioning on the resolution level \(r\). Together, these two sharing mechanisms make the refinement pipeline resolution-aware without requiring independently trained forecast or correction networks at every scale.

\begin{figure}[htbp]
    \centering
    \includegraphics[width=\linewidth]{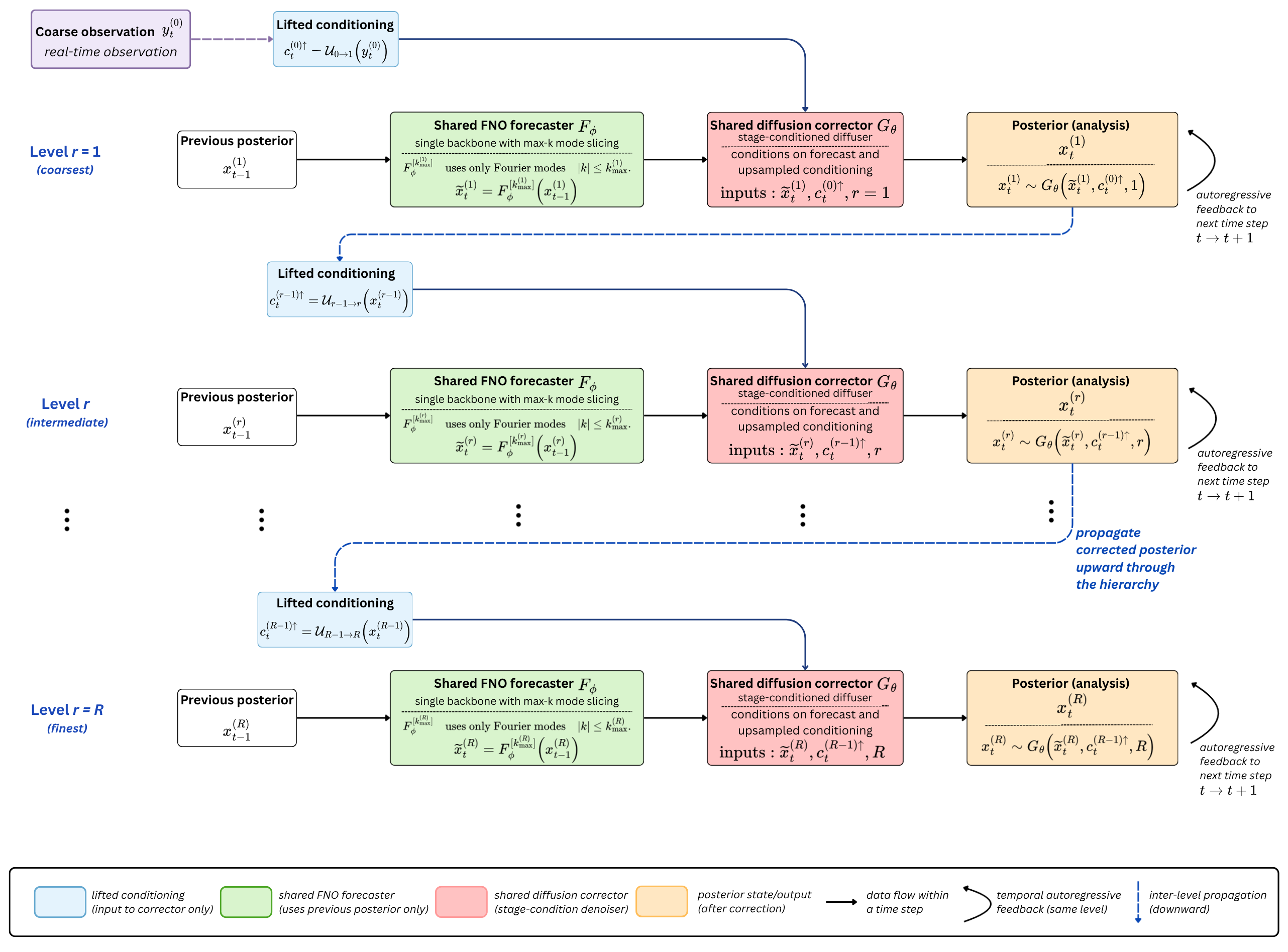}
   \caption{Detailed one-step forecast--analysis cascade for iterative refinement. 
At each target resolution level \(r\), the shared mode-sliced FNO forecaster 
advances the previous posterior \(x_{t-1}^{(r)}\) to produce the forecast prior 
\(\widetilde{x}_t^{(r)}\). The lifted conditioning signal 
\(c_t^{(r-1)\uparrow}\) bypasses the forecaster and enters only the shared 
diffusion corrector, which produces the posterior analysis \(x_t^{(r)}\). 
The posterior from each level is then propagated upward as conditioning for 
the next finer refinement stage.}
\label{fig:ir_one_step_cascade}
\end{figure}

Figure~\ref{fig:ir_one_step_cascade} illustrates this decomposition for one physical time step. The forecaster receives only the previous posterior at the same resolution, while the lifted coarse-level conditioning enters the analysis stage through the diffusion corrector.

\subsection{Iterative Multiscale Inference Procedure}
\label{sec:iterative_multiscale_inference}

The full reconstruction at time step \(t\) is obtained by applying the level-wise forecast--analysis update from Section~\ref{sec:forecast_analysis_decomposition} sequentially across the resolution hierarchy. Starting from the current coarse observation, inference proceeds from coarse to fine,
\begin{equation}
y_t^{(0)}
\longrightarrow
x_t^{(1)}
\longrightarrow
x_t^{(2)}
\longrightarrow
\cdots
\longrightarrow
x_t^{(R)} ,
\end{equation}
where \(x_t^{(R)}\) is taken as the final high-resolution analysis.

At the first refinement stage, the conditioning signal is the current coarse observation \(y_t^{(0)}\). After each analysis update, the resulting posterior becomes the conditioning signal for the next finer stage. Thus, within a single physical time step, corrected information is propagated upward through the hierarchy:
\begin{equation}
x_t^{(r)}
\longrightarrow
\mathcal{U}_{r\to r+1}
\!\left(
x_t^{(r)}
\right),
\qquad r=1,\dots,R-1 .
\end{equation}
Equivalently, the lifted posterior from level \(r\) becomes the conditioning input \(c_t^{(r)\uparrow}\) for the forecast--analysis update at target level \(r+1\). This posterior propagation distinguishes iterative refinement from a cascade in which every stage is repeatedly conditioned only on the original coarse observation.

At the initial time step, no previous posterior state is available. We therefore initialize each non-observed resolution level by spectrally upsampling the first coarse observation,
\begin{equation}
x_0^{(r)}
=
\mathcal{U}_{0\to r}
\!\left(
y_0^{(0)}
\right),
\qquad r=1,\dots,R .
\end{equation}
For subsequent time steps, the posterior at each resolution is fed back into the shared mode-sliced FNO forecaster to form the next forecast prior,
\begin{equation}
\widetilde{x}_{t+1}^{(r)}
=
F_{\phi}^{[k_{\max}^{(r)}]}
\!\left(
x_t^{(r)}
\right),
\qquad r=1,\dots,R .
\end{equation}

The resulting inference procedure is a closed-loop learned assimilation cascade. At each physical time step, corrected states are first propagated forward in time by the shared FNO forecaster and then propagated upward across resolution levels by the diffusion-based analysis updates. This allows new coarse observations to be assimilated progressively while maintaining temporally coherent high-resolution reconstructions.

\subsection{Shared FNO Forecaster: Architecture and Spectral Mode Slicing}

The dynamical prior at every resolution level is provided by a shared Fourier Neural Operator (FNO)~\cite{li2021fourier}. 
The choice of FNO is motivated by its ability to learn mappings between function spaces through spectral parameterizations of integral operators, making it well suited for spatially extended PDE-governed systems~\cite{li2021fourier,kovachki2023neural}. In particular, FNOs capture long-range interactions efficiently by applying learned transformations in Fourier space while retaining local nonlinear transformations in physical space. Fourier/operator-based architectures have also been used successfully in large-scale physical forecasting settings, including data-driven weather prediction~\cite{pathak2022fourcastnet}.

\paragraph{Architecture}
The shared forecaster follows the standard FNO structure: a lifting layer, a sequence of Fourier layers, and a projection layer. The lifting layer maps the input physical field to a higher-dimensional latent representation. This latent state is then processed by multiple spectral convolution blocks, each combining a global Fourier-space operator with a local pointwise convolutional bypass:
\begin{equation}
z
=
\sigma\!\left(
\mathcal{K}_{\mathrm{spec}}^{[k_{\max}^{(r)}]}(h)
+
\mathcal{K}_{\mathrm{loc}}(h)
\right),
\end{equation}
where \(h\) denotes the incoming latent feature field, \(\sigma\) is a nonlinear activation, \(\mathcal{K}_{\mathrm{spec}}^{[k_{\max}^{(r)}]}\) is the mode-sliced spectral convolution, and \(\mathcal{K}_{\mathrm{loc}}\) is a pointwise convolution. The final projection maps the latent representation back to the physical state space.

For the 1D Burgers benchmark, the same design is implemented using one-dimensional Fourier transforms and one-dimensional pointwise convolutions. For the 2D Kraichnan benchmark, the corresponding two-dimensional FNO is used. In both cases, the sharing is performed across resolution levels within the same benchmark: one 1D shared FNO is used for the Burgers hierarchy, and one 2D shared FNO is used for the Kraichnan hierarchy.

\paragraph{Spectral mode slicing}
Let \(W_\ell\) denote the learned complex spectral weights in Fourier layer \(\ell\), parameterized at the maximum spectral bandwidth required by the finest resolution. At level \(r\), only the leading low-frequency block of these weights is used:
$
W_{\ell}^{(r)}
=
W_{\ell}
\big|_{|k|\le k_{\max}^{(r)}} ,
$ in the 1D case, and analogously
$
W_{\ell}^{(r)}
=
W_{\ell}
\big|_{|k_x|\le k_{\max,x}^{(r)},\; |k_y|\le k_{\max,y}^{(r)}}
$ in the 2D case. The spectral convolution at level \(r\) therefore applies the same learned Fourier operator, but restricted to the modes resolvable and active at that resolution.

\begin{figure}[htbp]
    \centering
    \includegraphics[width=\linewidth]{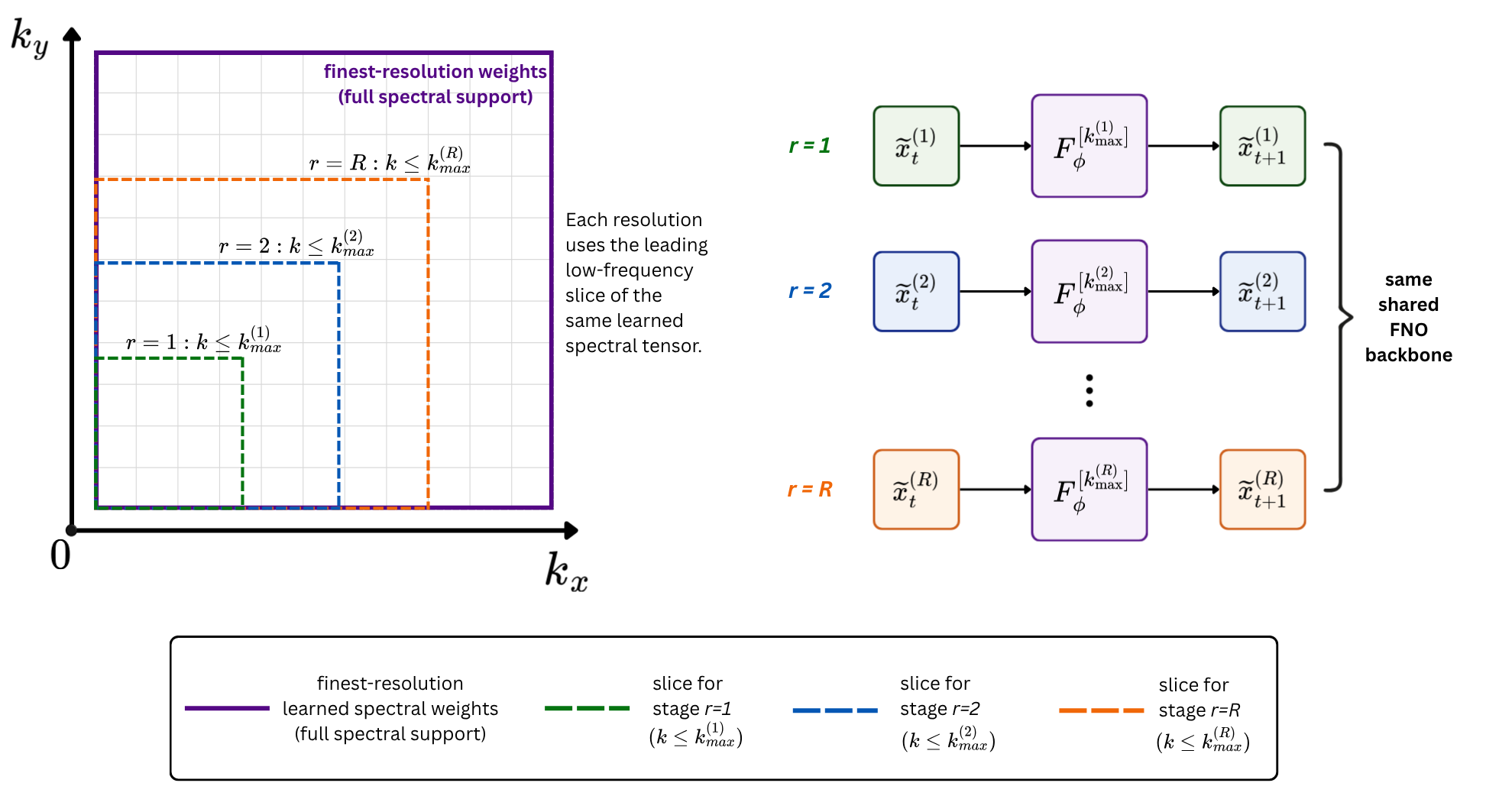}
    \caption{A single FNO is initialized at the finest spectral bandwidth; coarser-resolution inference uses progressively smaller low-frequency slices of the same learned weights.}
    \label{fig:k_mode_shared_forecaster}
\end{figure}

This slicing rule is natural for the multiresolution hierarchy because the lower-resolution fields contain only low-frequency information (as shown in Fig. \ref{fig:k_mode_shared_forecaster}). Coarser levels should not access spectral weights corresponding to modes that are not represented on their grids. Conversely, the finest level can use the full learned spectral bandwidth. The result is a single forecaster whose effective operator changes with resolution through the active Fourier support.
The shared mode-sliced design also reduces architectural redundancy across the hierarchy: the model learns one spectral forecasting operator that can be evaluated consistently at multiple resolutions.

\subsection{Shared Diffusion Corrector}

\begin{figure}[htbp]
  \centering
  \includegraphics[width=1.0\textwidth]{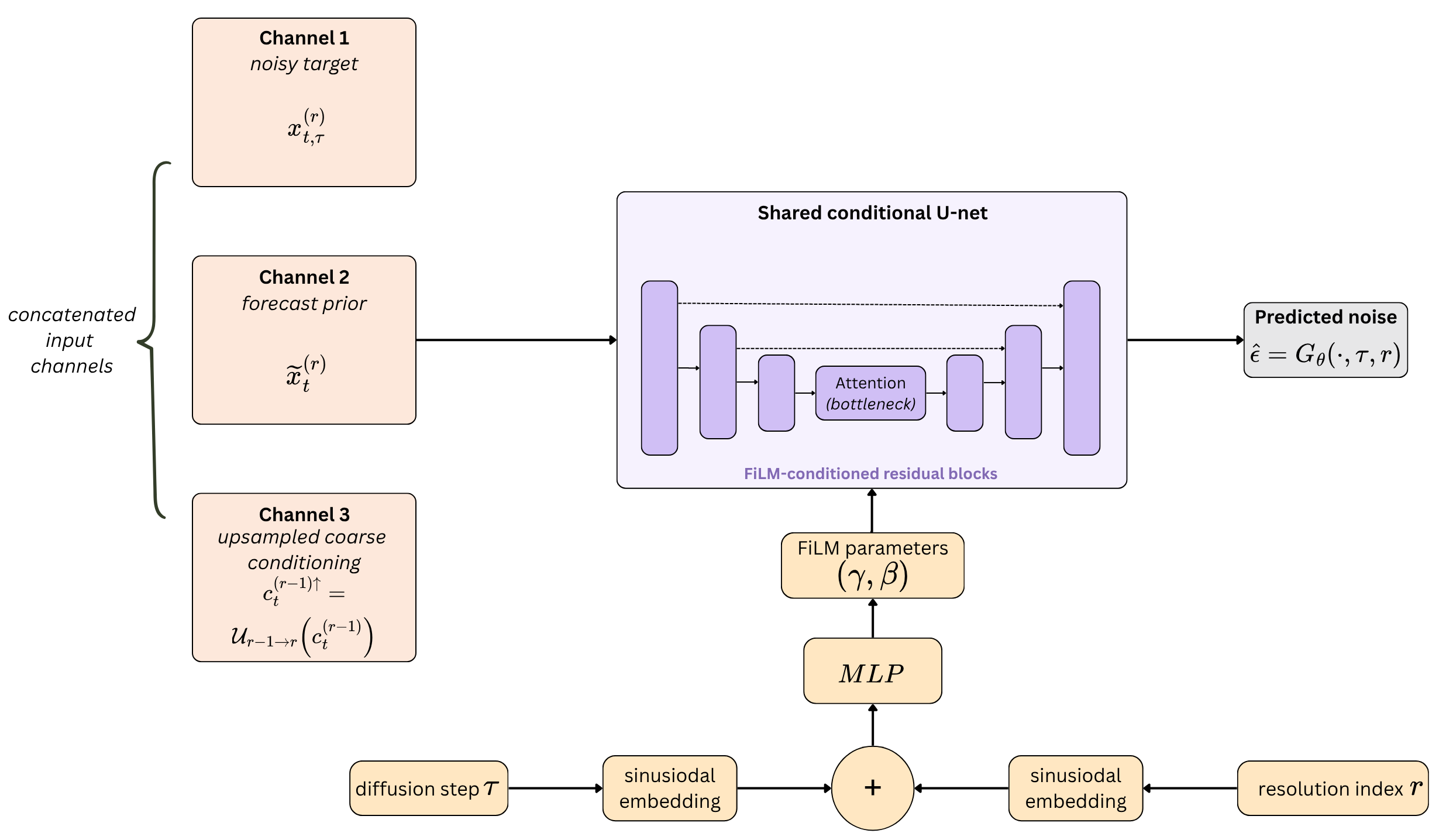}
  \caption{Shared conditional diffusion corrector used to implement the analysis step at each refinement stage. The denoiser receives three aligned spatial channels: the noisy target sample, the FNO forecast prior, and the upsampled conditioning field from the previous coarser level. These channels are concatenated and passed through a conditional U-Net that predicts the diffusion noise. The diffusion timestep and resolution index are embedded and injected into the network through FiLM conditioning, allowing the same corrector \(G_\theta\) to adapt across diffusion steps and resolution levels.}
  \label{fig:shared_diff}
\end{figure}

The analysis step at each refinement stage is implemented using a conditional denoising diffusion model~\cite{ho2020ddpm}. Given a forecast prior from the shared mode-sliced FNO and a conditioning signal from the previous coarser level, the corrector generates an analysis/posterior sample by modeling
\begin{equation}
x_t^{(r)}
\sim
p_\theta\!\left(
x_t^{(r)}
\,\middle|\,
\widetilde{x}_t^{(r)}, \;
c_t^{(r-1)\uparrow}, \;
r
\right),
\qquad r=1,\dots,R,
\end{equation}
where \(\widetilde{x}_t^{(r)}\) denotes the FNO forecast prior at level \(r\), and
\begin{equation}
c_t^{(r-1)\uparrow}
=
\mathcal{U}_{r-1\to r}
\!\left(
c_t^{(r-1)}
\right)
\end{equation}
is the lifted conditioning signal from the previous coarser level. This conditional distribution is realized using a shared neural network \(G_\theta\) that is reused across all refinement stages. While the FNO forecaster adapts across resolutions through spectral mode slicing, the diffusion corrector adapts through explicit conditioning on the resolution level \(r\).

\paragraph{Diffusion formulation}
We adopt the standard denoising diffusion probabilistic model (DDPM) framework~\cite{ho2020ddpm}. The forward process progressively perturbs the clean target state \(x_t^{(r)}\) with Gaussian noise:
\begin{equation}
x_{t,\tau}^{(r)}
=
\sqrt{\bar{\alpha}_{\tau}}\,x_t^{(r)}
+
\sqrt{1-\bar{\alpha}_{\tau}}\,\epsilon,
\qquad
\epsilon \sim \mathcal{N}(0,I),
\end{equation}
where \(\tau\) indexes the diffusion timestep and \(\bar{\alpha}_{\tau}\) is determined by a predefined noise schedule. In our implementation, we use a cosine noise schedule, following common practice in improved diffusion models~\cite{nichol2021improved}.

The denoising network is trained to predict the injected noise \(\epsilon\) from the corrupted sample \(x_{t,\tau}^{(r)}\), conditioned on the forecast prior, the lifted coarse state, the diffusion timestep \(\tau\), and the resolution index \(r\):
\begin{equation}
\widehat{\epsilon}
=
G_\theta
\!\left(
x_{t,\tau}^{(r)},
\widetilde{x}_t^{(r)},
c_t^{(r-1)\uparrow},
\tau,
r
\right).
\end{equation}
At inference, posterior samples are generated by iteratively denoising an initial Gaussian field using a DDIM-style sampler~\cite{song2021ddim}. The sampler includes a stochasticity parameter \(\eta\), with \(\eta=0\) corresponding to deterministic DDIM sampling and larger values introducing additional reverse-process noise. Because this sampling choice can affect both reconstruction quality and temporal smoothness, we report a sensitivity study over \(\eta\) and the number of DDIM reverse steps in ~\ref{app:sampler_sensitivity}. These results show that moderate stochasticity is particularly important for the 2D Kraichnan benchmark, while increasing the number of reverse steps beyond a moderate value yields diminishing accuracy gains relative to its added computational cost.

\paragraph{Input parameterization}
The corrector operates on three aligned spatial inputs, as shown in Fig.~\ref{fig:shared_diff}:
\begin{itemize}
    \item the noisy target state \(x_{t,\tau}^{(r)}\),
    \item the forecast prior \(\widetilde{x}_t^{(r)}\) produced by the shared mode-sliced FNO, and
    \item the upsampled coarse conditioning field \(c_t^{(r-1)\uparrow}\).
\end{itemize}
These inputs are concatenated along the channel dimension to form a 3-channel tensor. This parameterization explicitly exposes the denoiser to both sources of information required for the analysis step: the dynamical prior and the current coarse-scale constraint.

\paragraph{Conditional U-Net architecture}
The denoising network \(G_\theta\) is implemented as a dimension-matched conditional U-Net: a 1D conditional U-Net for the Burgers benchmark and a 2D conditional U-Net for the Kraichnan turbulence benchmark. U-Net architectures are well suited for denoising tasks because their encoder--decoder structure combines multiscale context with skip-connected spatial detail~\cite{ronneberger2015unet}. In both the 1D and 2D implementations, the architecture consists of:
\begin{itemize}
    \item an encoder with residual blocks and downsampling layers,
    \item a bottleneck with self-attention~\cite{vaswani2017attention},
    \item a decoder with skip connections and upsampling layers,
\end{itemize}
followed by a projection to a single-channel output corresponding to the predicted noise field.

All spatial convolutions use circular padding to respect the periodic boundary conditions of the underlying physical domain.

\paragraph{Conditioning mechanism}
The model is conditioned on both the diffusion timestep \(\tau\) and the resolution level \(r\). These are embedded into a fixed-dimensional latent vector and injected into each residual block using Feature-wise Linear Modulation (FiLM)~\cite{perez2018film}:
\begin{equation}
\mathrm{FiLM}(h)
=
(1+\gamma)\odot h + \beta,
\end{equation}
where \((\gamma,\beta)\) are functions of the timestep and resolution embeddings.

This mechanism allows the same corrector \(G_\theta\) to adapt its denoising behavior across diffusion time and across resolution levels. The resolution embedding is particularly important because the analysis task differs across refinement stages: early stages recover relatively coarse missing structure, while later stages refine smaller-scale details.



\section{Training Strategy}

\begin{figure}[htpb]
  \centering
  \includegraphics[width=\textwidth]{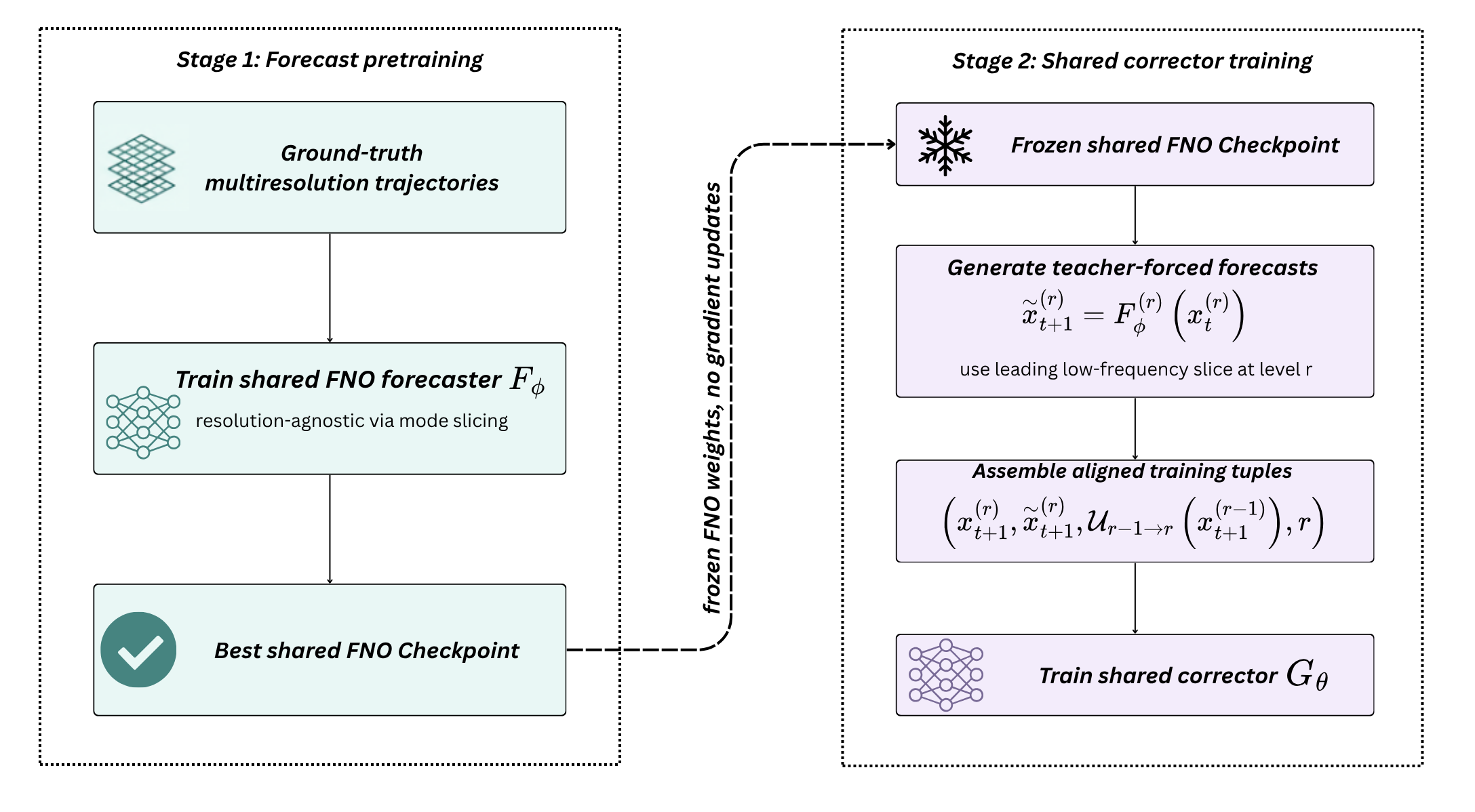}
  \caption{Two-stage training strategy for the forecast--analysis pipeline. In Stage 1, a single shared FNO forecaster is trained across resolution levels using mode slicing. In Stage 2, the FNO weights are frozen and the trained forecasters are run in teacher-forced mode to generate one-step forecast priors. These forecasts are time-aligned with the corresponding current coarse states and target fine states to form diffusion training tuples, which are then used to train the shared corrector $G_\theta$.}
  \label{fig:train_strategy}
  
\end{figure}

Training the proposed framework (Fig. \ref{fig:train_strategy}) requires more than independently fitting the forecasting and correction modules. Because the diffusion corrector operates on imperfect dynamical priors during inference, its training distribution must reflect the actual error characteristics of the learned forecasters. We therefore adopt a two-stage training strategy in which the shared mode-sliced FNO forecaster is first trained across all target resolutions on ground-truth one-step dynamics, after which their teacher-forced outputs are used to construct the training set for the shared diffusion corrector.

\subsection{Construction of the Multiresolution Training Dataset}

Let \(\{x_t^{(r)}\}_{t=0}^{T}\) denote a trajectory at resolution level \(r\), with \(r=0\) corresponding to the coarsest observable level and \(r=R\) corresponding to the finest target level. From each finest-resolution simulation, we construct a multiresolution pyramid using fixed resolution-transfer operators, yielding aligned trajectories at all intermediate scales. This produces, for every trajectory and time index, a set of states
\begin{equation}
\left\{
x_t^{(0)}, x_t^{(1)}, \dots, x_t^{(R)}
\right\}.
\end{equation}

The shared FNO forecaster and diffusion corrector are trained on different views of this hierarchy. The shared FNO forecaster is trained using consecutive ground-truth pairs at each non-observed resolution level. For a target level \(r=1,\dots,R\), the forecast training examples are
\begin{equation}
\left(
x_t^{(r)},\;
x_{t+1}^{(r)},\;
r
\right),
\end{equation}
where the level index \(r\) determines the active spectral slice \(k_{\max}^{(r)}\) used by the shared FNO.

The diffusion corrector is trained on tuples of the form
\begin{equation}
\left(
x_{t+1}^{(r)},\;
\widetilde{x}_{t+1}^{(r)},\;
c_{t+1}^{(r-1)\uparrow},\;
r
\right),
\qquad r=1,\dots,R,
\end{equation}
where \(x_{t+1}^{(r)}\) is the clean target field at the target resolution, \(\widetilde{x}_{t+1}^{(r)}\) is the teacher-forced forecast produced by the shared mode-sliced FNO, and
\begin{equation}
c_{t+1}^{(r-1)\uparrow}
=
\mathcal{U}_{r-1\to r}
\!\left(
c_{t+1}^{(r-1)}
\right)
\end{equation}
is the lifted conditioning field from the previous coarser level. During training, the conditioning signal is taken from the aligned ground-truth hierarchy:
\begin{equation}
c_{t+1}^{(r-1)}
=
\begin{cases}
y_{t+1}^{(0)}, & r=1,\\
x_{t+1}^{(r-1)}, & r=2,\dots,R.
\end{cases}
\end{equation}
Thus, the first refinement stage conditions on the observed coarse field, while later stages condition on the corresponding ground-truth state at the previous resolution level.

These tuples are aggregated across all trajectories, all time indices, and all resolution transitions into a shared training set for the diffusion corrector. This construction exposes the corrector to every refinement stage within a common formulation, enabling a single resolution-conditioned diffusion model to learn an analysis rule across the full hierarchy.

\subsection{Training of the Shared Forecast Model}

The forecast model is a single shared FNO \(F_{\phi}\) trained jointly across all non-observed resolution levels. For a training sample at target level \(r=1,\dots,R\), the model is evaluated using the spectral slice associated with that level:
\begin{equation}
\widetilde{x}_{t+1}^{(r)}
=
F_{\phi}^{[k_{\max}^{(r)}]}
\!\left(
x_t^{(r)}
\right).
\end{equation}
Here, \(F_{\phi}^{[k_{\max}^{(r)}]}\) denotes the shared FNO backbone evaluated using only the leading Fourier modes active at resolution \(r\). Coarser levels therefore use smaller low-frequency slices of the same learned spectral weights, while the finest level uses the largest spectral bandwidth.

The shared forecaster is optimized using a mean-squared one-step prediction loss aggregated across all active resolutions:
\begin{equation}
\mathcal{L}_{\mathrm{FNO}}
=
\mathbb{E}_{r,t}
\left[
\left\|
F_{\phi}^{[k_{\max}^{(r)}]}
\!\left(
x_t^{(r)}
\right)
-
x_{t+1}^{(r)}
\right\|_2^2
\right],
\qquad r=1,\dots,R.
\end{equation}
In practice, the expectation is approximated by sampling training pairs from the multiresolution dataset. Each sample specifies both a state pair and a target resolution level, which determines the mode slice used during the spectral convolution.

This joint objective trains a single forecasting operator to support one-step prediction across the entire hierarchy. The model is trained only at the target resolutions used during inference, excluding the coarsest observed level \(r=0\). After training, the best validation checkpoint of the shared FNO is retained. The forecaster is not used as a stand-alone final predictor in the proposed system; instead, it supplies the dynamical priors that are subsequently analyzed by the diffusion corrector.

\subsection{Training of the Diffusion Corrector}

Once the shared mode-sliced FNO forecaster has been trained, it is frozen and run in teacher-forced mode over the training and validation trajectories to generate forecast priors at each target resolution. For each trajectory, time index, and target level \(r=1,\dots,R\), the shared forecaster receives the ground-truth state at time \(t\) and predicts the next state:
\begin{equation}
\widetilde{x}_{t+1}^{(r)}
=
F_{\phi}^{[k_{\max}^{(r)}]}
\!\left(
x_t^{(r)}
\right).
\end{equation}
These teacher-forced forecasts are stored and paired with the aligned coarse conditioning states and ground-truth targets to train the shared diffusion corrector.

The corrector is trained using the standard DDPM epsilon-prediction objective. Given a clean target field \(x_{t+1}^{(r)}\), a diffusion timestep \(\tau\), and Gaussian noise \(\epsilon \sim \mathcal{N}(0,I)\), the forward process constructs
\begin{equation}
x_{t+1,\tau}^{(r)}
=
\sqrt{\bar{\alpha}_\tau}\,
x_{t+1}^{(r)}
+
\sqrt{1-\bar{\alpha}_\tau}\,
\epsilon .
\end{equation}
Here, \(x_{t+1,\tau}^{(r)}\) denotes the noisy version of the target state \(x_{t+1}^{(r)}\) at diffusion timestep \(\tau\).

The lifted conditioning field for target level \(r\) is
\begin{equation}
c_{t+1}^{(r-1)\uparrow}
=
\mathcal{U}_{r-1\to r}
\!\left(
c_{t+1}^{(r-1)}
\right),
\end{equation}
where
\begin{equation}
c_{t+1}^{(r-1)}
=
\begin{cases}
y_{t+1}^{(0)}, & r=1,\\
x_{t+1}^{(r-1)}, & r=2,\dots,R.
\end{cases}
\end{equation}

The denoiser then predicts the added noise under conditioning by the FNO forecast prior, the lifted coarse input, the diffusion timestep, and the target resolution index:
\begin{equation}
\mathcal{L}_{\mathrm{diff}}
=
\mathbb{E}_{r,t,\tau,\epsilon}
\left[
\left\|
\epsilon
-
G_\theta
\!\left(
x_{t+1,\tau}^{(r)},\;
\widetilde{x}_{t+1}^{(r)},\;
c_{t+1}^{(r-1)\uparrow},\;
k,\;
r
\right)
\right\|_2^2
\right],
\qquad r=1,\dots,R.
\end{equation}

A single diffusion corrector is trained jointly across all refinement stages. The target resolution index \(r\) is included explicitly in the conditioning pathway so that the same network parameters can adapt their denoising behavior to different scale transitions. In this way, the training pipeline mirrors the shared-model structure of the inference procedure: the FNO shares parameters through spectral mode slicing, while the diffusion corrector shares parameters through resolution-conditioned denoising.

\subsection{Forecast--Target Time Alignment}

A critical aspect of the training pipeline is the alignment between the teacher-forced forecasts and the corresponding supervision targets. Since the forecast generated from the ground-truth state at time \(t\) is intended to predict the state at time \(t+1\), the diffusion training tuples must be constructed with a one-step shift:
\begin{equation}
\widetilde{x}_{t+1}^{(r)}
=
F_{\phi}^{[k_{\max}^{(r)}]}
\!\left(
x_t^{(r)}
\right),
\qquad
\text{target}
=
x_{t+1}^{(r)},
\qquad
\text{conditioning}
=
c_{t+1}^{(r-1)\uparrow}.
\end{equation}
For the first refinement stage, the conditioning signal is obtained from the observed coarse field,
\begin{equation}
c_{t+1}^{(0)\uparrow}
=
\mathcal{U}_{0\to 1}
\!\left(
y_{t+1}^{(0)}
\right).
\end{equation}
For higher stages, the conditioning signal is obtained from the previous resolution level,
\begin{equation}
c_{t+1}^{(r-1)\uparrow}
=
\mathcal{U}_{r-1\to r}
\!\left(
x_{t+1}^{(r-1)}
\right),
\qquad r=2,\dots,R.
\end{equation}

Accordingly, the diffusion dataset pairs the stored forecast sequence indexed over \(t=0,\dots,T-1\) with the shifted ground-truth and coarse-resolution sequences indexed over \(t=1,\dots,T\). This time alignment ensures that the corrector learns to refine the actual one-step prior produced by the shared FNO, rather than an artificially synchronized or mismatched input.

This design choice is central to what we refer to as \emph{FNO-output teacher-forced training}: the corrector is trained on outputs of the learned shared forecaster itself, evaluated with the appropriate spectral mode slice at each resolution. As a result, the diffusion model learns to handle the characteristic forecast-error distribution of the mode-sliced FNO prior before it is deployed in the closed-loop sequential inference pipeline.

\subsection{Resolution-Balanced Batch Sampling and Input Augmentation}

The shared corrector is trained on samples drawn from multiple refinement stages, each with different spatial dimensions. To preserve efficient batching while keeping each batch shape-consistent, we use a resolution-aware batch sampler that groups training examples by refinement stage. Every mini-batch therefore contains samples from exactly one resolution transition, while training cycles across all stages in a balanced manner.

This batching strategy serves two purposes. First, it avoids the need for padding or resizing across incompatible spatial grids. Second, it prevents the training process from being dominated by the most abundant or cheapest stage, thereby encouraging uniform performance across the full cascade.

To improve robustness to inference-time mismatch, we optionally perturb the conditioning inputs during diffusion training. Specifically, additive Gaussian noise is injected into the forecast prior and coarse conditioning channels with a prescribed probability, using stage-dependent noise magnitudes. This augmentation is motivated by the fact that, during inference, the corrector does not receive perfect conditioning: the forecast prior is generated from previous posteriors rather than ground truth, and higher stages condition on earlier refined posteriors rather than exact coarse states. Input perturbation therefore acts as a simple robustness mechanism that exposes the model to mild imperfections during training.

\subsection{Optimization and Implementation Details}

The forecasting and correction modules are optimized separately. The shared FNO forecaster is trained with AdamW~\cite{loshchilov2019decoupled} and a cosine-decay learning-rate schedule~\cite{loshchilov2017sgdr}, using validation loss to select the best checkpoint. The diffusion corrector is optimized with AdamW under a linear warmup followed by cosine decay. During diffusion training, we use mixed-precision arithmetic~\cite{micikevicius2018mixed}, gradient clipping~\cite{pascanu2013difficulty}, and an exponential moving average (EMA) of network parameters, a parameter-averaging strategy related to classical stochastic approximation averaging~\cite{polyak1992acceleration}. The EMA weights are maintained throughout training and used at inference time, as they consistently yield more stable sampling behavior.

The diffusion model is trained over a fixed noise horizon using a cosine noise schedule, following improved diffusion-model practice~\cite{nichol2021improved}, while inference is performed with a DDIM sampler configured with a reduced number of reverse steps~\cite{song2021ddim}. Checkpoints are saved periodically throughout training, and the final deployed corrector corresponds to the EMA-smoothed model parameters. The same overall training protocol is used in both the 1D and 2D testbeds, with problem-specific hyperparameters adjusted only for data dimensionality and computational scale. Complete architectural details along with the training information is presented in \ref{tab:hyperparams_arch} and \ref{tab:hyperparams_training}.

\section{Experimental Setup}
\subsection{Testbeds}

We evaluate the proposed iterative refinement framework on two multiscale dynamical systems of increasing difficulty: the 1D stochastically forced Burgers equation and 2D Kraichnan turbulence. The first serves as a controlled proof-of-concept benchmark with sharp shock-like structures and temporally evolving fine-scale ambiguity, while the second provides a substantially more challenging turbulence setting in which coherent vortical structures and thin filaments coexist across a broad range of active scales~\cite{bec2007burgers,kraichnan1967inertial,boffetta2012two}. In both cases, high-resolution trajectories are generated using pseudo-spectral solvers and then converted into multiresolution datasets for training and evaluation~\cite{canuto2006spectral}. Full PDE parameters and dataset configuration for the two testbeds are shown in \ref{tab:hyperparams_data}.

\subsubsection{1D Stochastic Burgers Equation}

As a first benchmark, we consider the stochastically forced one-dimensional Burgers equation on a periodic domain,
\begin{equation}
\frac{\partial u}{\partial t}
+
u \frac{\partial u}{\partial x}
=
\nu \frac{\partial^2 u}{\partial x^2}
+
f(x,t),
\qquad x \in [0,2\pi].
\end{equation}
The Burgers equation is a canonical nonlinear advection--diffusion model and has long served as a simplified setting for studying shock formation, dissipation, and turbulence-like behavior~\cite{burgers1974nonlinear,bec2007burgers}. The system is evolved using a pseudo-spectral solver on a fine grid with $N=512$ points. Diffusion is handled implicitly through a Crank--Nicolson step in Fourier space, while the nonlinear advection and stochastic forcing are advanced explicitly with a fourth-order Runge--Kutta scheme. The nonlinear term is written in conservative form, and a standard $2/3$ dealiasing rule is applied in Fourier space~\cite{canuto2006spectral}.

The forcing is modeled as a finite set of Ornstein--Uhlenbeck processes acting on the first few Fourier modes, producing temporally correlated stochastic excitation~\cite{uhlenbeck1930brownian}. In the implementation used here, the forcing acts on the first $16$ modes with amplitudes that decay as $k^{-1}$, thereby sustaining statistically stationary dynamics while continually generating and interacting shock structures. The viscosity is set to $\nu = 5\times 10^{-3}$, and the internal solver time step is $\Delta t = 10^{-4}$. Trajectories are saved every $500$ solver steps, corresponding to an output cadence of $\delta = 0.05$. 

This system is a useful testbed because coarse observations are often insufficient to uniquely determine the location, steepness, and merger history of shocks. While low-resolution inputs constrain the large-scale profile, temporal information is essential for recovering fine-scale structure consistently over time. Following the data-generation setup used in this work, we construct a resolution hierarchy
\begin{equation}
64 \rightarrow 128 \rightarrow 256 \rightarrow 512,
\end{equation}
and use the coarsest level as the observed input and the finest level as the reconstruction target. The dataset contains $50$ trajectories, split into $40$ training, $5$ validation, and $5$ test trajectories, with $800$ saved snapshots per trajectory after an initial burn-in period. 

\subsubsection{2D Kraichnan Turbulence}

Our primary benchmark is forced two-dimensional turbulence in vorticity form,
\begin{equation}
\frac{\partial \omega}{\partial t}
+
J(\psi,\omega)
=
\nu \nabla^2 \omega
-
\mu \omega
+
f(x,y,t),
\qquad (x,y)\in[0,2\pi]^2,
\end{equation}
where $\omega$ is vorticity, $\psi$ is the streamfunction, and $J(\psi,\omega)$ denotes the Jacobian nonlinearity. Two-dimensional turbulence is a classical setting in which energy and enstrophy are transferred across scales through distinct cascade processes~\cite{kraichnan1967inertial,batchelor1969energy,boffetta2012two}. The system is evolved on a doubly periodic $256\times256$ grid using a pseudo-spectral solver with fourth-order Runge--Kutta time integration and standard $2/3$ dealiasing~\cite{canuto2006spectral}. The streamfunction is recovered in Fourier space through the Poisson relation, and the velocity field used in the nonlinear advection term is obtained spectrally from $\psi$.

The forcing is band-limited in Fourier space and updated through an Ornstein--Uhlenbeck-type process with finite temporal correlation~\cite{uhlenbeck1930brownian}. In the implementation used here, the forcing is concentrated around wavenumber $k_f \approx 4$ with a narrow spectral width and correlation time $\tau = 0.5$. The viscosity and Ekman drag are set to $\nu = 10^{-3}$ and $\mu = 0.05$, respectively, and the solver time step is $\Delta t = 0.005$. Snapshots are saved every $10$ solver steps, corresponding to an output cadence of $\delta = 0.05$. An initial burn-in phase is performed before saving begins in order to reach a statistically stationary regime. 

This testbed is significantly more demanding than Burgers because energetically relevant structures exist across all scales. The inverse energy cascade and direct enstrophy cascade produce coherent vortices, thin filaments, and merger debris that are only partially visible at coarse resolution~\cite{kraichnan1967inertial,boffetta2012two}. As a result, reconstructing the fine state from a coarse snapshot alone is highly ambiguous, while the previous high-resolution state provides strong temporal constraints on the admissible filament geometry and small-scale organization. For this problem, we use the resolution hierarchy
\begin{equation}
32\times32 \rightarrow 64\times64 \rightarrow 128\times128 \rightarrow 256\times256,
\end{equation}
with the $32\times32$ field treated as the observed input and the $256\times256$ field taken as the target state. The dataset contains $50$ trajectories, split into $44$ training, $5$ validation, and $3$ test trajectories, with $200$ saved snapshots per trajectory after burn-in. 

\subsection{Spectral downsampling and hierarchy construction}

The multiresolution hierarchy is constructed by spectrally coarse-graining the highest-resolution field. This is natural for the present testbeds because the underlying solvers are pseudo-spectral and the fields are periodic. Let $x^{(R)}$ denote a field on the finest grid, and let $x^{(r)}$ denote its representation on a coarser grid with fewer spatial points. For each target resolution, the coarse field is obtained by truncating high-frequency Fourier modes and transforming back to physical space, following standard Fourier spectral discretization principles~\cite{canuto2006spectral,trefethen2000spectral}.

\textbf{1D case.}
Let $u \in \mathbb{R}^{N_{\mathrm{src}}}$ be a periodic field on a uniform grid, and let $\widehat{u}_k$ denote its discrete Fourier coefficients. To construct a coarse representation on a grid of size $N_{\mathrm{tgt}} < N_{\mathrm{src}}$, we retain only the Fourier modes resolvable on the target grid:
\begin{equation}
\widehat{u}^{\downarrow}_k =
\widehat{u}_k,
\qquad
0 \le k \le \frac{N_{\mathrm{tgt}}}{2},
\end{equation}
and discard all higher modes. The downsampled field is then defined by
\begin{equation}
u^{\downarrow}
=
\frac{N_{\mathrm{tgt}}}{N_{\mathrm{src}}}
\,
\mathcal{F}^{-1}_{N_{\mathrm{tgt}}}
\!\left(
\widehat{u}^{\downarrow}
\right),
\end{equation}
where $\mathcal{F}^{-1}_{N_{\mathrm{tgt}}}$ denotes the inverse discrete Fourier transform on the target grid. The factor $N_{\mathrm{tgt}}/N_{\mathrm{src}}$ preserves the physical amplitude of the field across resolutions.

\textbf{2D case.}
Let $\omega \in \mathbb{R}^{N_y^{\mathrm{src}} \times N_x^{\mathrm{src}}}$ be a doubly periodic field, and let
\begin{equation}
\widehat{\omega} = \mathcal{F}_{2}\!\left(\omega\right)
\end{equation}
denote its 2D Fourier transform. To obtain a coarse field on a grid
$N_y^{\mathrm{tgt}} \times N_x^{\mathrm{tgt}}$, we truncate the spectrum to the Fourier modes resolvable on the target grid. Using the real FFT representation, this requires retaining:
\begin{itemize}
    \item the low nonnegative modes in the $k_x$ direction, and
    \item both the positive and negative low modes in the $k_y$ direction.
\end{itemize}
Accordingly, if $\widehat{\omega}$ is stored in $\mathrm{rfft2}$ format, the truncated spectrum $\widehat{\omega}^{\downarrow}$ is formed by
\begin{align}
\widehat{\omega}^{\downarrow}[0:\tfrac{N_y^{\mathrm{tgt}}}{2},\; 0:\tfrac{N_x^{\mathrm{tgt}}}{2}+1]
&=
\widehat{\omega}[0:\tfrac{N_y^{\mathrm{tgt}}}{2},\; 0:\tfrac{N_x^{\mathrm{tgt}}}{2}+1], \\
\widehat{\omega}^{\downarrow}[-\tfrac{N_y^{\mathrm{tgt}}}{2}:,\; 0:\tfrac{N_x^{\mathrm{tgt}}}{2}+1]
&=
\widehat{\omega}[-\tfrac{N_y^{\mathrm{tgt}}}{2}:,\; 0:\tfrac{N_x^{\mathrm{tgt}}}{2}+1],
\end{align}
with all remaining entries set to zero. The coarse field is then reconstructed as
\begin{equation}
\omega^{\downarrow}
=
\frac{N_y^{\mathrm{tgt}} N_x^{\mathrm{tgt}}}
     {N_y^{\mathrm{src}} N_x^{\mathrm{src}}}
\,
\mathcal{F}^{-1}_{2,\mathrm{tgt}}
\!\left(
\widehat{\omega}^{\downarrow}
\right).
\end{equation}
The multiplicative factor
\(
(N_y^{\mathrm{tgt}} N_x^{\mathrm{tgt}})
/
(N_y^{\mathrm{src}} N_x^{\mathrm{src}})
\)
is required to preserve the physical amplitude of the field under resolution transfer.

In both 1D and 2D, the hierarchy is constructed by applying this spectral restriction directly from the finest-resolution trajectory to each target level, rather than by repeated chained downsampling. Denoting the finest state by $x_t^{(R)}$, the field at level $r$ is therefore defined as
\begin{equation}
x_t^{(r)} = \mathcal{D}_{R \to r}\!\left(x_t^{(R)}\right),
\qquad r=0,\dots,R-1,
\end{equation}
where $\mathcal{D}_{R \to r}$ is the spectral truncation operator described above. This yields a resolution pyramid that is consistent with the Fourier structure of the underlying pseudo-spectral simulations. The truncation of unresolved high-wavenumber modes is also consistent with standard spectral filtering ideas used to control aliasing and remove non-resolvable components~\cite{orszag1971aliasing,canuto2006spectral}.  

\subsection{Baselines}

We compare the proposed iterative refinement framework against three baselines (features distinguished in Table \ref{tab:baseline_matrix}) designed to isolate the contribution of hierarchical correction, temporal conditioning, and learned dynamics (learnable parameters shown in \ref{tab:params}).

\begin{table}[htbp]
\centering

\begin{tabular}{lcccc}
\toprule
Method & Temporal prior & Obs.\ correction & Hierarchy & Stochastic \\
\midrule
Spectral upsampling      & --  & --  & --  & -- \\
EDSR                     & --  & \checkmark & -- & -- \\
One-shot diffusion       & -- & \checkmark & -- & \checkmark \\
FNO-only forecaster      & \checkmark & --  & --  & -- \\
\textbf{Iterative refinement} & \checkmark & \checkmark & \checkmark & \checkmark \\
\bottomrule
\end{tabular}
\caption{Comparison of baseline methods.}
\label{tab:baseline_matrix}
\end{table}

\subsubsection{Spectral Upsampling}

The simplest baseline reconstructs the target-resolution field by directly upsampling the coarse observation without using temporal information or a learned model. Since the data are generated and downsampled spectrally, this baseline is implemented by zero-padding the coarse Fourier coefficients to the target resolution and transforming back to physical space. Let $x_t^{(0)}$ denote the coarsest observed field. The reconstructed target field is given by
\begin{equation}
\hat{x}_t^{(R)} = \mathcal{U}_{0 \to R}\!\left(x_t^{(0)}\right),
\end{equation}
where $\mathcal{U}_{0 \to R}$ denotes the spectral upsampling operator from the observation grid to the finest-resolution grid.

This baseline preserves the Fourier modes present in the coarse observation but introduces no learned or dynamically inferred fine-scale content beyond the observed bandwidth. It therefore provides a non-learned reference for assessing whether the learned methods genuinely recover unresolved high-wavenumber structure.

\subsubsection{FNO-Only Autoregressive Forecasting}

To assess the value of the diffusion-based analysis stage, we evaluate a purely dynamical baseline that uses the same shared FNO forecaster as the proposed method, but omits the diffusion corrector entirely. At each target resolution, the trajectory is initialized from the upsampled first coarse observation and then advanced autoregressively:
\begin{equation}
\widehat{x}_0^{(r)}
=
\mathcal{U}_{0 \to r}\!\left(y_0^{(0)}\right),
\qquad
\widehat{x}_t^{(r)}
=
F_{\phi}^{[k_{\max}^{(r)}]}
\!\left(
\widehat{x}_{t-1}^{(r)}
\right),
\quad t \ge 1 .
\end{equation}

This baseline uses the identical shared mode-sliced FNO forecaster as the iterative refinement model and therefore isolates the benefit of assimilating current observations through the diffusion corrector. In practice, it tests whether learned dynamics alone are sufficient for long-horizon reconstruction in multiscale systems. 

\subsubsection{One-Shot Diffusion Super-Resolution}

We also compare against a one-shot generative baseline that maps directly from the coarsest observation to the finest resolution in a single diffusion pass, conditioned on the previous finest-resolution state. For the 2D case, this baseline reconstructs a $256\times256$ field directly from the upsampled $32\times32$ observation:
\begin{equation}
x_t^{(R)} \sim p_{\phi}\!\left(
x_t^{(R)}
\mid
x_{t-1}^{(R)},
\mathcal{U}_{0 \to R}(x_t^{(0)})
\right).
\end{equation}

Unlike the proposed method, the one-shot model does not perform intermediate refinement across scales and does not use a resolution-conditioned shared corrector. Its denoising network is architecturally similar to the proposed diffusion corrector but slightly larger in capacity, so that the comparison does not unfairly favor the iterative model through parameter count alone. In particular, the one-shot baseline uses a wider U-Net and conditions only on diffusion timestep, since no multiresolution hierarchy is present. 

This baseline isolates the benefit of hierarchical decomposition. If one-shot generation performs comparably to iterative refinement, then the intermediate forecast--analysis cascade is unnecessary; if not, the comparison supports the claim that multistage refinement substantially improves conditioning and reconstruction fidelity.

\subsubsection{Deterministic EDSR Super-Resolution}

We additionally compare against a deterministic one-shot super-resolution baseline based on the Enhanced Deep Super-Resolution (EDSR) architecture~\cite{lim2017edsr}. Unlike the diffusion-based baselines, this model produces a single deterministic reconstruction and does not sample from a posterior distribution. For the 2D Kraichnan benchmark, the model maps the coarsest $32\times32$ vorticity observation directly to the target $256\times256$ field:
\begin{equation}
\hat{x}_t^{(R)} = S_{\psi}\!\left(x_t^{(0)}\right),
\end{equation}
where $S_{\psi}$ denotes the trained EDSR super-resolution network.

The EDSR baseline uses residual blocks without batch normalization, residual scaling for stable training, and sub-pixel convolution via pixel-shuffle layers to perform the $8\times$ upsampling. To respect the doubly periodic domain, all spatial convolutions use circular padding. Importantly, the model uses no temporal context: each frame is super-resolved independently from the current coarse observation. This baseline therefore isolates the value of temporal conditioning and sequential correction relative to a strong deterministic image-style super-resolution model.

\subsection{Evaluation Metrics}

We evaluate all methods using complementary metrics that quantify reconstruction accuracy, spectral fidelity, and temporal coherence. Let $\hat{x}_{i,t}$ denote the predicted field for trajectory $i$ at time $t$, and let $x_{i,t}$ denote the corresponding ground-truth field.

\subsubsection{Root Mean Squared Error}

Our primary scalar metric is the root mean squared error (RMSE), computed over all spatial degrees of freedom. For a single trajectory, this is defined as
\begin{equation}
\mathrm{RMSE}(\hat{x},x)
=
\sqrt{
\frac{1}{TN}
\sum_{t=1}^{T}
\left\|
\hat{x}_t - x_t
\right\|_2^2
},
\end{equation}
where $N$ is the number of spatial grid points at the evaluated resolution. In reporting aggregate performance, we compute RMSE separately for each test trajectory and then report the mean and standard deviation across trajectories. This provides both an average error level and a measure of trajectory-to-trajectory stability.

\subsubsection{RMSE Over Time}

To assess temporal error growth or stabilization, we also compute the per-timestep RMSE profile
\begin{equation}
\mathrm{RMSE}_t
=
\sqrt{
\frac{1}{N}
\left\|
\hat{x}_t - x_t
\right\|_2^2
},
\qquad t=1,\dots,T.
\end{equation}
This yields an error curve over the rollout horizon and is particularly informative for comparing the stability of autoregressive baselines against corrected sequential methods.

\subsubsection{Spectral RMSE}

Since both benchmarks involve multiscale dynamics, we measure error in Fourier space in addition to physical space. Let $\widehat{x}_t(k)$ and $\widehat{\hat{x}}_t(k)$ denote the Fourier coefficients of the true and predicted fields. The spectral RMSE is computed mode-wise as
\begin{equation}
\mathrm{SpecRMSE}(k)
=
\sqrt{
\frac{1}{T}
\sum_{t=1}^{T}
\left|
\widehat{\hat{x}}_t(k) - \widehat{x}_t(k)
\right|^2
}.
\end{equation}
This metric reveals whether a method reproduces the correct scale-dependent energy content, especially at high wavenumbers where simple interpolation and one-shot super-resolution methods may fail.

\subsubsection{Temporal Consistency}

To quantify frame-to-frame smoothness and dynamical coherence, we compute the temporal consistency of a predicted trajectory as the average displacement between consecutive reconstructed states:
\begin{equation}
\mathrm{TC}(\hat{x})
=
\frac{1}{T-1}
\sum_{t=1}^{T-1}
\left\|
\hat{x}_{t+1} - \hat{x}_t
\right\|_2.
\end{equation}
We again report the mean and standard deviation across test trajectories. This metric is not an accuracy measure by itself; rather, it characterizes the temporal behavior of the predicted sequence and helps identify unstable or overly noisy reconstructions. 

\subsubsection{Reported Statistics}

For each evaluated resolution, we report the mean and standard deviation of the per-trajectory RMSE. When relevant, we also compare posterior reconstructions against the FNO-only baseline and the raw FNO forecast prior before correction. In this way, the reported metrics distinguish between errors due to imperfect dynamical forecasting and errors remaining after generative assimilation.

\section{Results}

We begin by comparing the proposed iterative refinement method against all baselines on aligned evaluation subsets for the 1D and 2D benchmarks. The 2D benchmark follows the Kraichnan turbulence example used in SuperBench, while one 1D benchmark uses a Burgers turbulence setting related to prior work on Burgers turbulence \cite{ren2025superbench,dhingra2024accelerated}. In each case, we report the mean RMSE and the standard deviation of per-trajectory RMSE over the selected test subset. These tables provide the clearest first comparison of reconstruction accuracy and rollout stability before turning to more detailed analyses of spectral behavior and temporal consistency.

\subsection{Results on 1D Stochastic Burgers}

Table~\ref{tab:burgers-benchmark} summarizes performance on the 1D Burgers benchmark, evaluated on an aligned subset of $5$ trajectories over $700$ time steps at the finest resolution of $512$ grid points. The results show that both diffusion-based methods substantially outperform the deterministic baselines. The one-shot diffusion super-resolution model achieves the lowest RMSE, with an error of $2.776\times10^{-3}$, followed by the proposed iterative refinement method with an RMSE of $4.355\times10^{-3}$. Both methods provide large improvements over EDSR and spectral upsampling, whose RMSE values are $2.1447\times10^{-2}$ and $8.4944\times10^{-2}$, respectively.

The comparison also highlights the relative behavior of deterministic and generative reconstruction methods in the corrected 1D Burgers setting. EDSR improves substantially over spectral upsampling, confirming that a learned deterministic super-resolution model can recover useful fine-scale structure from the coarse input. However, it remains less accurate than the diffusion-based approaches, indicating the value of probabilistic generative reconstruction for sharp-gradient dynamics. The FNO-only autoregressive forecaster fails catastrophically, yielding undefined RMSE values due to unstable rollout behavior. This confirms that learned dynamics alone are insufficient for robust long-horizon reconstruction without recurrent observation-driven correction or frame-wise generative reconstruction.

\begin{table}[htbp]
\centering
\caption{Benchmark results on the 1D stochastic Burgers testbed, evaluated on an aligned subset of $5$ trajectories over $700$ time steps at resolution $512$. Values are reported as mean RMSE with the standard deviation across trajectories shown in parentheses. Lower RMSE is better.}
\label{tab:burgers-benchmark}
\begin{tabular}{lc}
\toprule
Method & RMSE \\
\midrule
Spectral Upsample & 0.084944 (0.017270) \\
EDSR & 0.021447 (0.004078) \\
\textbf{One-Shot SR} & \textbf{0.002776} (0.000245) \\
Iterative Refinement & 0.004355 (0.000254) \\
FNO-only forecaster & -- \\
\bottomrule
\end{tabular}
\end{table}

\subsection{Results on 2D Kraichnan Turbulence}

Table~\ref{tab:kraichnan-benchmark} reports results on the 2D Kraichnan turbulence benchmark, evaluated on an aligned subset of $3$ trajectories over $200$ time steps at the finest resolution of $256\times256$. The proposed iterative refinement method achieves the best overall reconstruction quality, obtaining the lowest RMSE of $0.18401$ and the highest SSIM of $0.835870$. This improves substantially over spectral upsampling, one-shot diffusion super-resolution, deterministic EDSR, and the FNO-only autoregressive forecaster.

The advantage of iterative refinement is most pronounced in this 2D turbulence setting because the full $32\times32 \rightarrow 256\times256$ reconstruction is a highly ill-posed multiscale inverse problem. Thin vorticity filaments, vortex interfaces, and merger debris are not uniquely determined by the coarse observation alone. One-shot methods must infer the entire missing range of scales in a single step, whereas iterative refinement decomposes the task into a sequence of smaller resolution-wise correction problems. Each stage only has to recover one band of unresolved structure while being conditioned on both a learned dynamical prior and the current coarser-scale posterior. This makes the conditional reconstruction problem better posed and leads to more accurate recovery of fine-scale turbulent structure.

The deterministic EDSR baseline performs competitively among the non-diffusion one-shot methods, achieving lower RMSE than both spectral upsampling and one-shot diffusion. However, it still falls short of iterative refinement in both RMSE and SSIM. This suggests that while a strong deterministic super-resolution network can recover useful spatial structure from the coarse field, the sequential forecast--analysis cascade provides additional benefit by incorporating temporal dynamics and hierarchical correction. The FNO-only forecaster, while able to produce finite outputs in the 2D case, accumulates large autoregressive error and performs dramatically worse than methods that use current observational information.

\begin{table}[htbp]
\centering
\caption{Benchmark results on the 2D Kraichnan turbulence testbed, evaluated on an aligned subset of $3$ trajectories over $200$ time steps at resolution $256\times256$. Values are reported as mean with the standard deviation across trajectories shown in parentheses. Lower RMSE is better; higher SSIM is better.}
\label{tab:kraichnan-benchmark}
\begin{tabular}{lcc}
\toprule
Method & RMSE & SSIM \\
\midrule
Spectral Upsample & 0.335469 (0.008368) & 0.669100 (0.110947) \\
One-Shot SR & 0.349025 (0.127302) & 0.748400 (0.078439) \\
EDSR & 0.238840 (0.004079) & 0.786113 (0.003680) \\
\textbf{Iterative Refinement} & \textbf{0.18401} (0.013212) & \textbf{0.835870} (0.004520) \\
FNO-only forecaster & 2.568060 (0.198535) & -- \\
\bottomrule
\end{tabular}
\end{table}

Taken together with the 1D Burgers results, these experiments show that the relative benefit of iterative refinement depends on the difficulty of the underlying reconstruction problem. In the corrected 1D Burgers setting, one-shot diffusion performs extremely well and slightly outperforms iterative refinement in RMSE. In the 2D Kraichnan case, however, the hierarchical forecast--analysis formulation is clearly superior, reflecting the greater ambiguity and multiscale complexity of turbulent reconstruction. This supports the main premise of the proposed framework: iterative refinement becomes increasingly valuable as the coarse-to-fine inverse problem becomes more strongly multiscale and underdetermined.

\subsection{Comparison Against Baselines}
\label{sec:baseline_comparison}

We next compare iterative refinement against two important classes of baselines: autoregressive forecasting without an analysis step, and one-shot super-resolution methods that reconstruct the finest-resolution state directly from the current coarse observation.

\paragraph{Autoregressive forecasting without analysis}
The FNO-only baseline isolates the role of repeated observation-driven analysis. This baseline uses the same shared mode-sliced FNO forecaster as the proposed method, but removes the diffusion corrector and rolls the learned dynamics forward autoregressively. As shown quantitatively in Tables~\ref{tab:burgers-benchmark} and~\ref{tab:kraichnan-benchmark}, this is insufficient for stable long-horizon reconstruction. In the 1D Burgers case, the FNO-only trajectory becomes unstable and eventually blows up, while in the 2D Kraichnan case it accumulates large autoregressive error and produces substantially worse RMSE than all methods that use current observational information. Additional qualitative diagnostics of these FNO-only failure modes are provided in ~\ref{app:fno_only_diagnostics}.

This behavior confirms that the learned forecaster should not be interpreted as a stand-alone high-resolution predictor. In the proposed framework, its role is to provide a dynamical prior that is repeatedly analyzed using current coarse observations. The diffusion corrector is therefore essential for suppressing forecast drift and maintaining consistency with the observed trajectory.

\paragraph{One-shot super-resolution baselines}
We also compare against two one-shot super-resolution baselines: deterministic EDSR and stochastic one-shot diffusion. These methods reconstruct the finest-resolution state directly from the current coarse observation, without using a multiresolution forecast--analysis cascade. This comparison isolates the benefit of progressive hierarchical analysis relative to direct coarse-to-fine reconstruction.

The 1D Burgers benchmark shows that the benefit of hierarchy depends on the difficulty of the inverse problem. In this setting, the coarse-to-fine reconstruction problem is sufficiently constrained that one-shot diffusion achieves the lowest RMSE, while iterative refinement remains close behind and substantially outperforms deterministic EDSR and spectral upsampling. Additional 1D snapshot and space--time comparisons are provided in ~\ref{app:oneshot_baseline_diagnostics}.

\begin{figure}[htbp]
  \centering
  \includegraphics[width=\textwidth]{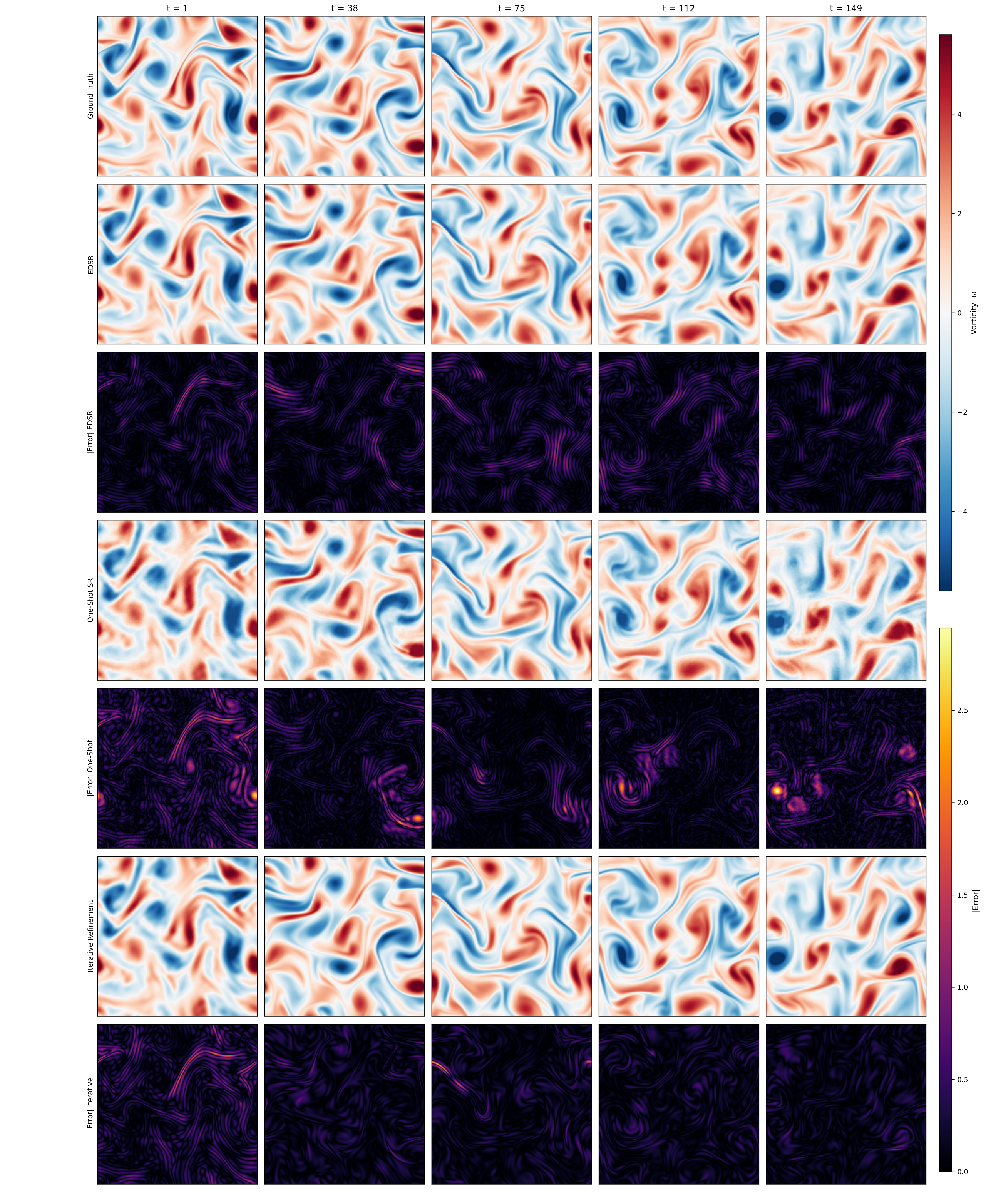}
  \caption{Temporal rollout comparison on the 2D Kraichnan turbulence benchmark at resolution \(256\times256\) for a representative trajectory. Rows show the ground truth, EDSR reconstruction and error, one-shot diffusion reconstruction and error, and iterative refinement reconstruction and error. While both one-shot baselines recover the large-scale vortex organization, iterative refinement more accurately preserves fine filaments and local vortex geometry, yielding consistently smaller residual error throughout the rollout.}
  \label{fig:2d_all_methods_rollout}
\end{figure}

The 2D Kraichnan benchmark tells a different story. Figure~\ref{fig:2d_all_methods_rollout} shows that both one-shot baselines produce visually plausible reconstructions and capture much of the large-scale vortex organization. However, their residual errors remain consistently larger than those of iterative refinement throughout the rollout. EDSR tends to produce smoother fields and misses some fine filamentary structure, while one-shot diffusion recovers sharp features in some regions but introduces less stable fine-scale texture. Iterative refinement achieves the best balance: it preserves the large-scale flow organization while more accurately reconstructing thin filaments, compact vortices, and stretched interfaces.

\begin{figure}[htbp]
  \centering
  \includegraphics[width=0.95\textwidth]{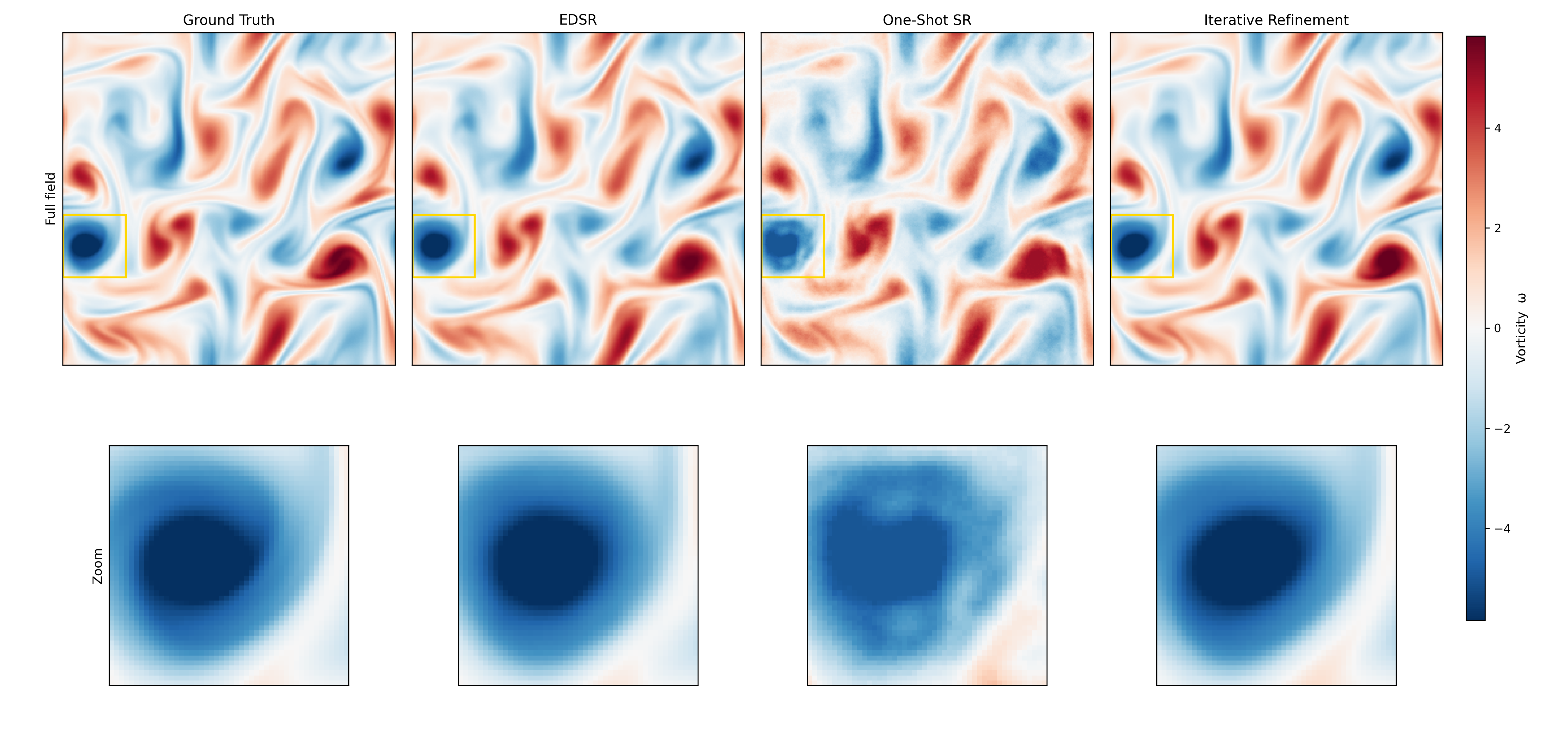}
  \caption{Local zoom comparison on the 2D Kraichnan benchmark for a representative final-time snapshot. The highlighted region shows that iterative refinement best reproduces the shape and intensity of the vortex core, whereas EDSR yields a smoother reconstruction and the one-shot diffusion model introduces visible local distortions. This close-up illustrates the advantage of hierarchical analysis for recovering fine-scale turbulent structure.}
  \label{fig:2d_all_methods_zoom}
\end{figure}

The local zoom comparison in Fig.~\ref{fig:2d_all_methods_zoom} further clarifies this distinction. In the highlighted region, iterative refinement most closely reproduces the shape and intensity of the vortex core, whereas EDSR yields a smoother reconstruction and the one-shot diffusion model introduces visible local distortions. This local comparison illustrates that the improvement is not only a reduction in aggregate RMSE, but also a more accurate recovery of coherent small-scale structure.

Taken together, these comparisons support the central interpretation of the benchmark results. Direct one-shot generative reconstruction can be highly effective when the inverse problem is sufficiently constrained, as in the 1D Burgers case. In the more underdetermined 2D turbulent setting, however, the hierarchical forecast--analysis cascade provides a clear advantage by decomposing the difficult \(32\times32 \rightarrow 256\times256\) recovery problem into a sequence of smaller, better-conditioned refinement steps.

\subsection{Spectrum reconstruction}

To assess whether the reconstructed fields recover the correct multiscale content, we compare the energy spectra of the predicted solutions against the ground truth. This provides a stricter diagnostic than pointwise error alone, since a method may achieve reasonable spatial reconstructions while still misrepresenting the distribution of energy across scales.

\begin{figure}[htbp]
  \centering
  \includegraphics[width=\textwidth]{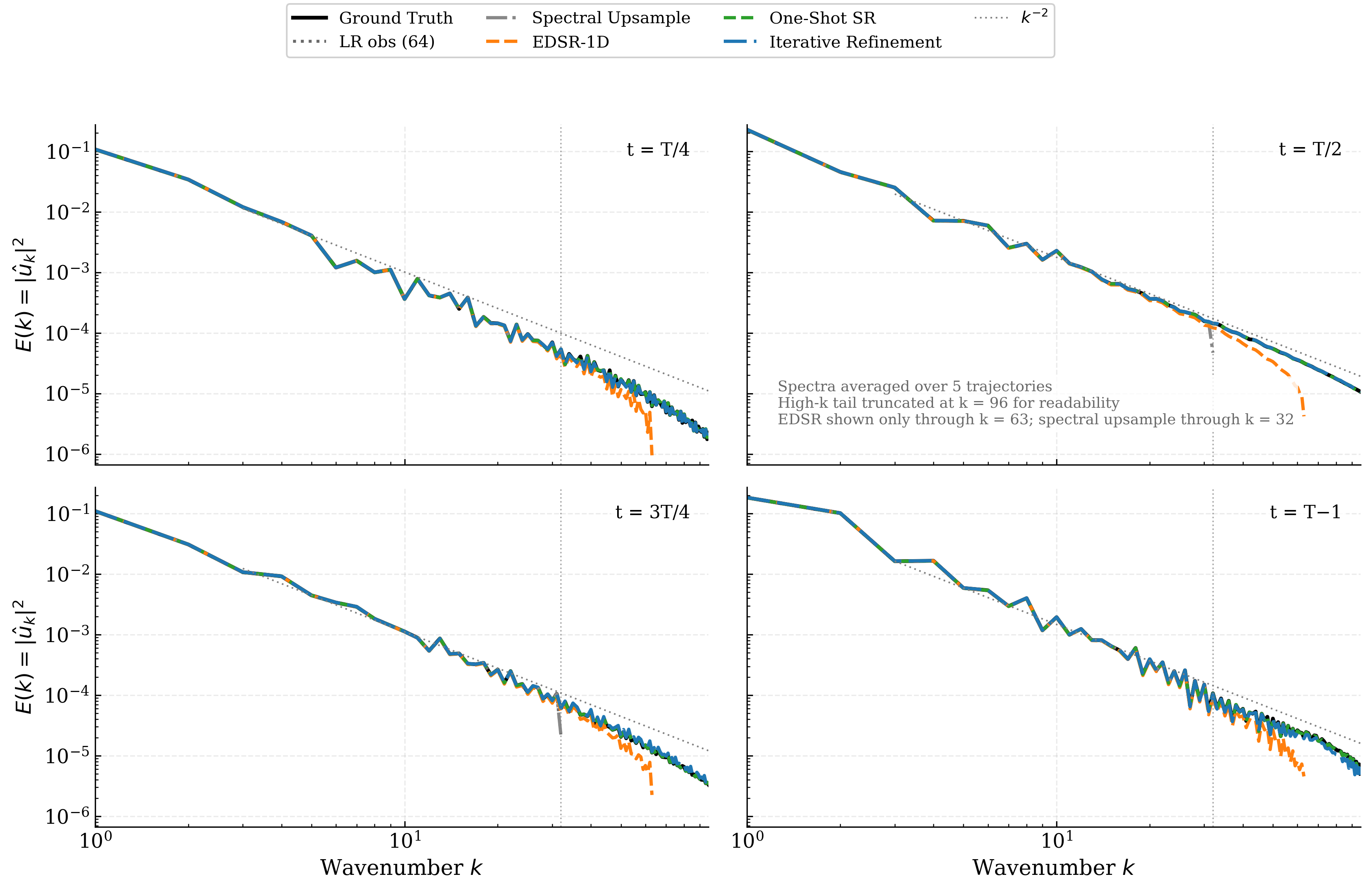}
  \caption{Trajectory-averaged energy spectrum \(E(k)=|\hat{u}_k|^2\) for the 1D stochastic Burgers benchmark at four representative times (\(t=T/4,\,T/2,\,3T/4,\) and \(T\)). spectral upsampling is limited by the coarse observation bandwidth, while EDSR exhibits an earlier high-wavenumber roll-off. The one-shot diffusion and iterative refinement models both closely track the ground-truth spectrum across the resolved wavenumber range, indicating accurate recovery of fine-scale spectral content.}
  \label{fig:burgers_spectrum}
  
\end{figure}

In the 1D Burgers case, Fig.~\ref{fig:burgers_spectrum} shows the trajectory-averaged energy spectrum \(E(k)=|\hat{u}_k|^2\) at four representative times. spectral upsampling is fundamentally limited by the coarse observation bandwidth and therefore cannot recover energy beyond the low-resolution Nyquist cutoff. Among the learned baselines, EDSR exhibits an earlier high-wavenumber roll-off, indicating a tendency toward over-smoothing at the smallest resolved scales. By contrast, the one-shot diffusion and iterative refinement models both track the ground-truth spectrum very closely over nearly the entire resolved range. Their spectral behavior is nearly indistinguishable in this benchmark, which is consistent with the strong overall performance of both models on the corrected 1D dataset. Thus, in 1D Burgers, the main spectral conclusion is that the diffusion-based generative models recover the fine-scale energy content far more faithfully than either spectral upsampling or deterministic EDSR.

\begin{figure}[htbp]
  \centering
  \includegraphics[width=\textwidth]{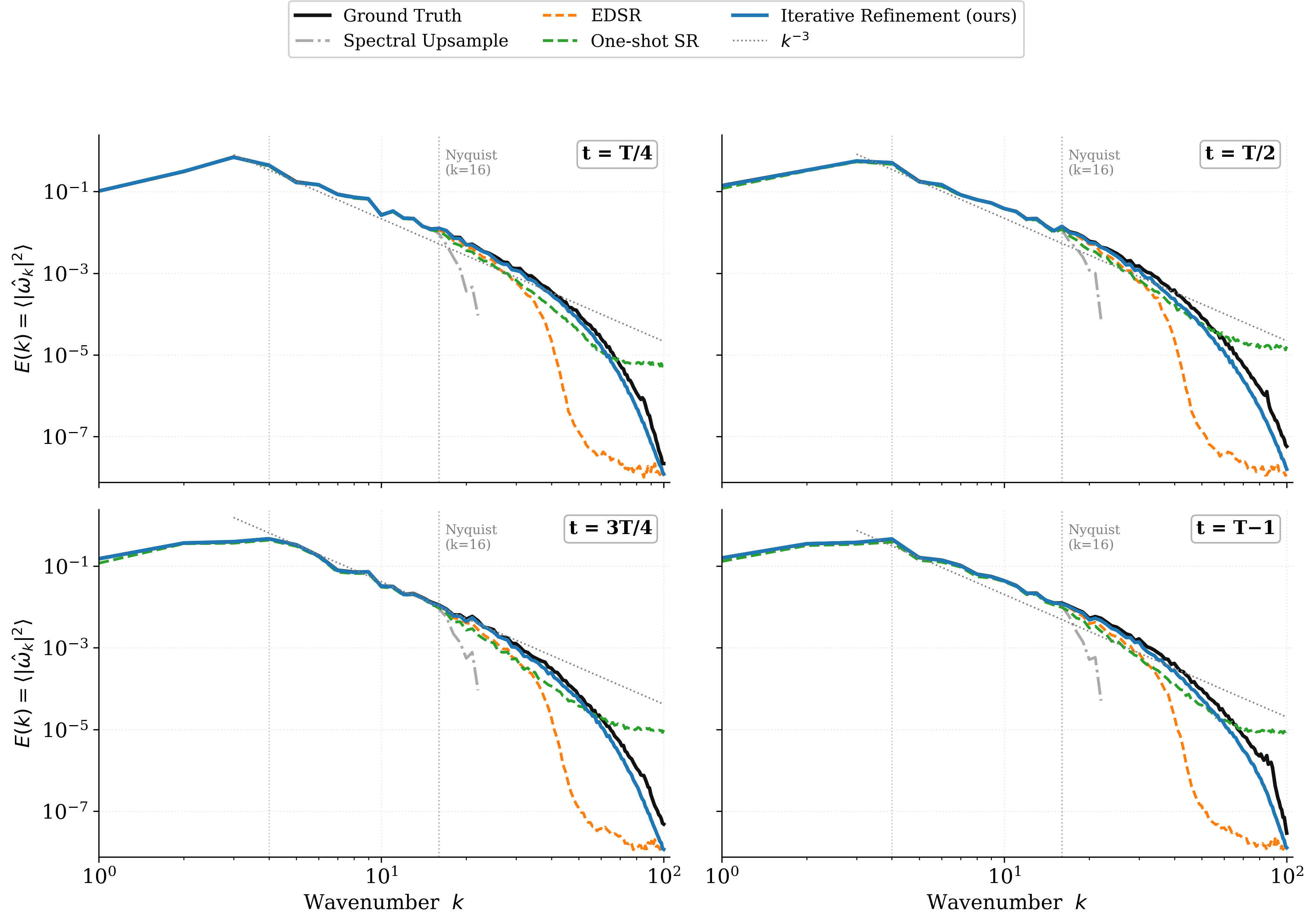}
  \caption{Radial energy spectrum for the 2D Kraichnan turbulence benchmark at 
  resolution \(256\times256\), shown at four evenly spaced trajectory times 
  \(t=T/4\), \(T/2\), \(3T/4\), and \(T-1\), with spectra averaged over the 
  test trajectories. The vertical dotted line marks the coarse-grid Nyquist 
  limit (\(k=16\)). spectral upsampling loses energy rapidly beyond this 
  limit, while EDSR remains overly dissipative at moderate and high 
  wavenumbers. One-shot diffusion improves the spectral reconstruction but 
  still underestimates the high-\(k\) tail. Iterative refinement provides the 
  closest match to the ground-truth spectrum across all four time points, 
  demonstrating more accurate and temporally persistent recovery of fine-scale 
  turbulent structure.}
  \label{fig:kraichnan_spectrum}
\end{figure}

The distinction becomes even clearer in the 2D Kraichnan turbulence case. 
Figure~\ref{fig:kraichnan_spectrum} shows the radial energy spectrum at four 
representative and evenly spaced trajectory times, 
$t=T/4,\; T/2,\; 3T/4,$ and $T-1$, with each spectrum averaged over the test 
trajectories. Across all four time points, spectral upsampling reproduces 
only the low-wavenumber content inherited from the coarse observation and 
rapidly loses energy beyond the coarse-grid Nyquist limit. EDSR improves over 
bicubic at intermediate wavenumbers, but its spectrum decays too aggressively 
in the high-\(k\) range, indicating an overly dissipative reconstruction of 
fine turbulent content. The one-shot diffusion model provides a better spectral 
match than EDSR, but it still underestimates the high-wavenumber tail, 
especially at later times.

By contrast, iterative refinement consistently yields the closest agreement 
with the ground-truth spectrum over the full rollout. In particular, it tracks 
the intermediate-to-high wavenumber range more accurately than the competing 
methods at all four sampled times, indicating superior recovery of the missing 
multiscale cascade. The temporal consistency of this behavior is important: the 
spectral advantage of iterative refinement is not confined to a single instant, 
but persists throughout the sequential reconstruction process. This confirms 
that the hierarchical forecast--analysis strategy is more effective at 
reconstructing fine-scale turbulent structure than either deterministic 
one-shot super-resolution or single-pass stochastic generation.

Taken together, these spectral diagnostics reinforce the main empirical picture from the benchmark results. In the simpler 1D Burgers setting, both diffusion-based approaches recover the correct spectral content very well, with little separation between one-shot and iterative refinement. In the more challenging 2D turbulent setting, however, iterative refinement provides the most faithful spectrum reconstruction, particularly at smaller scales where accurate recovery is most difficult and most important.

\subsection{Temporal Stability of the Iterative Refinement Process}
\label{sec:temporal_stability}

An important requirement for sequential reconstruction is that accuracy remain stable throughout the rollout rather than deteriorate over time. We therefore examine the temporal RMSE of the reconstructed trajectories and compare the frame-to-frame displacement magnitude against the ground-truth evolution. Since the strongest differences between methods occur in the more challenging 2D turbulence setting, we focus the main discussion on the Kraichnan benchmark and provide the corresponding 1D Burgers diagnostics in ~\ref{app:temporal_stability_1d}.

\begin{figure}[htbp]
  \centering
  \begin{subfigure}[t]{0.49\textwidth}
    \centering
    \includegraphics[width=\textwidth]{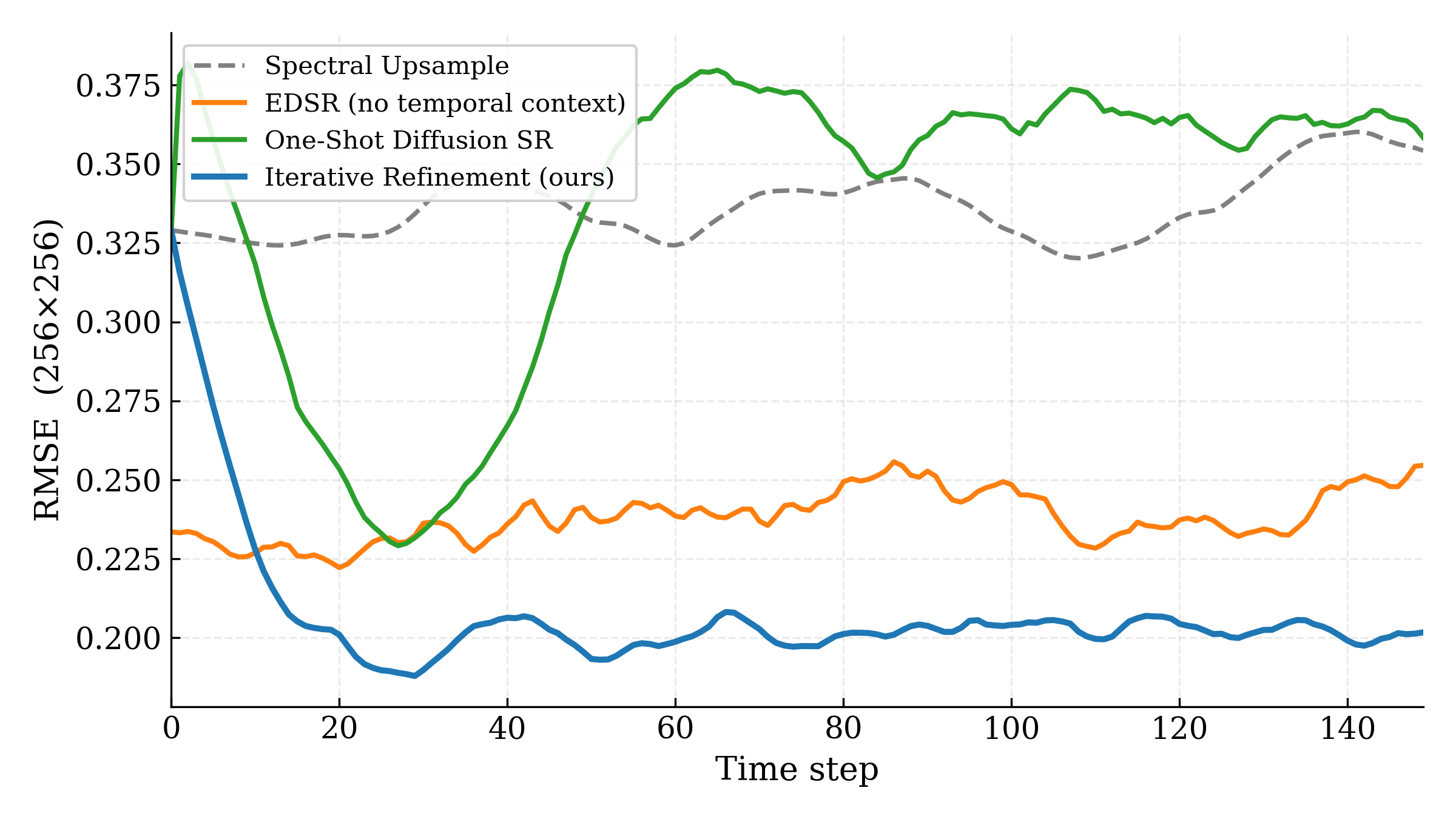}
    \caption{RMSE over time.}
    \label{fig:2d_trmse_comp}
  \end{subfigure}
  \hfill
  \begin{subfigure}[t]{0.49\textwidth}
    \centering
    \includegraphics[width=\textwidth]{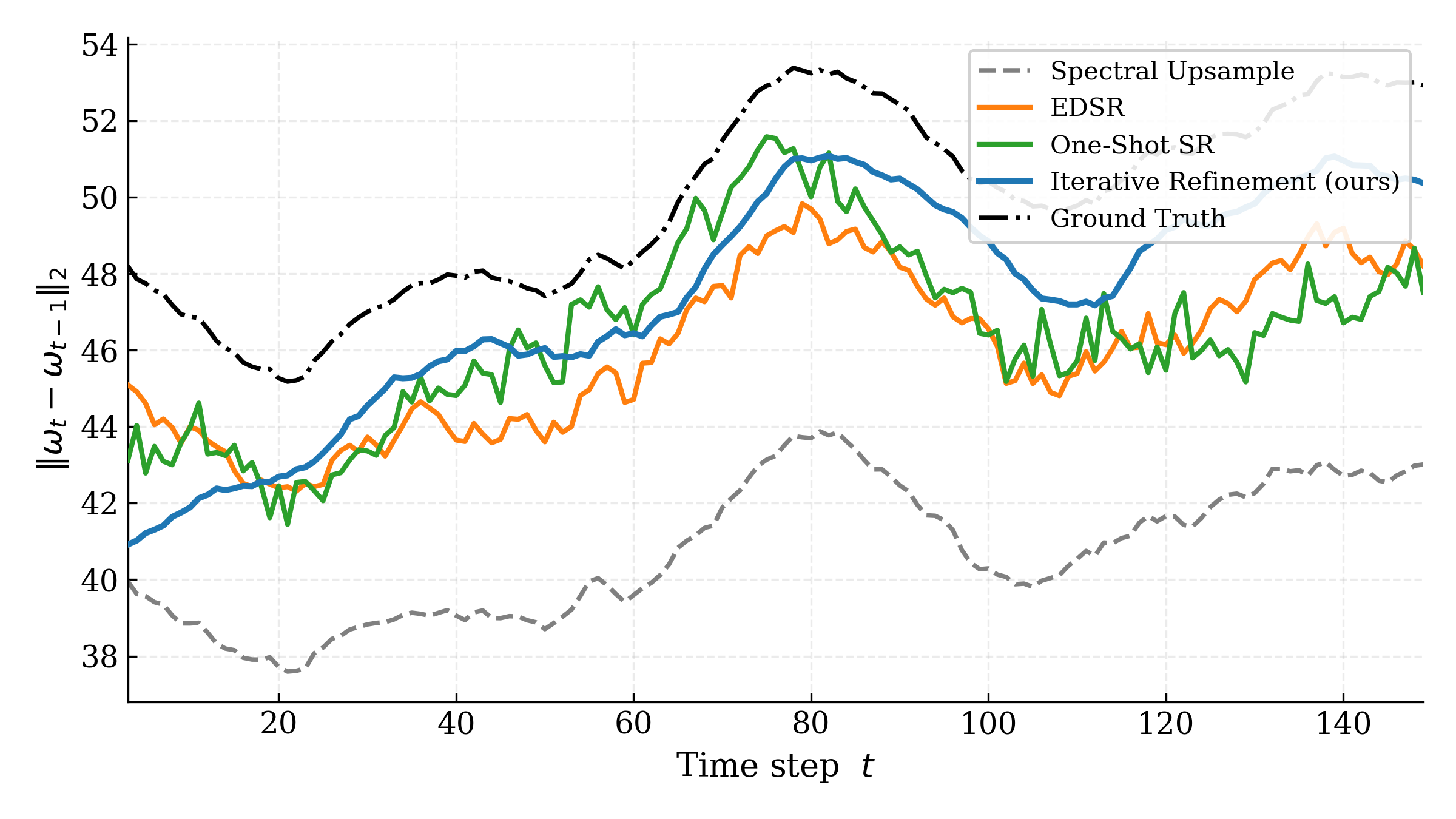}
    \caption{Frame-to-frame temporal consistency.}
    \label{fig:2d_temporal_consistency_comp}
  \end{subfigure}
  \caption{Temporal stability diagnostics on the 2D Kraichnan benchmark at resolution \(256\times256\). \textbf{(a)} RMSE over time. Iterative refinement maintains the lowest and flattest error profile throughout the rollout, outperforming one-shot diffusion, EDSR, spectral upsampling, and the autoregressive FNO-only baseline, which diverges rapidly. \textbf{(b)} Temporal consistency measured by the frame-to-frame displacement norm \(\|\omega_t-\omega_{t-1}\|_2\). Iterative refinement most closely tracks the ground-truth temporal evolution, whereas spectral upsampling is overly smooth and the one-shot and EDSR baselines exhibit larger deviations.}
  \label{fig:2d_temporal_stability}
\end{figure}

Figure~\ref{fig:2d_temporal_stability} shows the temporal stability diagnostics on the 2D Kraichnan benchmark. The RMSE trajectories demonstrate that iterative refinement maintains the lowest and most nearly constant error profile throughout the rollout. EDSR performs better than spectral upsampling and remains far more stable than the autoregressive FNO-only forecaster, but it still exhibits a higher error floor than iterative refinement. The one-shot diffusion model remains bounded, yet its error grows to a visibly larger plateau and stays consistently above the iterative method over most of the rollout. The FNO-only baseline accumulates error rapidly and diverges, confirming that repeated analysis using the current observation is essential for stable long-horizon reconstruction in the turbulent setting.

To assess whether the generated trajectories evolve in a physically plausible manner, Fig.~\ref{fig:2d_temporal_stability} also compares the frame-to-frame displacement norm \(\|\omega_t-\omega_{t-1}\|_2\) against the ground-truth evolution. Spectral upsampling yields the smallest displacement magnitude, indicating overly smooth temporal behavior inherited from the coarse observations. EDSR and one-shot diffusion capture more variability, but both show larger deviations from the ground-truth trend. Iterative refinement provides the closest overall match to the true temporal evolution, reproducing both the scale and trend of the frame-to-frame changes more faithfully.

These diagnostics show that the benefit of iterative refinement is not limited to lower pointwise reconstruction error. In the 2D turbulent setting, the forecast--analysis cascade also produces a more realistic sequential evolution, avoiding the drift of autoregressive forecasting while improving upon both deterministic and one-shot stochastic super-resolution baselines.

\subsection{Comparison with Data-Assimilation Baselines}
\label{sec:da_comparison}

We next examine the data-assimilation interpretation of iterative refinement in more detail on the 2D Kraichnan benchmark. We compare against two ensemble Kalman filtering baselines. The first is a solver-based EnKF that uses the pseudospectral Kraichnan solver as the ensemble forecast model. This serves as a strong classical DA reference because the forecast step has access to the underlying numerical time integrator; we therefore refer to it as a solver EnKF with \emph{oracle dynamics knowledge}. The second is a learned EnKF that replaces the solver forecast with the trained FNO at \(256\times256\), thereby using a learned surrogate as the ensemble forecast model. Both EnKF variants use \(N=20\) ensemble members and assimilate the same \(32\times32\) coarse observations. The observed Fourier modes up to the coarse Nyquist limit are directly constrained by the data, while unresolved modes must be inferred from the forecast dynamics and the update mechanism.

The accuracy--cost comparison in Fig.~\ref{fig:da_cost_accuracy_intro} summarizes the main trade-off. The solver EnKF provides the lowest aggregate RMSE, as expected for a tuned DA method with access to the true high-resolution solver. However, this accuracy requires repeatedly advancing an ensemble of full-resolution states through the pseudospectral model. Iterative refinement performs learned inference without online solver calls. The learned EnKF gives a more direct learned-forecast comparison to the proposed approach, but its reconstruction quality degrades substantially despite its low cost. Since the FNO forecast is deterministic, ensemble spread must be maintained through artificial perturbations rather than arising naturally from stochastic physical dynamics, making the method sensitive to covariance calibration and less effective at recovering fine-scale structure.

\begin{figure}[htbp]
  \centering
  \includegraphics[width=\textwidth]{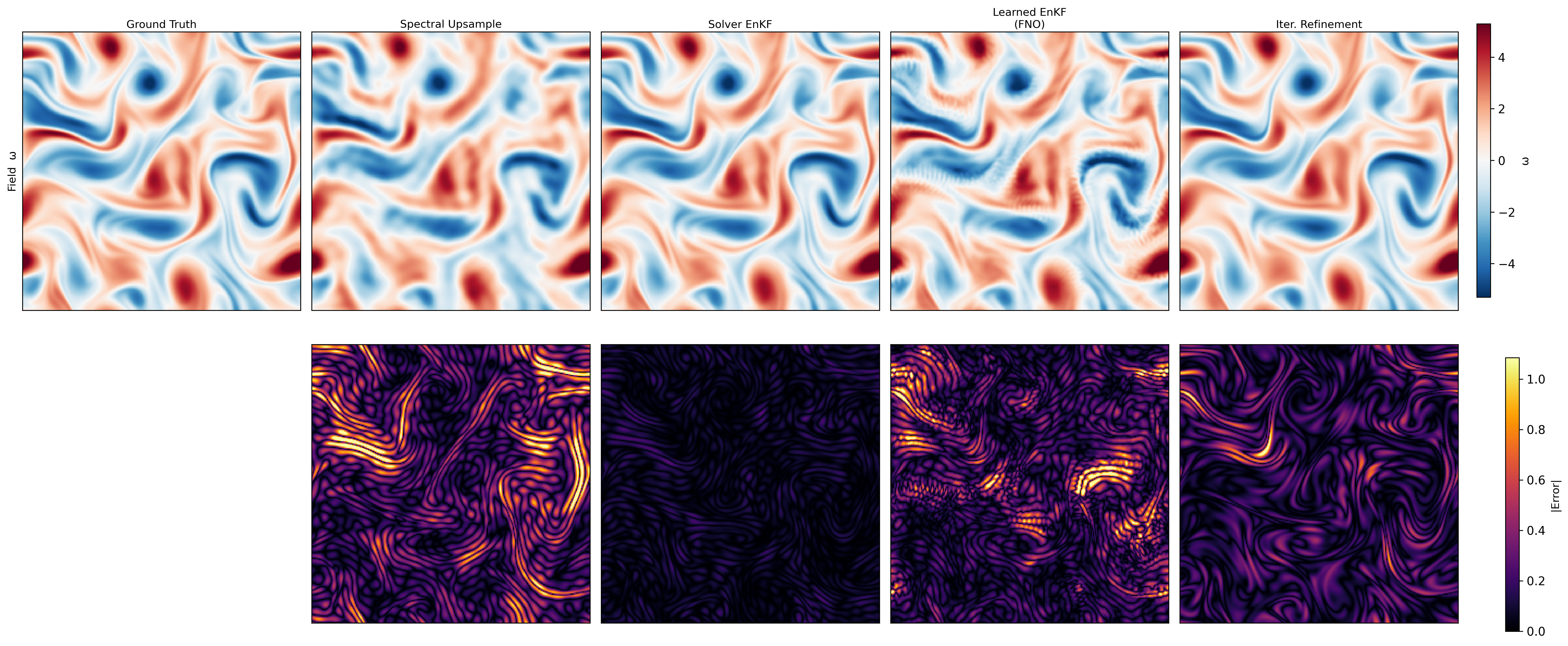}
  \caption{Snapshot comparison between data-assimilation baselines and iterative refinement on the 2D Kraichnan benchmark at resolution \(256\times256\). The top row shows the ground truth, spectral upsampling, solver EnKF, learned EnKF using an FNO forecast, and iterative refinement. The bottom row shows the absolute error for each reconstruction method. The solver EnKF and iterative refinement both recover the main vortex and filament structures well, while the learned EnKF exhibits larger localized artifacts and spectral upsampling misses substantial fine-scale content.}
  \label{fig:da_snapshot}
\end{figure}

The qualitative comparison in Fig.~\ref{fig:da_snapshot} illustrates these differences. The solver EnKF and iterative refinement both recover the dominant vortex and filament structures well, whereas the learned EnKF exhibits larger localized artifacts and less accurate fine-scale organization. Spectral upsampling, as expected, misses much of the unresolved turbulent content. Iterative refinement produces a smoother and more coherent posterior than the learned EnKF, indicating that the nonlinear diffusion-based analysis step is more effective than a linear ensemble update built on the same learned forecast family.

\begin{figure}[htbp]
  \centering
  \includegraphics[width=\textwidth]{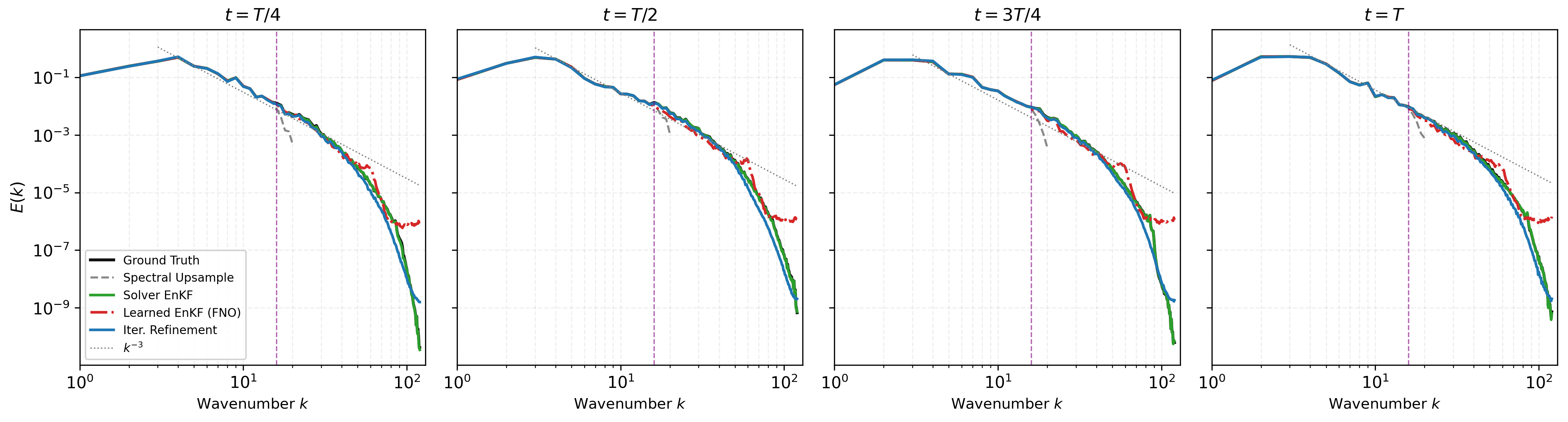}
  \caption{Radial energy spectrum comparison between EnKF variants and iterative refinement on the 2D Kraichnan benchmark. Spectra are shown at four representative times and averaged over test trajectories. The solver EnKF, which uses the pseudospectral solver as the ensemble forecast model, most closely follows the ground-truth spectrum. Iterative refinement also recovers the spectral decay well without requiring online solver access. The learned EnKF, which replaces the solver with an FNO forecast, is less reliable at high wavenumbers and may introduce nonphysical spectral content.}
  \label{fig:da_spectrum}
\end{figure}

The spectral comparison in Fig.~\ref{fig:da_spectrum} further clarifies the distinction. The solver EnKF and iterative refinement both track the ground-truth radial energy spectrum well across much of the resolved range. The learned EnKF can preserve or inject energy beyond the observation cutoff, but this energy is not distributed as reliably and may appear as nonphysical high-wavenumber content. Iterative refinement more consistently follows the ground-truth spectral decay, showing that the diffusion corrector provides a better learned mechanism for fine-scale recovery than a linear EnKF update based on the same FNO forecast model.

Overall, this comparison strengthens the interpretation of iterative refinement as a genuinely data-assimilative method rather than only a super-resolution model. Iterative refinement does not outperform a solver EnKF with oracle dynamics knowledge when the true high-resolution solver is available online. Instead, its value is as a learned super-resolved assimilation method that substantially improves over a learned EnKF using the same forecast family while avoiding repeated high-resolution ensemble forecasts.

\section{Ablation Studies}
\label{sec:ablations}

We now examine the main structural design choices in the iterative refinement framework. The quantitative ablations focus on two questions: whether intermediate refinement stages improve reconstruction quality, and whether propagating corrected posteriors between stages is necessary for accurate coarse-to-fine recovery. Additional prior--posterior visualizations, provided in ~\ref{app:forecast_prior_diagnostics}, show that the shared FNO forecast already supplies a meaningful dynamical prior and that the diffusion corrector primarily performs localized residual analysis rather than re-synthesizing the full state from scratch.

\subsection{Effect of Cascade Depth}

 The goal of this ablation is to determine whether the intermediate refinement stages are genuinely useful, or whether similar performance could be achieved with a shallower inference pipeline. To isolate the effect of cascade depth, all variants reuse the same trained shared FNO forecaster and the same shared diffusion corrector; only the inference-time sequence of refinement stages is modified.

For the 1D Burgers benchmark, we compare
\begin{align}
\text{1-stage:} \quad &64 \rightarrow 512, \\
\text{2-stage:} \quad &64 \rightarrow 256 \rightarrow 512, \\
\text{3-stage:} \quad &64 \rightarrow 128 \rightarrow 256 \rightarrow 512,
\end{align}
where the 3-stage variant is the full iterative refinement pipeline used in the main experiments.

For the 2D Kraichnan benchmark, we compare
\begin{align}
\text{1-stage:} \quad &32 \times 32 \rightarrow 256 \times 256, \\
\text{2-stage:} \quad &32 \times 32 \rightarrow 128 \times 128 \rightarrow 256 \times 256, \\
\text{3-stage:} \quad &32 \times 32 \rightarrow 64 \times 64 \rightarrow 128 \times 128 \rightarrow 256 \times 256,
\end{align}
with the 3-stage variant again corresponding to the default full cascade.

\begin{figure}[htbp]
  \centering

  \begin{subfigure}[t]{0.48\textwidth}
    \centering
    \includegraphics[width=\textwidth]{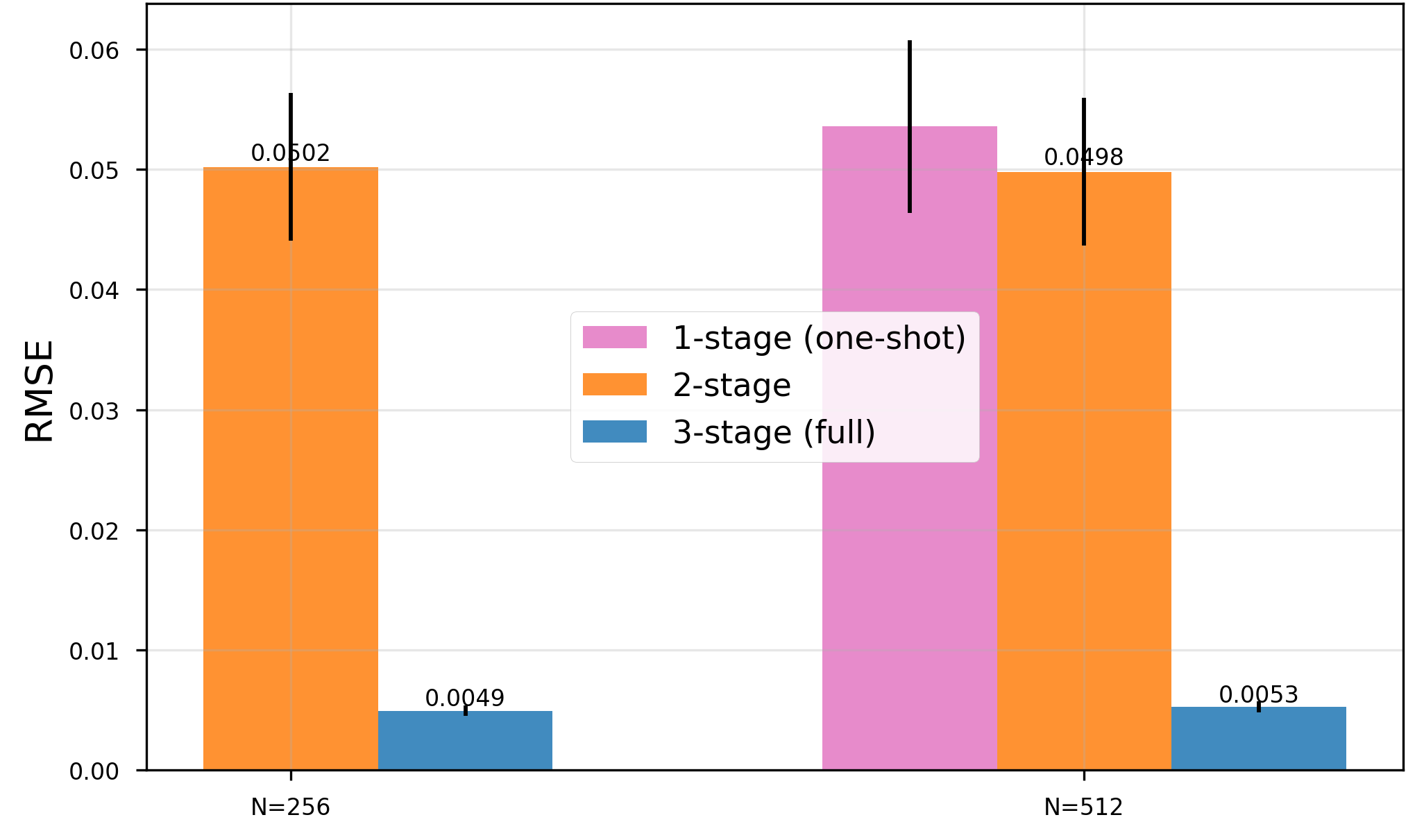}
    \caption{Aggregate RMSE at $N=256$ and $N=512$.}
    \label{fig:a1_rmse_bars}
  \end{subfigure}
  \hfill
  \begin{subfigure}[t]{0.48\textwidth}
    \centering
    \includegraphics[width=\textwidth]{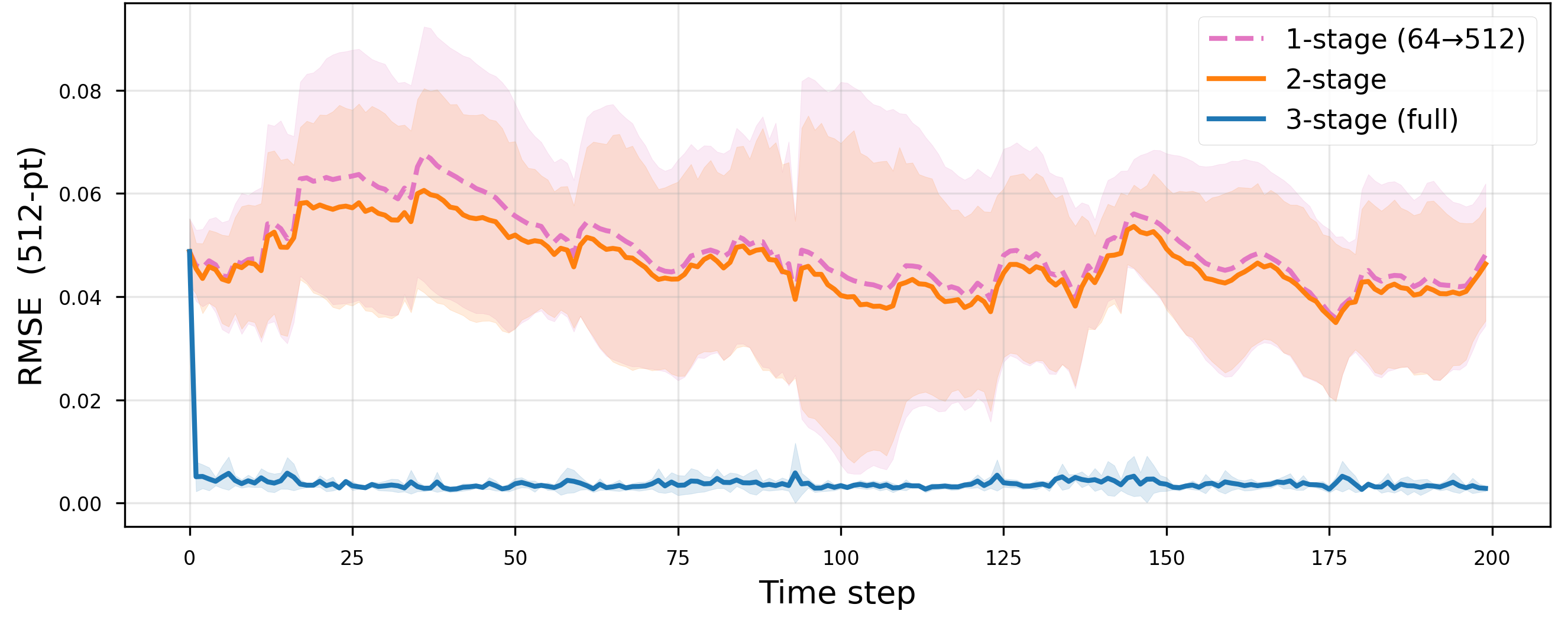}
    \caption{RMSE over time at $N=512$.}
    \label{fig:a1_rmse_time}
  \end{subfigure}

  \vspace{0.8em}

  \begin{subfigure}[t]{0.78\textwidth}
    \centering
    \includegraphics[width=\textwidth]{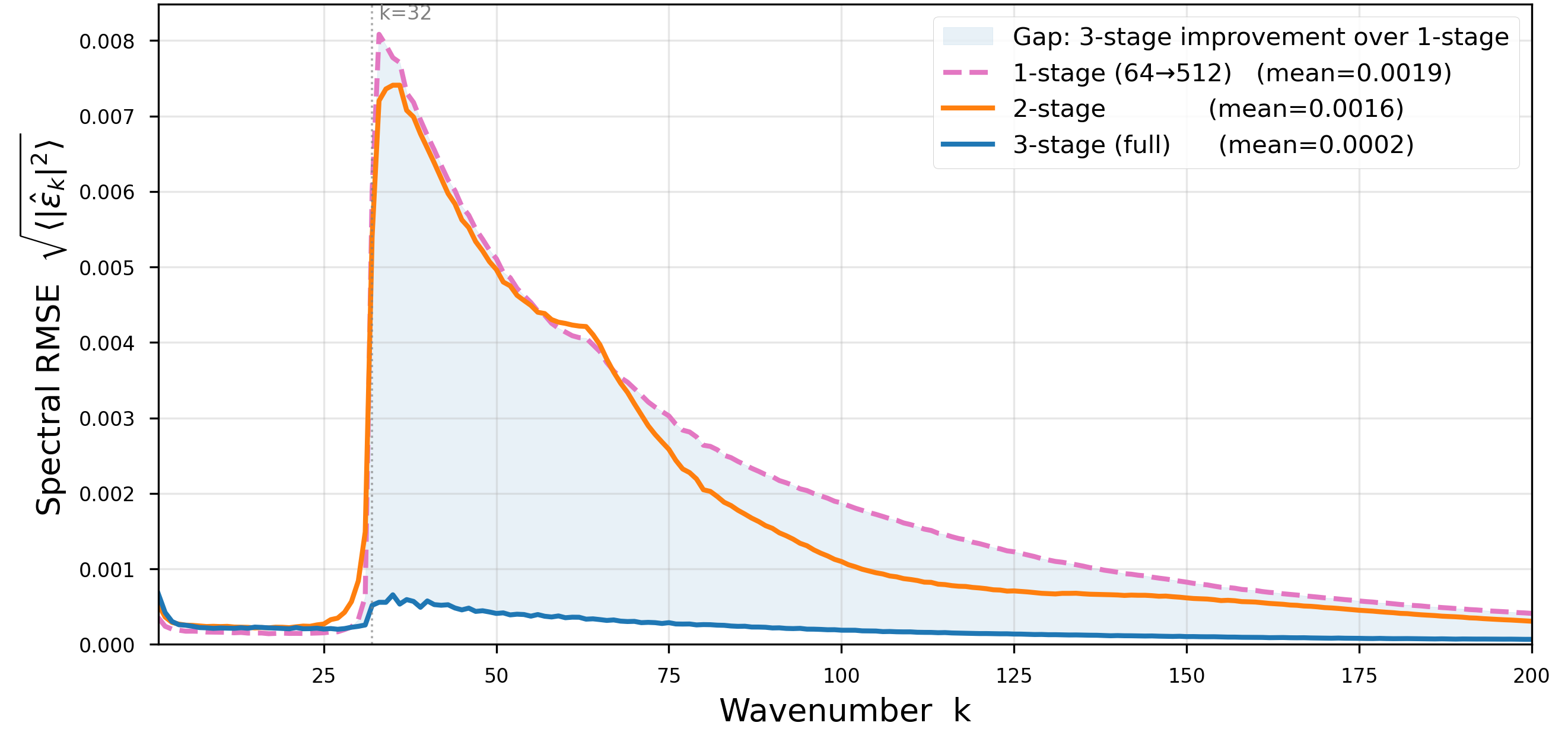}
    \caption{Per-mode spectral RMSE at $N=512$.}
    \label{fig:a1_spectral_rmse}
  \end{subfigure}

  \caption{Cascade-depth ablation on the 1D Burgers benchmark. All variants use the same trained forecasters and shared diffusion corrector; only the inference cascade is changed. The full 3-stage cascade, $64\rightarrow128\rightarrow256\rightarrow512$, achieves the lowest aggregate RMSE, maintains the most stable temporal error profile, and yields the smallest spectral error beyond the coarse-observation Nyquist limit.}
  \label{fig:ablation_cascade_depth_1d}
\end{figure}

The results show that cascade depth has a strong and consistent effect on performance in both testbeds. On the 1D Burgers problem (Fig.~\ref{fig:ablation_cascade_depth_1d}), the full 3-stage cascade substantially outperforms the shallower alternatives in aggregate RMSE, temporal stability, and spectral reconstruction. The 1-stage and 2-stage variants maintain significantly larger error floors throughout the rollout, whereas the full cascade rapidly settles into a low-error regime. The spectral error plot further shows that the benefit is concentrated in the high-wavenumber range beyond the coarse-observation Nyquist limit, indicating that the intermediate refinement stages are critical for recovering unresolved fine scales.

\begin{figure}[htbp]
  \centering

  \begin{subfigure}[t]{0.48\textwidth}
    \centering
    \includegraphics[width=\textwidth]{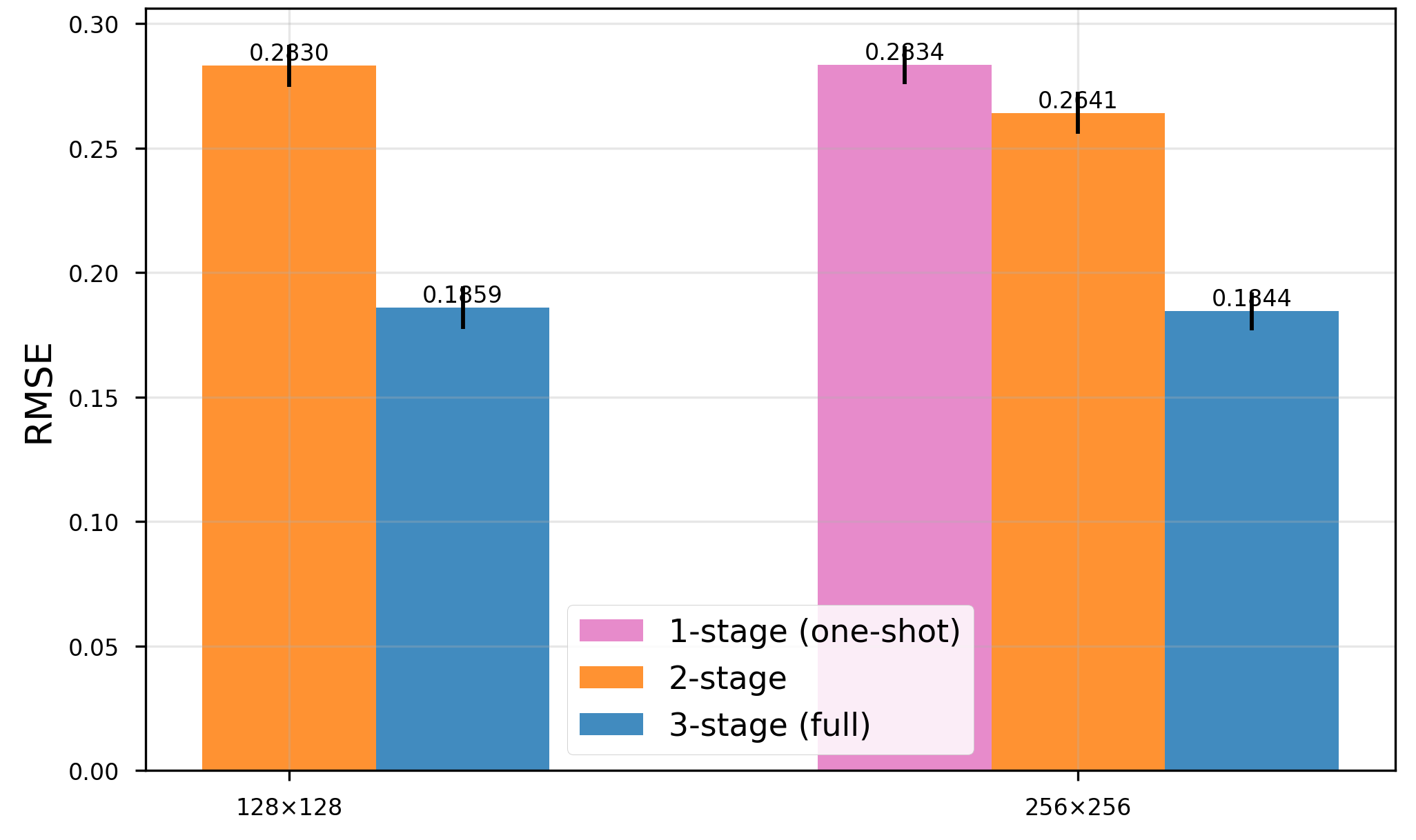}
    \caption{Aggregate RMSE at $128\times128$ and $256\times256$.}
    \label{fig:a1_2d_rmse_bars}
  \end{subfigure}
  \hfill
  \begin{subfigure}[t]{0.48\textwidth}
    \centering
    \includegraphics[width=\textwidth]{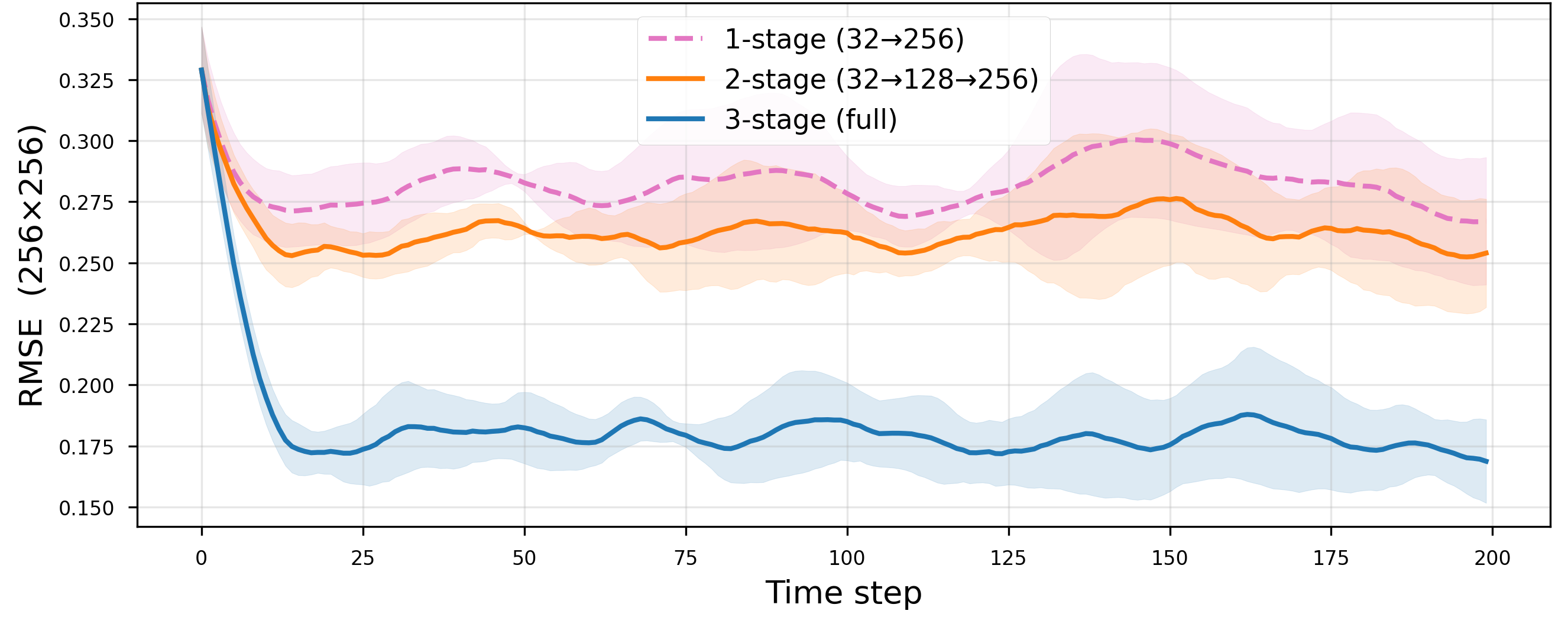}
    \caption{RMSE over time at $256\times256$.}
    \label{fig:a1_2d_rmse_time}
  \end{subfigure}

  \vspace{0.8em}

  \begin{subfigure}[t]{0.78\textwidth}
    \centering
    \includegraphics[width=\textwidth]{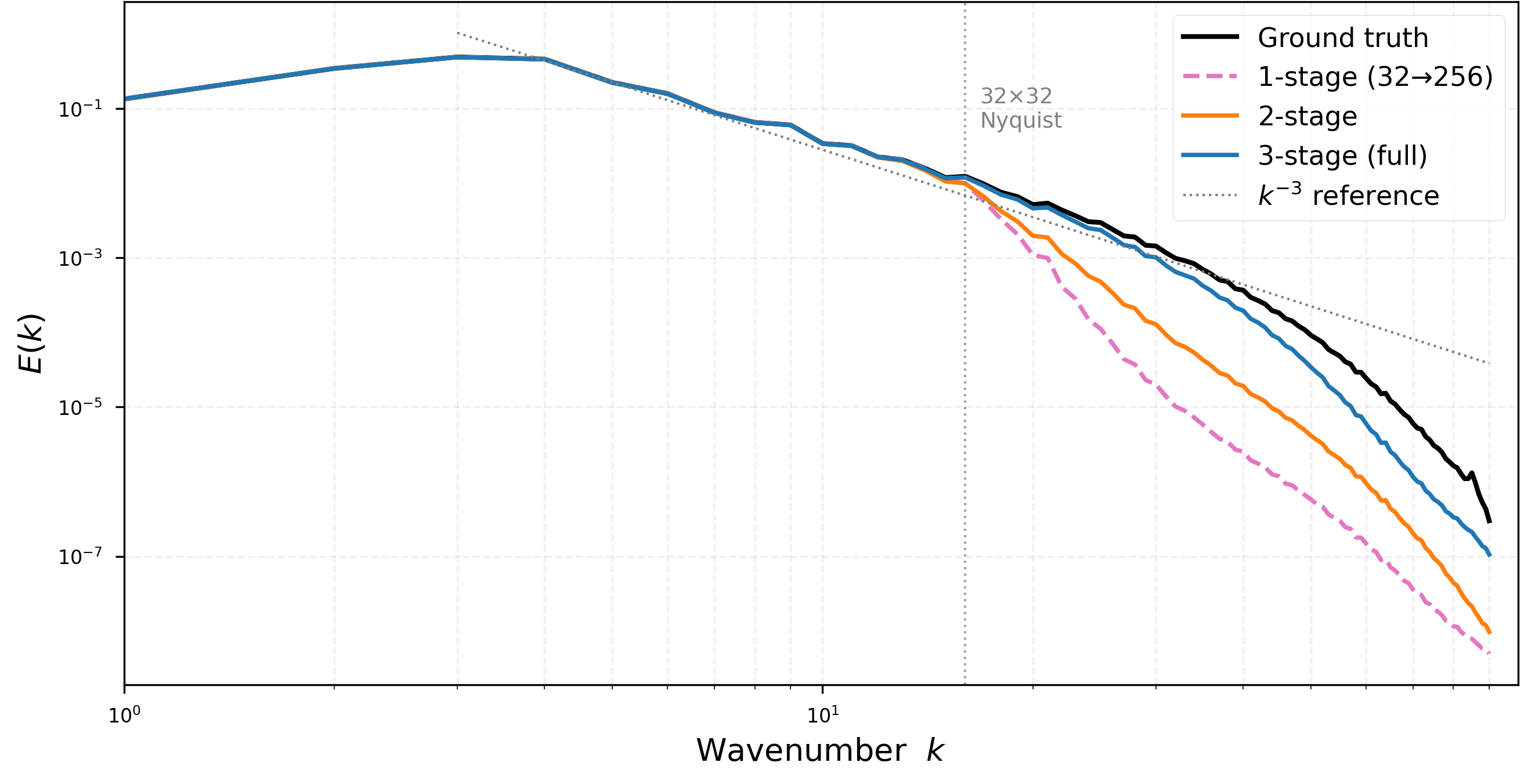}
    \caption{Radial energy spectrum at $256\times256$.}
    \label{fig:a1_2d_spectrum}
  \end{subfigure}

  \caption{Cascade-depth ablation on the 2D Kraichnan turbulence benchmark. The one-shot 1-stage variant attempts a direct $32\times32 \rightarrow 256\times256$ reconstruction, the 2-stage variant inserts an intermediate $128\times128$ refinement, and the full 3-stage cascade further includes a $64\times64$ stage. The deeper cascade consistently improves aggregate accuracy, lowers the temporal error floor, and most faithfully reconstructs the high-wavenumber portion of the energy spectrum beyond the coarse-grid Nyquist limit.}
  \label{fig:ablation_cascade_depth_2d}
\end{figure}

The same trend persists, and in fact becomes even clearer, in the 2D Kraichnan case (Fig.~\ref{fig:ablation_cascade_depth_2d}). At the finest resolution of $256\times256$, the one-shot 1-stage variant attains an RMSE of approximately $0.2834$, the 2-stage variant improves this to $0.2641$, and the full 3-stage cascade reduces it further to $0.1844$. At the intermediate resolution of $128\times128$, the 3-stage cascade also clearly outperforms the 2-stage alternative ($0.1859$ versus $0.2830$), showing that the benefit of a deeper hierarchy appears already before the final super-resolution step. The RMSE-over-time curves confirm that this gain is persistent over the full rollout rather than concentrated at isolated snapshots.

The 2D spectral comparison provides the clearest explanation. All variants closely match the ground-truth spectrum in the observed low-wavenumber range, but they diverge substantially beyond the $32\times32$ Nyquist limit. The 1-stage variant loses high-wavenumber energy most rapidly, the 2-stage variant improves the reconstruction but still remains too dissipative, and the full 3-stage cascade preserves the closest match to the ground-truth $E(k)$ curve over the intermediate and high-wavenumber range. Thus, in both 1D and 2D, deeper cascades yield better-conditioned correction problems and significantly improve the recovery of fine-scale structure.

Overall, this ablation supports the central design principle of the proposed method: rather than attempting to bridge a large coarse-to-fine resolution gap in one step, it is more effective to decompose the task into a sequence of smaller forecast--analysis refinements. The advantage of this staged reconstruction is modest but consistent in 1D, and becomes substantially more pronounced in the more underconstrained 2D turbulence setting.

\subsection{Effect of the Propagation Signal}

\begin{figure}[htbp]
    \centering
    \includegraphics[width=0.9\linewidth]{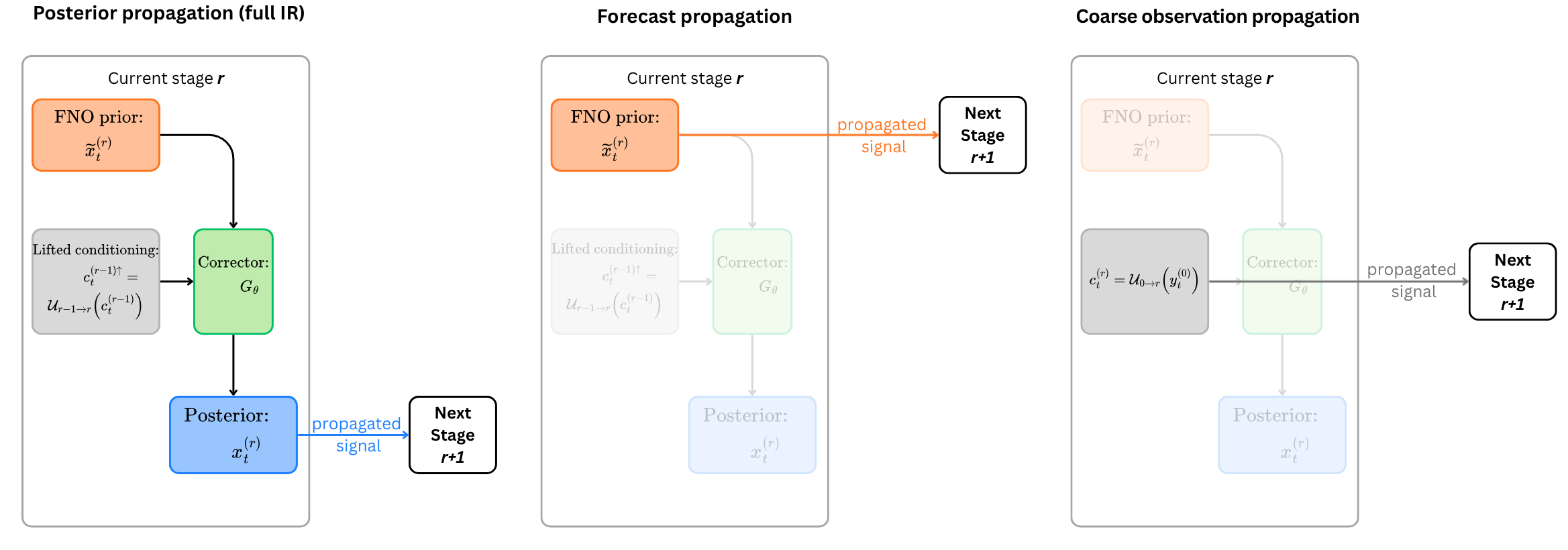}
    \caption{Schematic of the three propagation-signal variants considered in the ablation study. After each refinement stage, the next finer stage is conditioned on either (i) the corrected posterior from the previous stage, (ii) the FNO forecast prior at that stage, or (iii) the original coarse observation upsampled to the required intermediate resolution. The default iterative refinement method corresponds to posterior propagation.}
    \label{fig:prop_signal}
\end{figure}

We next study which signal should be propagated between refinement stages. In the default iterative refinement pipeline, the corrected posterior produced at one stage becomes the coarse conditioning input for the next finer stage. This choice assumes that each stage produces useful multiscale information that should be preserved and passed upward through the hierarchy. To test this assumption, we compare three propagation choices while keeping the trained FNO forecasters and diffusion corrector fixed:
\begin{align}
\text{(i) posterior:} \quad & c_t^{(r)} = x_t^{(r)}, \\
\text{(ii) forecast:} \quad & c_t^{(r)} = \widetilde{x}_t^{(r)}, \\
\text{(iii) raw observation:} \quad & c_t^{(r)} = \mathcal{U}_{0\to r}\!\left(y_t\right).
\end{align}
Here, the first option is the default method. The second replaces the corrected posterior with the FNO forecast at the same resolution, and the third ignores intermediate refined states entirely, instead repeatedly reusing the original coarse observation after upsampling it to the required resolution. The first refinement stage is identical across all variants, since it always conditions directly on the coarsest observation; only the propagated signal for later stages is changed.

\begin{figure}[htbp]
  \centering

  \begin{subfigure}[t]{0.48\textwidth}
    \centering
    \includegraphics[width=\textwidth]{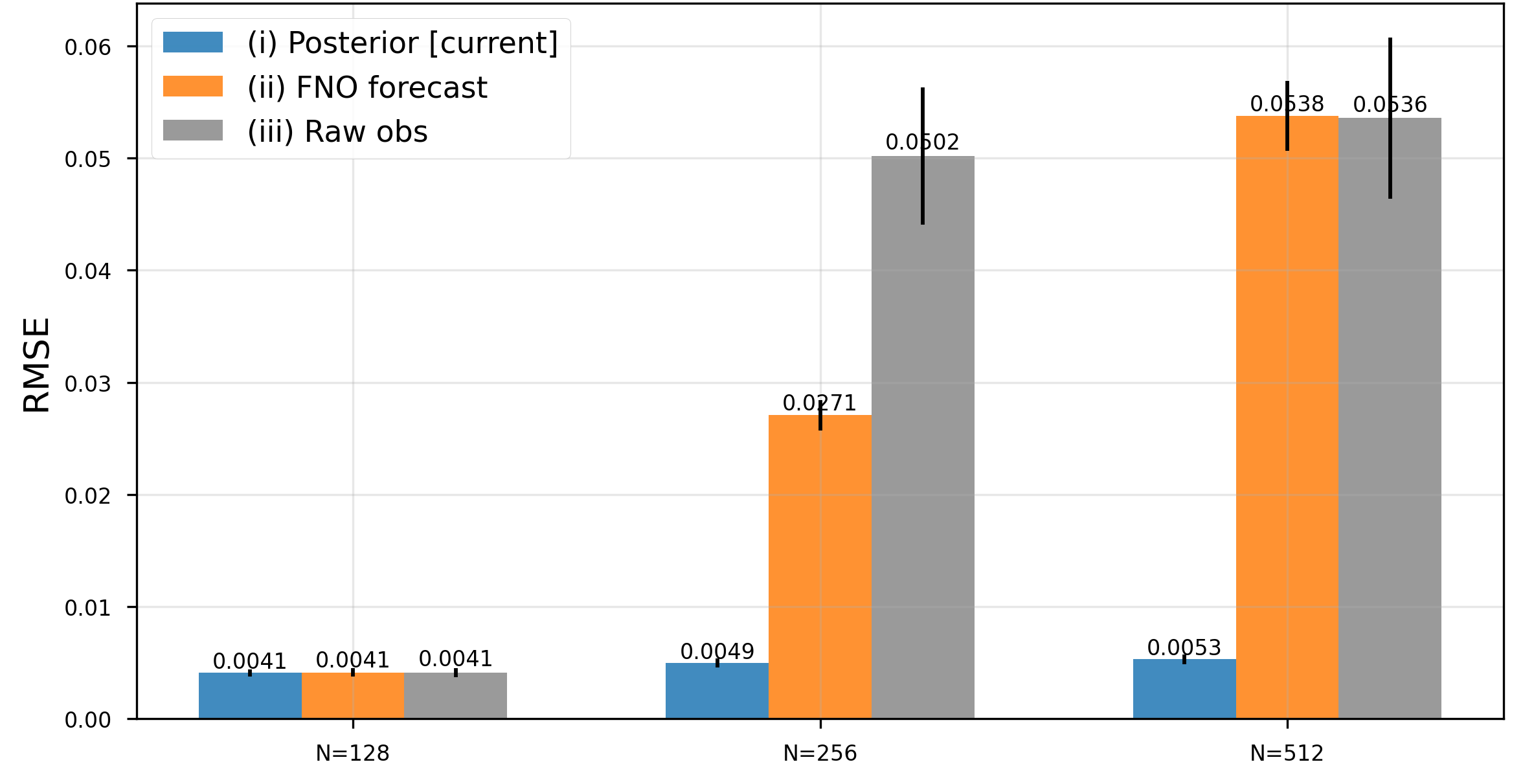}
    \caption{Aggregate RMSE at each resolution.}
    \label{fig:a2_rmse_bars_1d}
  \end{subfigure}
  \hfill
  \begin{subfigure}[t]{0.48\textwidth}
    \centering
    \includegraphics[width=\textwidth]{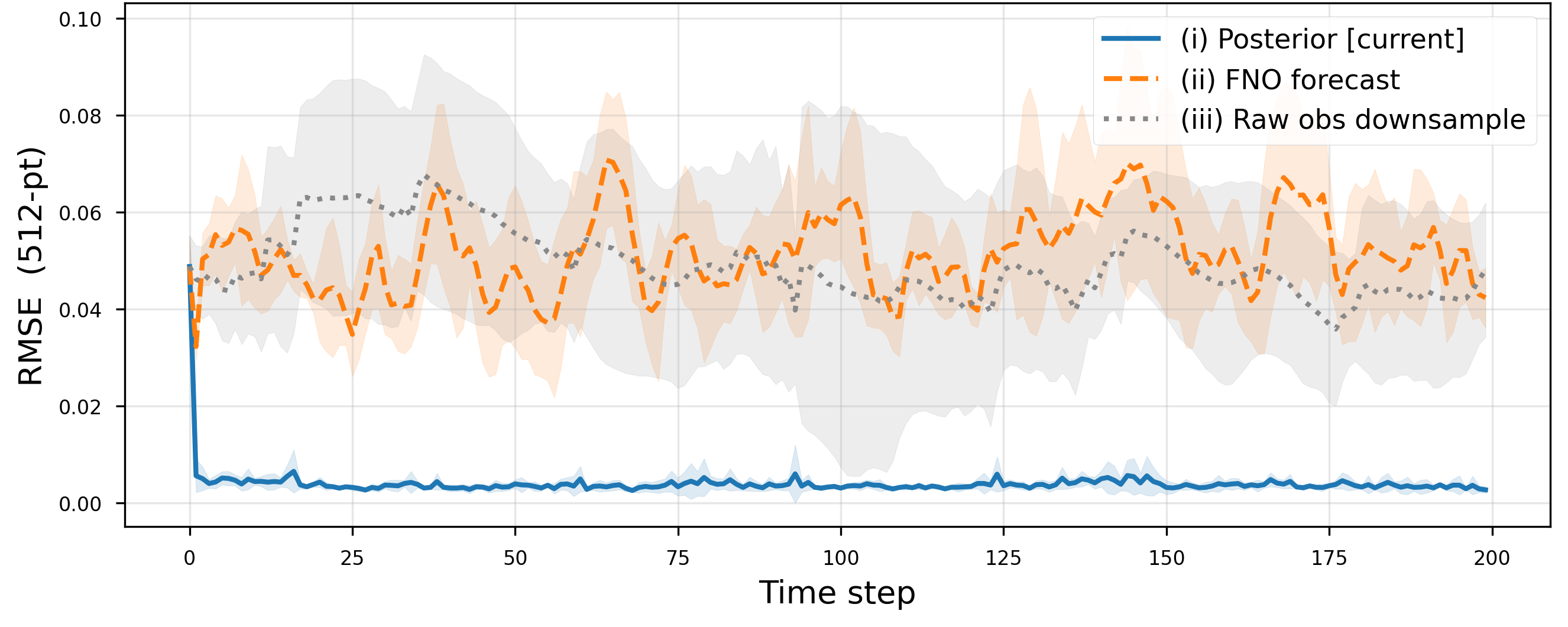}
    \caption{RMSE over time at \(N=512\).}
    \label{fig:a2_rmse_time_1d}
  \end{subfigure}

  \vspace{0.8em}

  \begin{subfigure}[t]{0.78\textwidth}
    \centering
    \includegraphics[width=\textwidth]{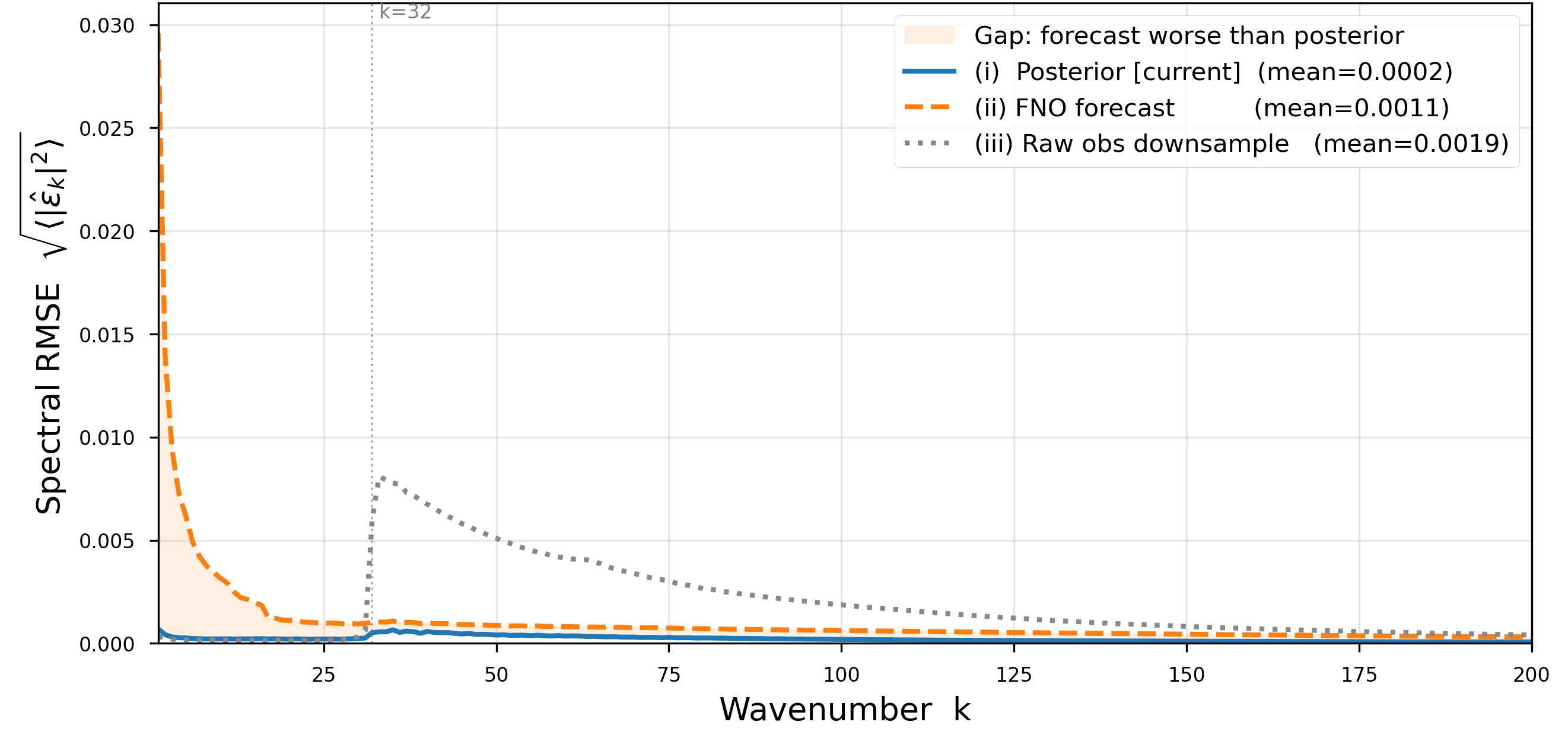}
    \caption{Per-mode spectral RMSE at \(N=512\).}
    \label{fig:a2_spectral_rmse_1d}
  \end{subfigure}

  \caption{Propagation-signal ablation on the 1D Burgers benchmark. All variants use the same trained FNO forecasters and diffusion corrector; only the signal passed from one refinement stage to the next is changed. Propagating the corrected posterior yields the lowest aggregate RMSE, the most stable temporal behavior, and the smallest spectral error. Using the FNO forecast or repeatedly reusing the upsampled coarse observation removes the benefit of progressive posterior refinement and substantially degrades the final high-resolution reconstruction.}
  \label{fig:ablation_propagation_signal_1d}
\end{figure}

The 1D Burgers results are shown in Fig.~\ref{fig:ablation_propagation_signal_1d}. At the first refined level, all three variants perform similarly because they all receive the same initial coarse observation. However, the differences become pronounced at finer resolutions. At \(N=256\), posterior propagation reduces the RMSE to approximately \(0.0049\), compared with \(0.0271\) for forecast propagation and \(0.0502\) for raw-observation propagation. At \(N=512\), the posterior variant remains near \(0.0053\), whereas the forecast and raw-observation variants stay around \(0.0538\) and \(0.0536\), respectively. The RMSE-over-time curves show that posterior propagation rapidly settles into a low-error regime and remains stable, while the other two variants maintain a persistent high-error floor. The spectral RMSE confirms that the corrected posterior carries the information needed for fine-scale recovery: using the posterior yields the lowest error across nearly all Fourier modes.

\begin{figure}[htbp]
  \centering

  \begin{subfigure}[t]{0.48\textwidth}
    \centering
    \includegraphics[width=\textwidth]{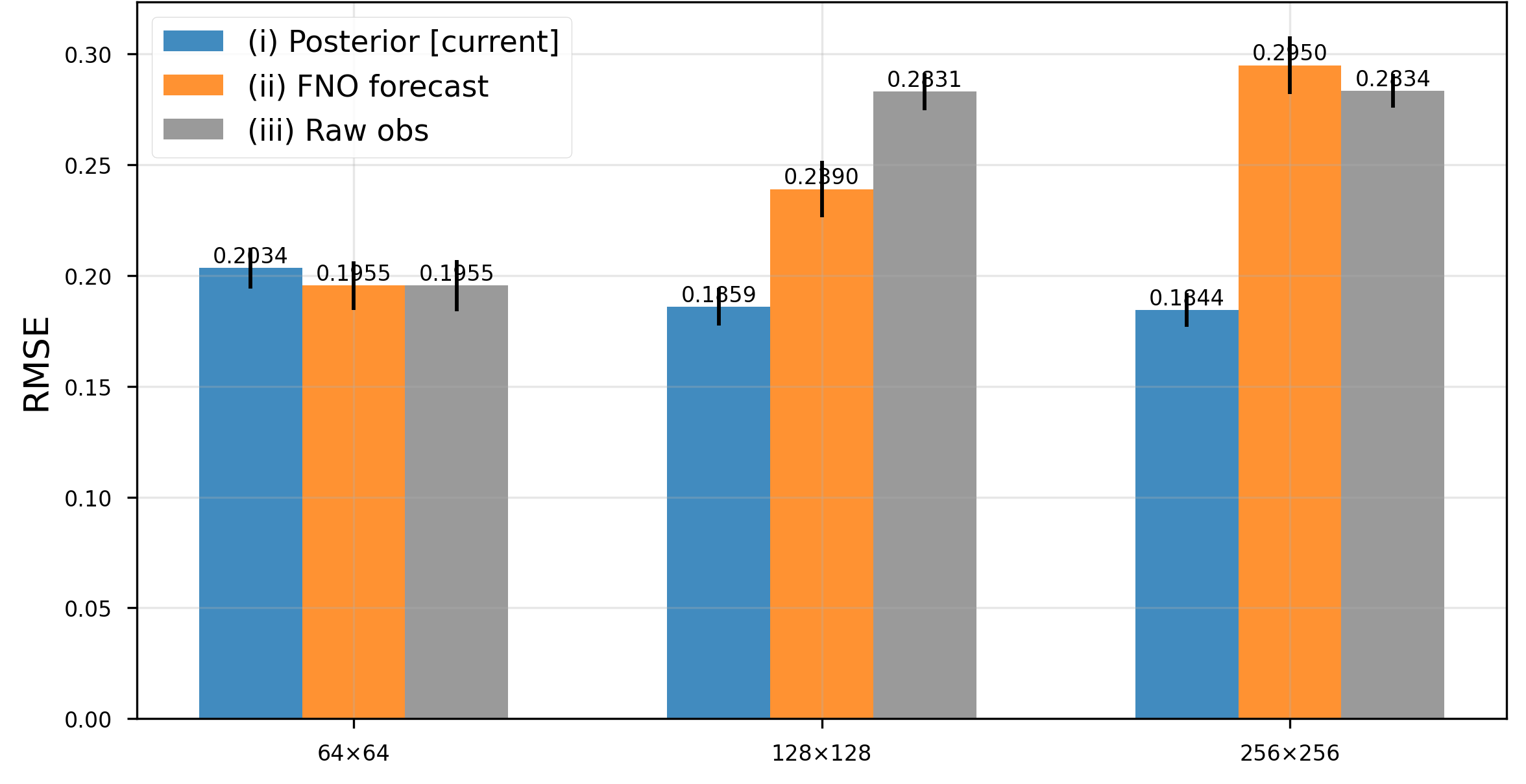}
    \caption{Aggregate RMSE at each resolution.}
    \label{fig:a2_rmse_bars_2d}
  \end{subfigure}
  \hfill
  \begin{subfigure}[t]{0.48\textwidth}
    \centering
    \includegraphics[width=\textwidth]{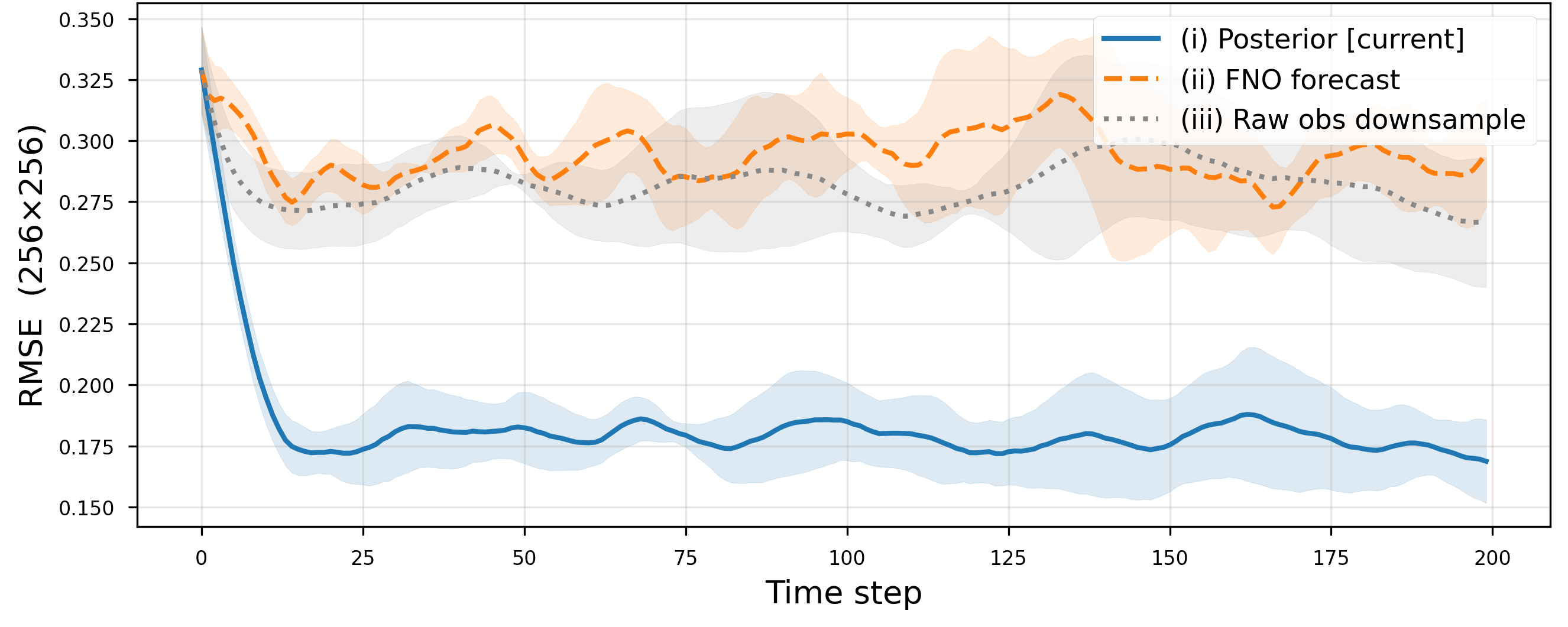}
    \caption{RMSE over time at \(256\times256\).}
    \label{fig:a2_rmse_time_2d}
  \end{subfigure}

  \vspace{0.8em}

  \begin{subfigure}[t]{0.78\textwidth}
    \centering
    \includegraphics[width=\textwidth]{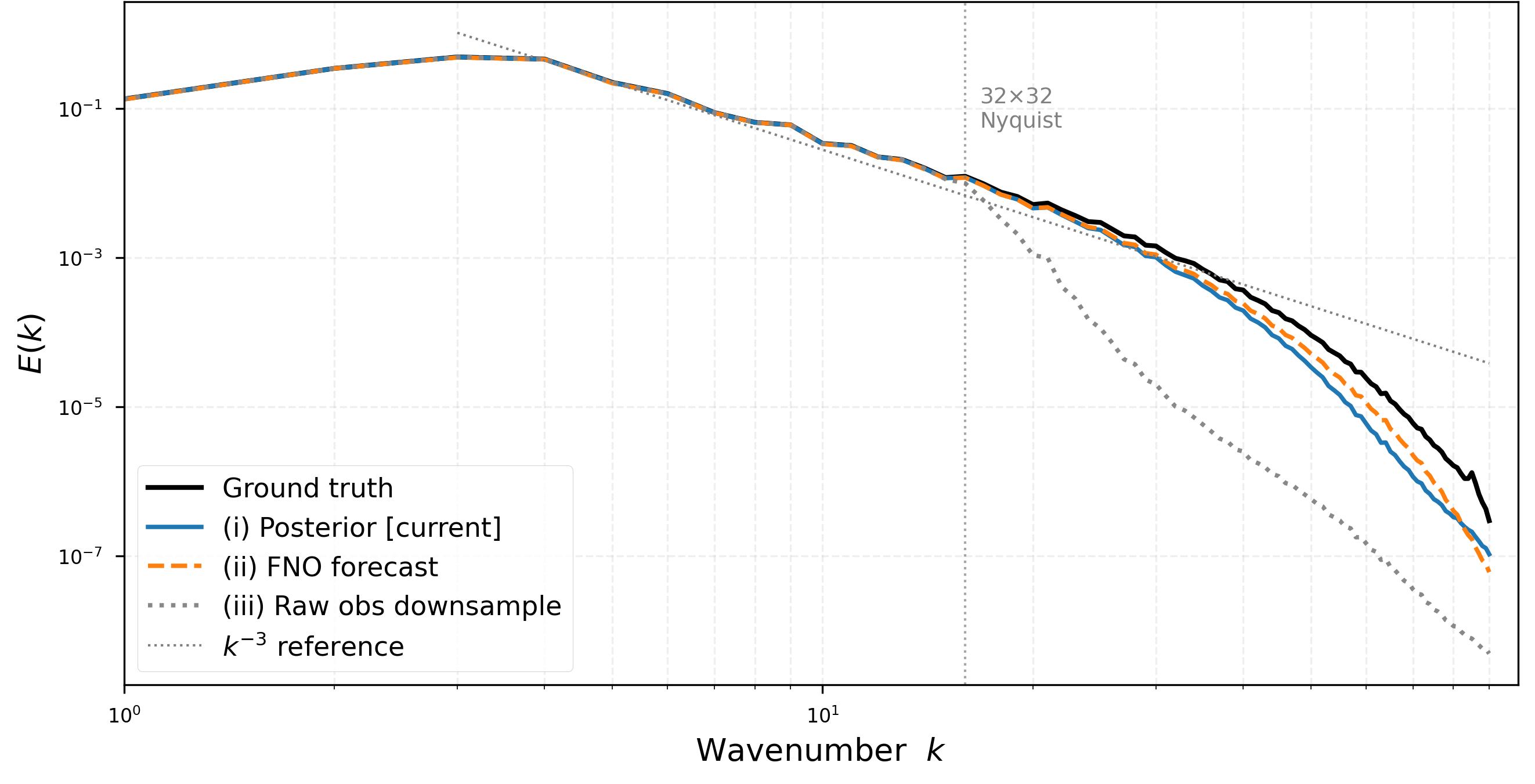}
    \caption{Radial energy spectrum at \(256\times256\).}
    \label{fig:a2_spectrum_2d}
  \end{subfigure}

  \caption{Propagation-signal ablation on the 2D Kraichnan turbulence benchmark. The posterior-propagation variant remains most accurate at the finer resolutions and achieves the lowest final RMSE. The forecast and posterior variants exhibit similar radial energy spectra, indicating that both retain comparable spectral amplitudes across wavenumbers; however, the posterior still yields substantially lower RMSE because the diffusion correction improves the spatial alignment and structural fidelity of vortical features. Reusing only the upsampled coarse observation performs worst at fine scales, showing that progressive posterior refinement is essential for accurate multiscale reconstruction.}
  \label{fig:ablation_propagation_signal_2d}
\end{figure}

The same qualitative conclusion holds in the 2D Kraichnan benchmark, as shown in Fig.~\ref{fig:ablation_propagation_signal_2d}. At the first refinement level, \(64\times64\), all three choices remain close in performance, with RMSE values near \(0.20\), since that stage is again driven directly by the same \(32\times32\) observation. At higher resolutions, however, the benefit of posterior propagation becomes clear. At \(128\times128\), posterior propagation attains an RMSE of \(0.1859\), compared with \(0.2390\) for forecast propagation and \(0.2831\) for raw-observation propagation. At the final \(256\times256\) resolution, the posterior variant remains best at \(0.1844\), while the forecast and raw-observation variants degrade to \(0.2950\) and \(0.2834\), respectively. The temporal RMSE curves show that posterior propagation consistently stabilizes the rollout at a substantially lower error level.

The 2D spectral comparison reveals an interesting and somewhat subtler picture. The posterior and forecast variants produce rather similar radial energy spectra, both remaining much closer to the ground truth than the raw-observation baseline in the high-\(k\) range. This indicates that the FNO forecast already carries a reasonable approximation of the \emph{spectral amplitude} distribution across wavenumbers. However, its corresponding spatial fields are still significantly misaligned, particularly in the positions and shapes of coherent vortical structures. Since the radial energy spectrum \(E(k)\) is insensitive to phase, it does not strongly penalize these positional errors, whereas RMSE does. The diffusion-corrected posterior therefore yields a much lower RMSE not because it dramatically changes the aggregate spectrum, but because it improves the \emph{structural fidelity} of the field---that is, the spatial placement and geometry of vortices. This is an important finding in its own right: the propagation signal affects not only how much fine-scale energy is present, but also whether that energy is organized correctly in physical space.

Overall, these results strongly support the propagation mechanism used in the proposed method. Intermediate diffusion posteriors are not merely auxiliary outputs; they are the vehicle through which corrected multiscale information is transferred from coarse to fine levels. Replacing them with uncorrected forecasts or repeatedly reusing the upsampled coarse observation discards that progressively refined information and substantially degrades the final reconstruction. The effect is dramatic in 1D and remains clearly beneficial in 2D, where posterior propagation improves both the quantitative accuracy and the structural coherence of the reconstructed turbulent fields.

\section{Summary Across Methods}

\begin{figure}[htbp]
  \centering
  \begin{subfigure}[t]{0.95\textwidth}
    \centering
    \includegraphics[width=\textwidth]{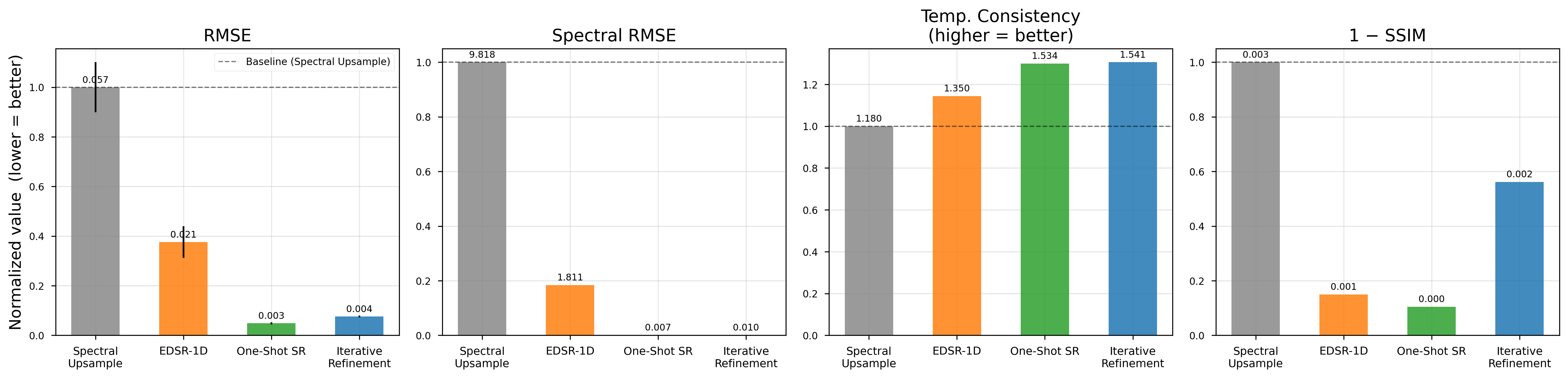}
    \caption{1D Burgers summary at resolution $512$.}
    \label{fig:summary_1d}
  \end{subfigure}

  \vspace{0.8em}

  \begin{subfigure}[t]{0.95\textwidth}
    \centering
    \includegraphics[width=\textwidth]{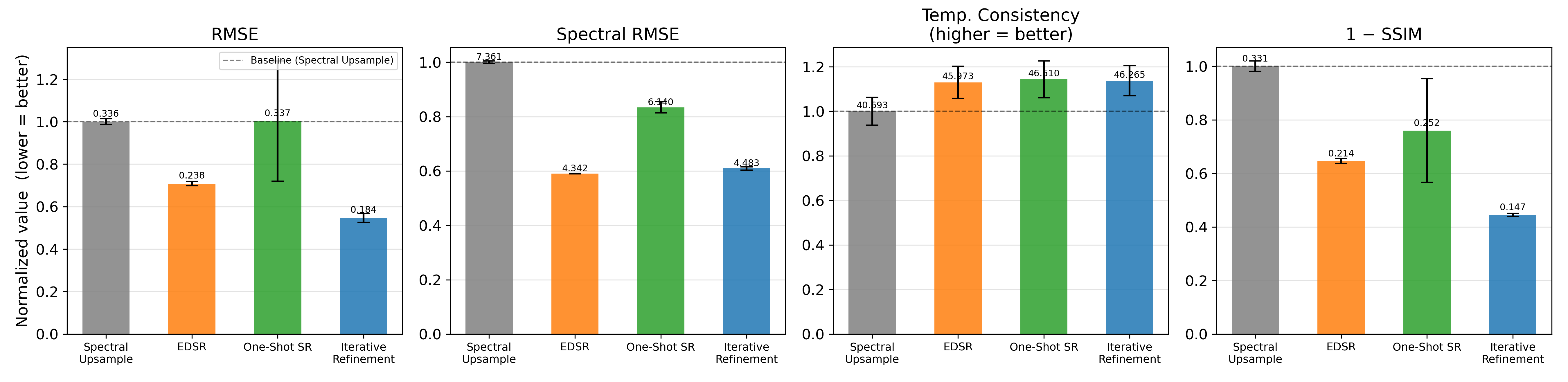}
    \caption{2D Kraichnan summary at resolution $256\times256$.}
    \label{fig:summary_2d}
  \end{subfigure}

  \caption{Summary of reconstruction metrics across all methods. \textbf{Top:} 1D Burgers benchmark, where one-shot diffusion achieves the lowest RMSE and spectral RMSE, with iterative refinement close behind. \textbf{Bottom:} 2D Kraichnan turbulence benchmark, where iterative refinement achieves the best RMSE, spectral RMSE, and SSIM, demonstrating the advantage of hierarchical correction in the more challenging multiscale setting.}
  \label{fig:summary_metrics}
\end{figure}

Figure~\ref{fig:summary_metrics} summarizes the main quantitative trends across the 1D Burgers and 2D Kraichnan benchmarks. In the 1D case, the one-shot diffusion model achieves the lowest RMSE and spectral error, with iterative refinement remaining close behind. Both stochastic methods substantially outperform deterministic EDSR and spectral upsampling. This indicates that, for the corrected 1D Burgers setting, the coarse-to-fine inverse problem is sufficiently constrained for direct one-shot generative reconstruction to perform extremely well.

The 2D Kraichnan benchmark shows a different behavior. Iterative refinement achieves the best RMSE, spectral RMSE, and SSIM among all methods, while also maintaining competitive temporal consistency. EDSR performs strongly among deterministic baselines and improves over spectral upsampling, but remains less accurate than iterative refinement. One-shot diffusion recovers plausible turbulent structure but exhibits larger errors and variability. These trends support the central claim that hierarchical forecast--analysis refinement becomes most beneficial when the reconstruction problem is strongly multiscale and underdetermined.

\section{Discussion}

The results reveal a nuanced picture of when hierarchical iterative refinement is most beneficial. On the corrected 1D Burgers benchmark, the one-shot diffusion baseline achieves the lowest RMSE and spectral error, while the proposed iterative refinement method remains close behind and substantially outperforms deterministic EDSR and spectral upsampling. In contrast, on the 2D Kraichnan turbulence benchmark, iterative refinement is the strongest method across the most important reconstruction metrics, including RMSE, spectral RMSE, and SSIM. This distinction is important: the proposed method is not simply a universally better super-resolution model, but a framework whose advantage becomes most pronounced when the inverse problem is strongly multiscale, temporally coupled, and underdetermined.

\subsection{Interpretation from a Data Assimilation Perspective}

From a data assimilation viewpoint, the proposed method can be interpreted as a learned multiscale filtering procedure. The shared FNO forecaster provides dynamical priors analogous to forecast models in classical sequential estimation, while the diffusion corrector acts as a nonlinear stochastic analysis operator. Instead of applying an explicit Kalman-style update, the analysis step is learned from data and conditioned on both the forecast prior and the current coarser-resolution state.

The prior--posterior comparisons and propagation-signal ablations show that the diffusion corrector does not merely synthesize a new field independently at every stage. Rather, the method progressively improves the state estimate as information moves upward through the resolution hierarchy. The corrected posterior at each intermediate level becomes an informative conditioning signal for the next stage. This is a key distinction from one-shot super-resolution: the method builds a sequence of corrected multiscale posteriors rather than attempting to recover all missing scales at once.

\subsection{When Does Hierarchical Refinement Help?}

The comparison between the 1D and 2D testbeds suggests that the value of hierarchical refinement depends on the complexity of the coarse-to-fine inverse problem. In 1D Burgers, the dynamics are strongly structured and the dominant discontinuity-like features are often well constrained by the coarse observation and temporal context. As a result, a sufficiently expressive one-shot diffusion model can directly reconstruct the high-resolution state with excellent accuracy. Iterative refinement remains stable and competitive, but the additional cascade does not provide a decisive advantage in this simpler setting.

In 2D Kraichnan turbulence, however, the reconstruction problem is much less constrained. Thin vorticity filaments, vortex interfaces, and localized merger structures are not uniquely determined by the coarse observation. One-shot methods must infer the full missing range of scales in a single step, which leads either to over-smoothing, as in deterministic EDSR, or to less stable fine-scale texture, as in one-shot diffusion. Iterative refinement reduces this difficulty by decomposing the full $32\times32 \rightarrow 256\times256$ recovery into smaller resolution-wise corrections. This produces better spectral recovery, higher SSIM, and lower RMSE in the 2D benchmark.

\subsection{Role of the Cascade}

The cascade-depth and propagation-signal ablations provide direct evidence that the hierarchy is not merely an architectural convenience. Reducing the testbeds from the full cascade to shallower variants substantially increases RMSE and high-wavenumber spectral error. Similarly, replacing intermediate posterior propagation with either raw observations or uncorrected forecasts degrades the final reconstruction. These results show that the intermediate posteriors carry useful corrected information that compounds across stages.

This supports the central mechanism of the method: each stage produces a posterior that is both dynamically informed by the forecast prior and observationally constrained by the coarser state. Passing this posterior to the next stage yields a progressively refined multiscale estimate. Without this corrected propagation signal, the cascade loses much of its advantage.

\subsection{Limitations}

Despite its strengths, the current formulation has several limitations. First, the cascade structure introduces stage-to-stage error propagation. Since only the first refinement stage directly observes the true coarse input, later stages depend on generated intermediate posteriors. Errors at lower levels may therefore propagate upward, especially in longer rollouts or more chaotic regimes.

Second, the shared corrector imposes a strong parameter-sharing assumption across resolution transitions. This makes the method compact and resolution-conditioned, but the statistics of different refinement tasks may not be identical. For example, the correction from $64\rightarrow128$ may involve different structures than the correction from $128\rightarrow256$ or $256\rightarrow512$. A single shared corrector may therefore be suboptimal for some scales.

Third, the diffusion sampler introduces practical sensitivities. The stochasticity parameter and number of reverse steps affect both accuracy and cost, especially in the 2D turbulence case. Although the results suggest that moderate step counts are sufficient, sampler tuning remains an important component of the full pipeline.

Finally, the present training strategy still relies on teacher-forced forecast priors. This reduces train--test mismatch relative to training on idealized priors, but it does not fully reproduce inference conditions, where forecasters receive previous generated posteriors rather than ground-truth states. More inference-aware training strategies may further improve stability and long-horizon performance.

\subsection{Computational Trade-offs and Future Improvements}

Iterative refinement is more computationally expensive than spectral upsampling, deterministic EDSR, or a single one-shot diffusion model, because it performs diffusion-based correction at multiple resolution levels. The results suggest that this extra cost is most justified in difficult multiscale settings such as 2D turbulence, where the method provides clear gains in accuracy, structure, and spectral fidelity.

A promising direction is to relax the fully shared-corrector assumption. In the present work, a single corrector is reused across all refinement stages for conceptual simplicity and parameter efficiency. However, a stronger iterative refinement model may be obtained through \emph{aggregate training}, where each resolution transition is trained separately or partially specialized. Such a strategy would allow the correction model at each scale to adapt to the statistics of that specific refinement task, potentially improving the quality of the final posterior. This would increase training cost and model storage, but it may be worthwhile for complex turbulence or high-dimensional geophysical applications.

\section{Conclusion}

We introduced an iterative refinement framework for super-resolved data assimilation of multiscale physical systems. The central idea is to replace a single coarse-to-fine reconstruction with a sequence of resolution-wise forecast--analysis updates across a multiresolution hierarchy. At each target resolution, a shared Fourier Neural Operator forecaster provides a learned dynamical prior using spectral mode slicing, while a shared conditional diffusion corrector produces an analysis posterior conditioned on the current coarser-resolution state. This design combines temporal forecasting, observational correction, and generative fine-scale reconstruction within a single learned assimilation pipeline.

The experiments show that the value of iterative refinement depends strongly on the difficulty of the underlying inverse problem. On the 1D stochastic Burgers benchmark, direct one-shot diffusion achieves the lowest RMSE and spectral error, indicating that a sufficiently expressive generative model can solve the reconstruction task very effectively when the coarse observation strongly constrains the missing fine-scale structure. Iterative refinement remains close in accuracy and substantially outperforms deterministic EDSR and spectral upsampling, but it does not dominate the one-shot diffusion baseline in this simpler setting. This result is important because it shows that hierarchical refinement is not merely a universally stronger super-resolution model; rather, its advantage emerges when the reconstruction problem is sufficiently underdetermined.

On the 2D Kraichnan turbulence benchmark, the benefit of the proposed framework becomes much clearer. In the challenging \(32\times32 \rightarrow 256\times256\) reconstruction setting, iterative refinement achieves the best RMSE, spectral fidelity, and SSIM among the learned super-resolution baselines. It also produces temporally stable reconstructions with improved recovery of vortex interfaces, thin filaments, and localized turbulent structures. These results support the central premise of the method: decomposing a difficult multiscale inverse problem into a sequence of smaller forecast--analysis refinements makes the reconstruction better conditioned and improves the recovery of unresolved scales.

The ablation studies further clarify the mechanism behind these gains. Increasing cascade depth improves reconstruction quality, showing that intermediate refinement stages are not merely architectural overhead but contribute directly to fine-scale recovery. Similarly, propagating corrected diffusion posteriors between stages is essential; replacing these posteriors with uncorrected forecasts or repeatedly upsampled coarse observations substantially degrades performance. These findings indicate that the method's advantage comes from the structure of the multiscale assimilation cascade itself, not only from the use of a powerful diffusion model.

The comparison with ensemble Kalman filtering baselines highlights the computational role of the proposed approach. A solver-based EnKF remains a strong reference when the high-resolution physical solver is available online, and can achieve lower aggregate RMSE. However, this accuracy requires repeatedly advancing an ensemble of full-resolution states. Iterative refinement instead provides a learned super-resolved assimilation alternative that avoids online full-solver forecasts, while substantially improving over a learned EnKF based on the same FNO forecast family. Thus, the proposed method occupies a useful middle ground between inexpensive learned filtering and expensive solver-based assimilation.

Several directions remain open. Future work should explore inference-aware training to further reduce residual train--test mismatch, explicit spectral or physics-informed regularization to improve high-wavenumber fidelity, and uncertainty quantification through ensemble posterior sampling. Extensions to partially observed, noisy, three-dimensional, or more strongly chaotic systems are also important. Another promising direction is aggregate or stage-specialized training, in which separate correctors are trained for different resolution transitions rather than sharing a single corrector across all scales. Such specialization may improve reconstruction quality in complex turbulence or geophysical applications, at the cost of additional parameters and training effort.

Overall, the results support the view that learned data assimilation benefits from combining temporal priors with hierarchical generative correction. One-shot diffusion can be sufficient for simpler and more constrained systems, but for complex multiscale turbulence, iterative refinement provides a more accurate and physically faithful pathway from coarse observations to high-resolution state estimates.


\section*{Declaration of competing interest}
    The authors declare that they have no known competing financial interests or personal relationships that could have appeared to influence the work reported in this paper.

\section*{Acknowledgments}
This work was supported by the AFOSR Grant FA9550-24-1-0327.

\section*{Data availability} 
Data supporting the findings of this study are available from the corresponding author upon reasonable request.

\section*{Code availability}
The implementation associated with this study is publicly available in the GitHub repository \texttt{Iterative\_refinement\_DA}: \url{https://github.com/dmrigank/Iterative_refinement_DA}. A citable archival version of the repository is available through Zenodo at \url{https://doi.org/10.5281/zenodo.21494987}~\cite{dhingra2026iterativecode}.

\section*{Declaration of Generative AI use}
During the preparation of this manuscript, the authors used generative AI–assisted tools (Claude Sonnet 5.0) solely for minor language editing, including correction of spelling and grammatical errors. The tools were not used to generate scientific content, data, analyses, or interpretations. All technical content, results, and conclusions were developed by the authors, who reviewed and approved the final manuscript and take full responsibility for its contents.

\bibliographystyle{unsrtnat}
\bibliography{references}


\appendix

\section{Additional Baseline Comparisons}
\label{app:baseline_comparisons}

This appendix provides additional qualitative diagnostics for the baseline comparisons discussed in Section~\ref{sec:baseline_comparison}. These figures support the main conclusions reported in Tables~\ref{tab:burgers-benchmark} and~\ref{tab:kraichnan-benchmark}: autoregressive forecasting without analysis is unstable over long rollouts, one-shot diffusion is highly effective for the simpler 1D Burgers benchmark, and iterative refinement provides the strongest reconstruction quality in the more underdetermined 2D Kraichnan setting.

\subsection{Autoregressive FNO-Only Rollout Diagnostics}
\label{app:fno_only_diagnostics}

The FNO-only baseline removes the diffusion analysis step and rolls out the shared mode-sliced FNO forecaster autoregressively. This diagnostic isolates whether the learned dynamical prior alone is sufficient for stable high-resolution reconstruction.

\begin{figure}[htbp]
  \centering
  \includegraphics[width=\textwidth]{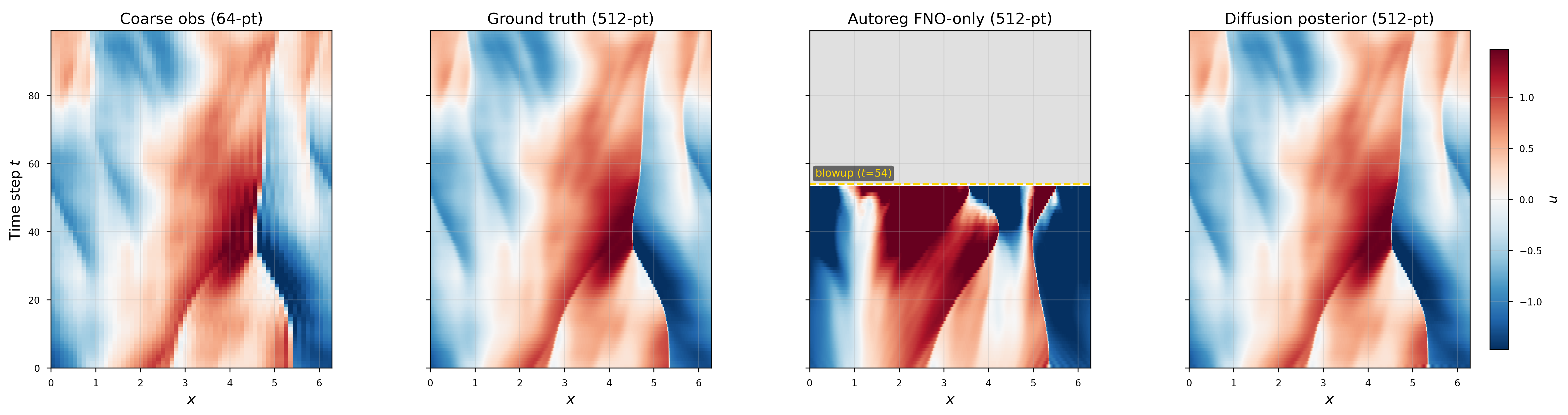}
  \caption{Autoregressive FNO-only diagnostic on the 1D Burgers benchmark at resolution \(512\). The FNO-only rollout initially follows the large-scale structure but becomes unstable and blows up at \(t=54\), after which the predicted trajectory loses physical consistency. The diffusion posterior remains closely aligned with the ground-truth shock evolution over the full time horizon.}
  \label{fig:app_1d_fno_hovmoller}
\end{figure}

In the 1D Burgers benchmark, the FNO-only rollout initially tracks the large-scale shock structure but rapidly drifts from the ground truth and blows up around \(t\approx54\), as shown in Fig.~\ref{fig:app_1d_fno_hovmoller}. After this point, the predicted field loses physical coherence and no longer represents a plausible Burgers trajectory. By contrast, the diffusion posterior remains well aligned with the ground-truth Hovmöller structure over the full time horizon.

\begin{figure}[htbp]
  \centering
  \includegraphics[width=\textwidth]{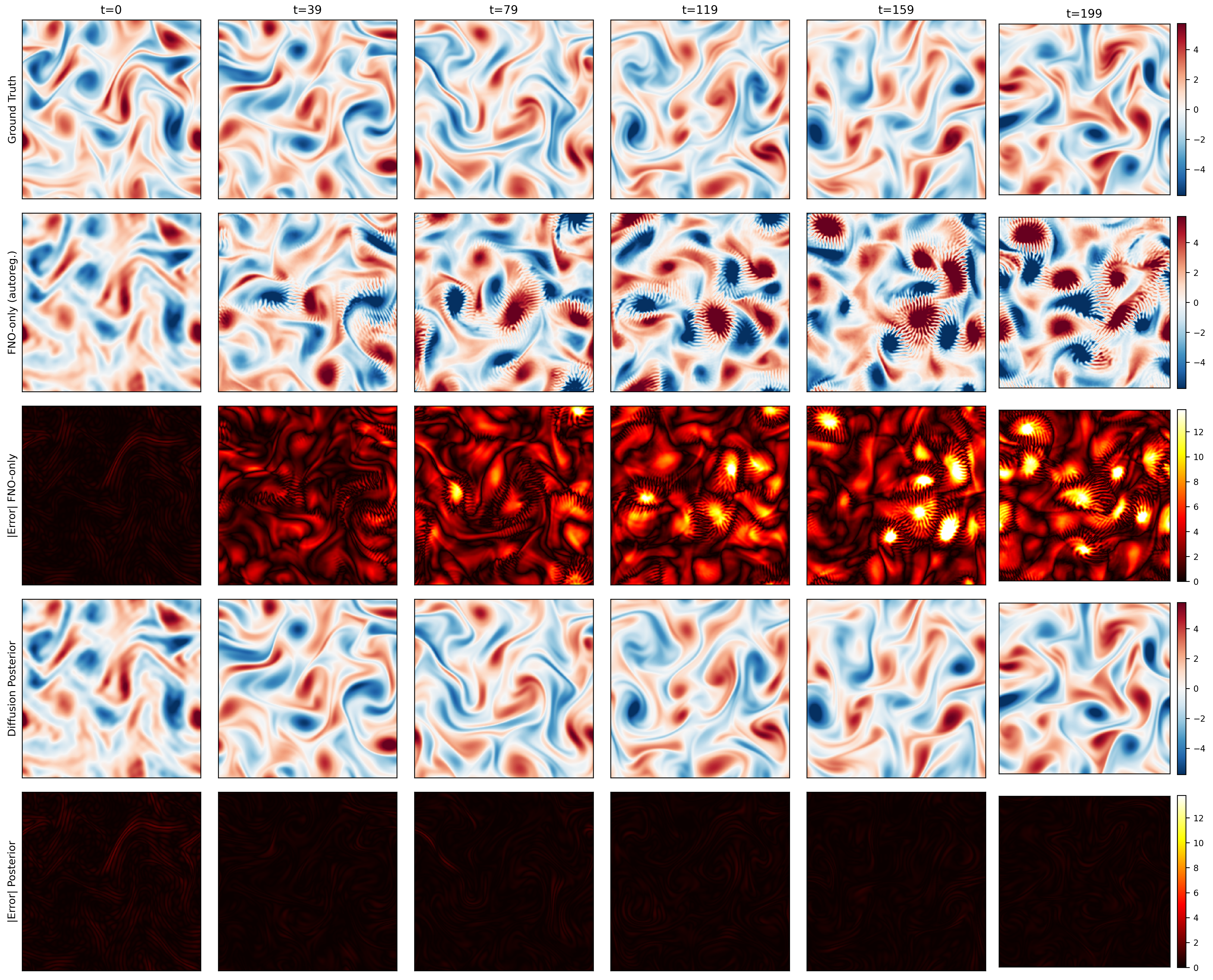}
  \caption{Autoregressive FNO-only diagnostic on the 2D Kraichnan benchmark at resolution \(256\times256\) for a representative trajectory. Columns show increasing time, while rows compare ground truth, the FNO-only baseline, its absolute error, the diffusion posterior, and its absolute error. The FNO-only rollout accumulates substantial error and develops spurious small-scale structures, whereas the diffusion posterior preserves the correct vortex and filament organization with much smaller error throughout the sequence.}
  \label{fig:app_2d_fno_rollout}
\end{figure}

The 2D Kraichnan case exhibits a less abrupt but still substantial failure mode. Figure~\ref{fig:app_2d_fno_rollout} shows that the FNO-only rollout accumulates error, develops spurious small-scale oscillations, and departs from the correct vortex--filament organization over time. The diffusion posterior preserves the large-scale morphology and fine-scale coherence of the ground truth with substantially smaller residual error. These results confirm that the shared FNO is useful as a dynamical prior, but repeated observation-driven analysis is essential for stable sequential reconstruction.

\subsection{Additional 1D One-Shot Baseline Diagnostics}
\label{app:oneshot_baseline_diagnostics}

We provide additional qualitative comparisons against the deterministic EDSR and stochastic one-shot diffusion baselines on the 1D Burgers benchmark. These one-shot methods reconstruct the finest-resolution state directly from the current coarse observation, without propagating intermediate posteriors through a multiresolution hierarchy.

\begin{figure}[htbp]
  \centering
  \includegraphics[width=\textwidth]{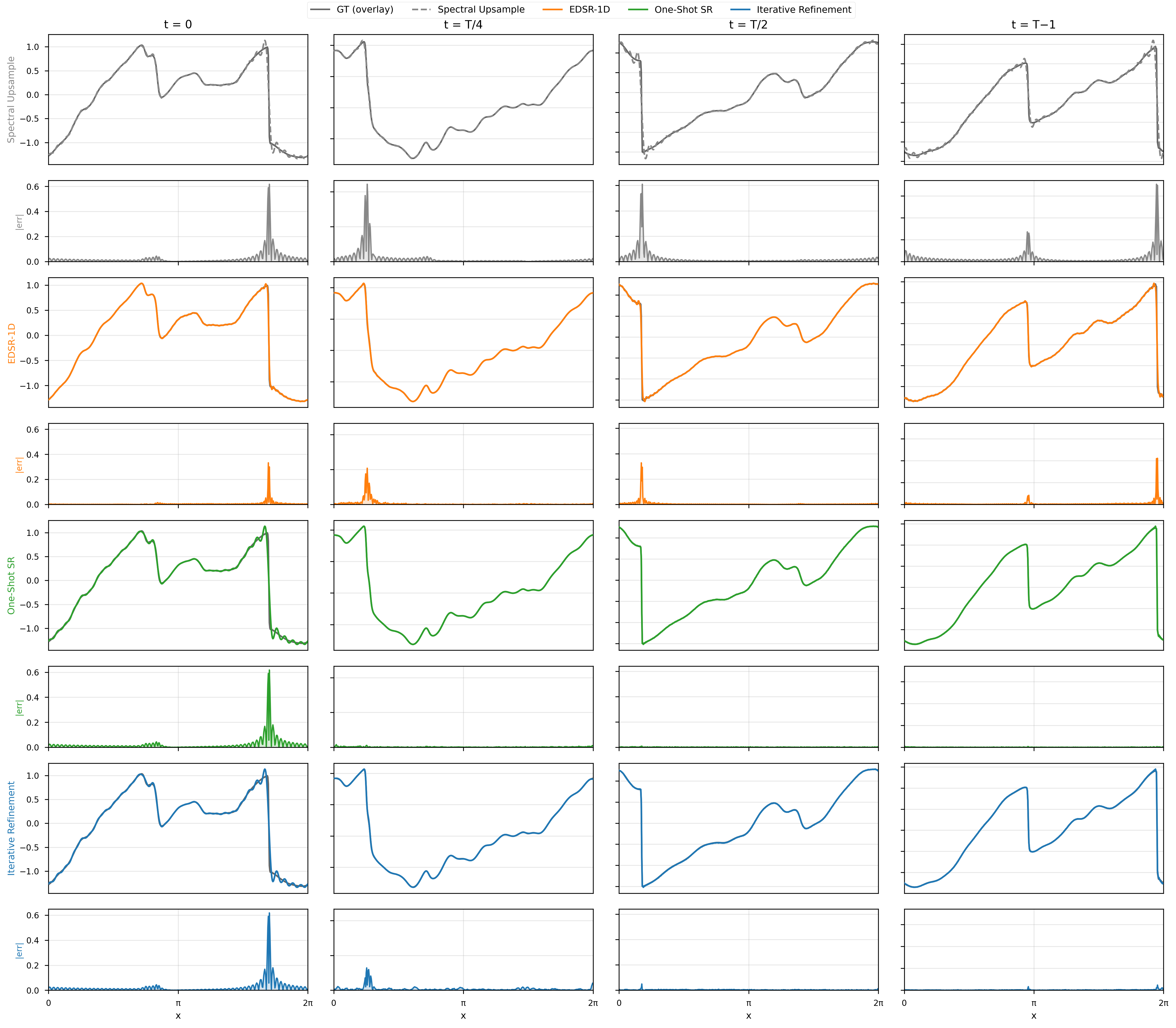}
  \caption{Snapshot comparison on the 1D stochastic Burgers benchmark at resolution \(512\) for a representative trajectory at four times (\(t=0\), \(T/4\), \(T/2\), and \(T-1\)). For each method, the reconstructed signal is shown together with the pointwise absolute error relative to the ground truth. Both the one-shot diffusion and iterative refinement models closely track the Burgers shock structure, while EDSR improves over spectral upsampling but exhibits larger localized deviations near sharp transitions.}
  \label{fig:app_1d_all_methods_snap}
\end{figure}

\begin{figure}[htbp]
  \centering
  \includegraphics[width=0.9\textwidth]{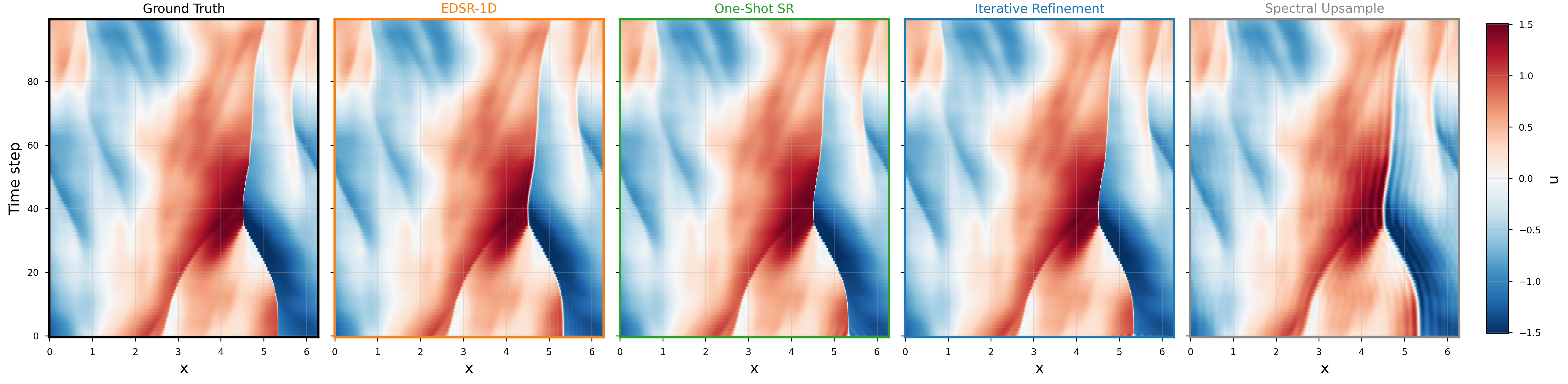}
  \caption{Space--time comparison on the 1D stochastic Burgers benchmark for a representative trajectory. The Hovmöller plots show the full temporal evolution for the ground truth and the reconstructions from EDSR, one-shot diffusion, iterative refinement, and spectral upsampling. The learned methods recover the dominant space--time structure well, with one-shot diffusion and iterative refinement remaining closest to the ground truth over the full rollout.}
  \label{fig:app_1d_all_methods_hovmoller}
\end{figure}

As shown in Figs.~\ref{fig:app_1d_all_methods_snap} and~\ref{fig:app_1d_all_methods_hovmoller}, both one-shot diffusion and iterative refinement closely track the ground-truth shock structure. EDSR improves over spectral upsampling but exhibits larger localized deviations near sharp gradients. These results are consistent with the quantitative ordering in Table~\ref{tab:burgers-benchmark}, where one-shot diffusion slightly outperforms iterative refinement.

\section{Additional Temporal Stability Diagnostics}
\label{app:temporal_stability}

\subsection{Temporal Stability on the 1D Burgers Benchmark}
\label{app:temporal_stability_1d}

\begin{figure}[htbp]
  \centering
  \begin{subfigure}[t]{0.49\textwidth}
    \centering
    \includegraphics[width=\textwidth]{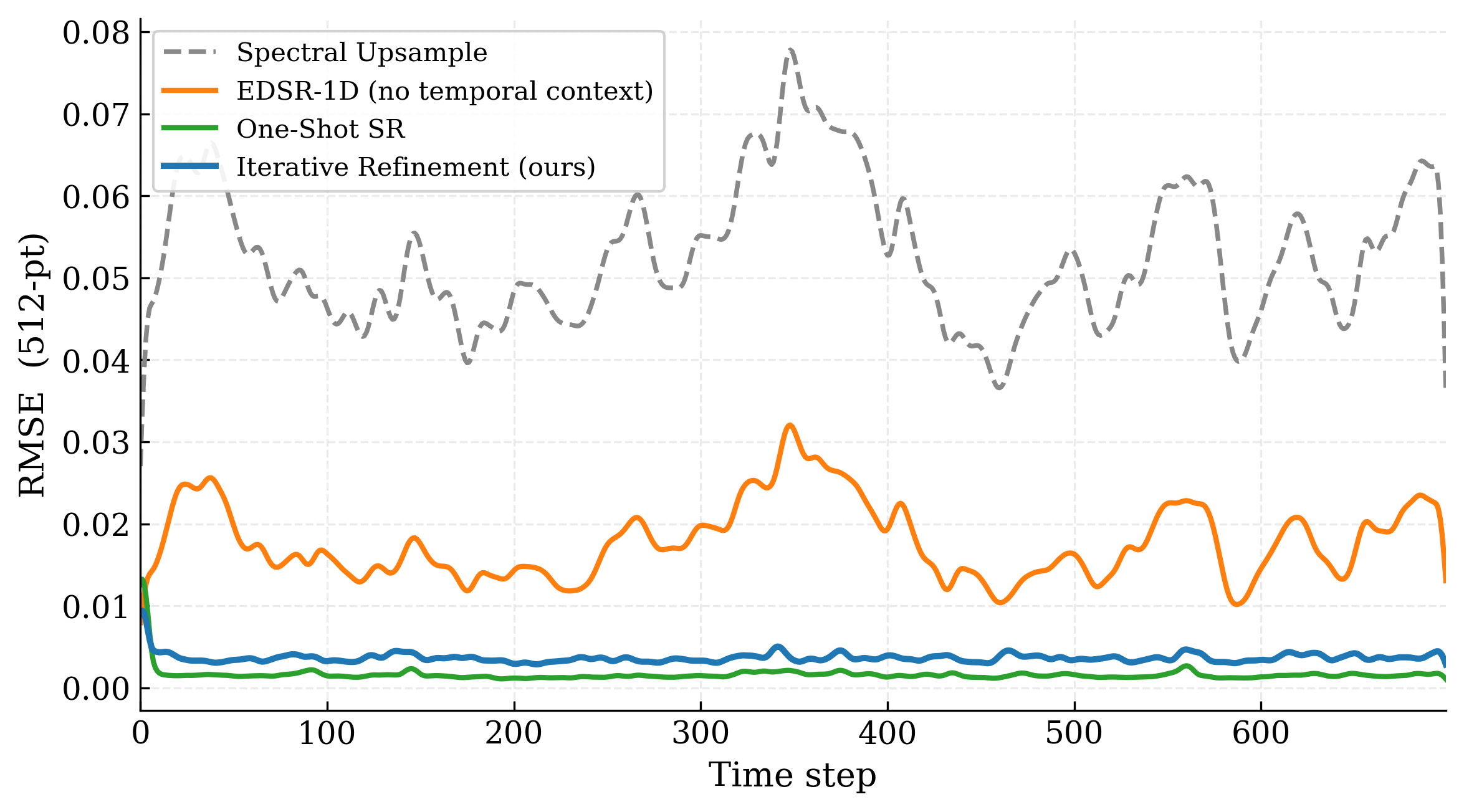}
    \caption{RMSE over time.}
    \label{fig:1d_trmse_comp_app}
  \end{subfigure}
  \hfill
  \begin{subfigure}[t]{0.49\textwidth}
    \centering
    \includegraphics[width=\textwidth]{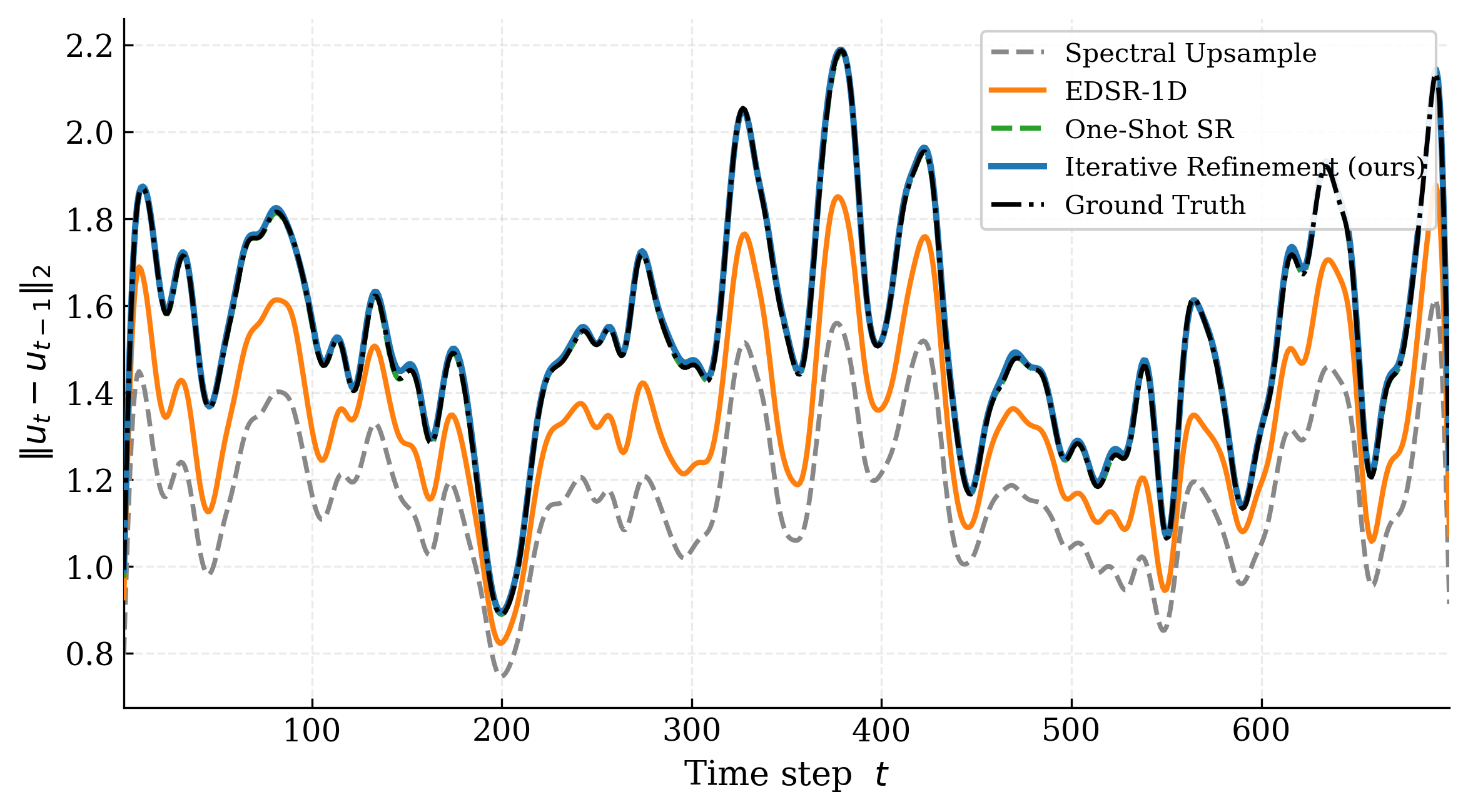}
    \caption{Frame-to-frame temporal consistency.}
    \label{fig:1d_temporal_consistency_comp_app}
  \end{subfigure}
  \caption{Temporal stability diagnostics on the 1D Burgers benchmark at resolution \(512\), averaged over the test trajectories. \textbf{(a)} RMSE over time. One-shot diffusion achieves the lowest temporal RMSE, while iterative refinement remains close and stable throughout the rollout. \textbf{(b)} Temporal consistency measured by the frame-to-frame displacement norm \(\|u_t-u_{t-1}\|_2\). One-shot diffusion and iterative refinement closely follow the ground-truth temporal variation, whereas EDSR and spectral upsampling are more dissipative.}
  \label{fig:1d_temporal_stability_app}
\end{figure}

Figure~\ref{fig:1d_temporal_stability_app} reports the corresponding temporal stability diagnostics for the 1D Burgers benchmark. The left panel shows the RMSE over time at the finest resolution \(N=512\), averaged over the test trajectories. All learned super-resolution methods remain bounded and substantially outperform spectral upsampling over the full horizon. Among them, the one-shot diffusion model attains the lowest RMSE across most of the rollout, while iterative refinement remains a close second and preserves a similarly flat, low-error profile after the initial transient. EDSR improves substantially over spectral upsampling but remains consistently less accurate than the diffusion-based models.

The right panel compares the frame-to-frame displacement norm \(\|u_t-u_{t-1}\|_2\), again averaged over the test trajectories. One-shot diffusion and iterative refinement closely track the ground-truth temporal variation, indicating that both methods preserve the dominant temporal dynamics of the Burgers solution. EDSR underestimates some of the sharper temporal changes, while spectral upsampling is systematically too smooth. These results are consistent with the main quantitative finding for the 1D benchmark: direct one-shot diffusion is slightly more accurate, but iterative refinement remains temporally stable and dynamically consistent over long rollouts.

\section{Forecast-Prior Diagnostics}
\label{app:forecast_prior_diagnostics}

To better understand the role of the learned forecast prior, we compare the ground truth, the one-step FNO forecast, and the diffusion-corrected analysis posterior. These visualizations show that the forecast prior already provides a strong approximation to the next-step state, while the diffusion corrector refines this prior in a structured and localized manner rather than reconstructing the full field from scratch.

\begin{figure}[htbp]
  \centering
  \includegraphics[width=\textwidth]{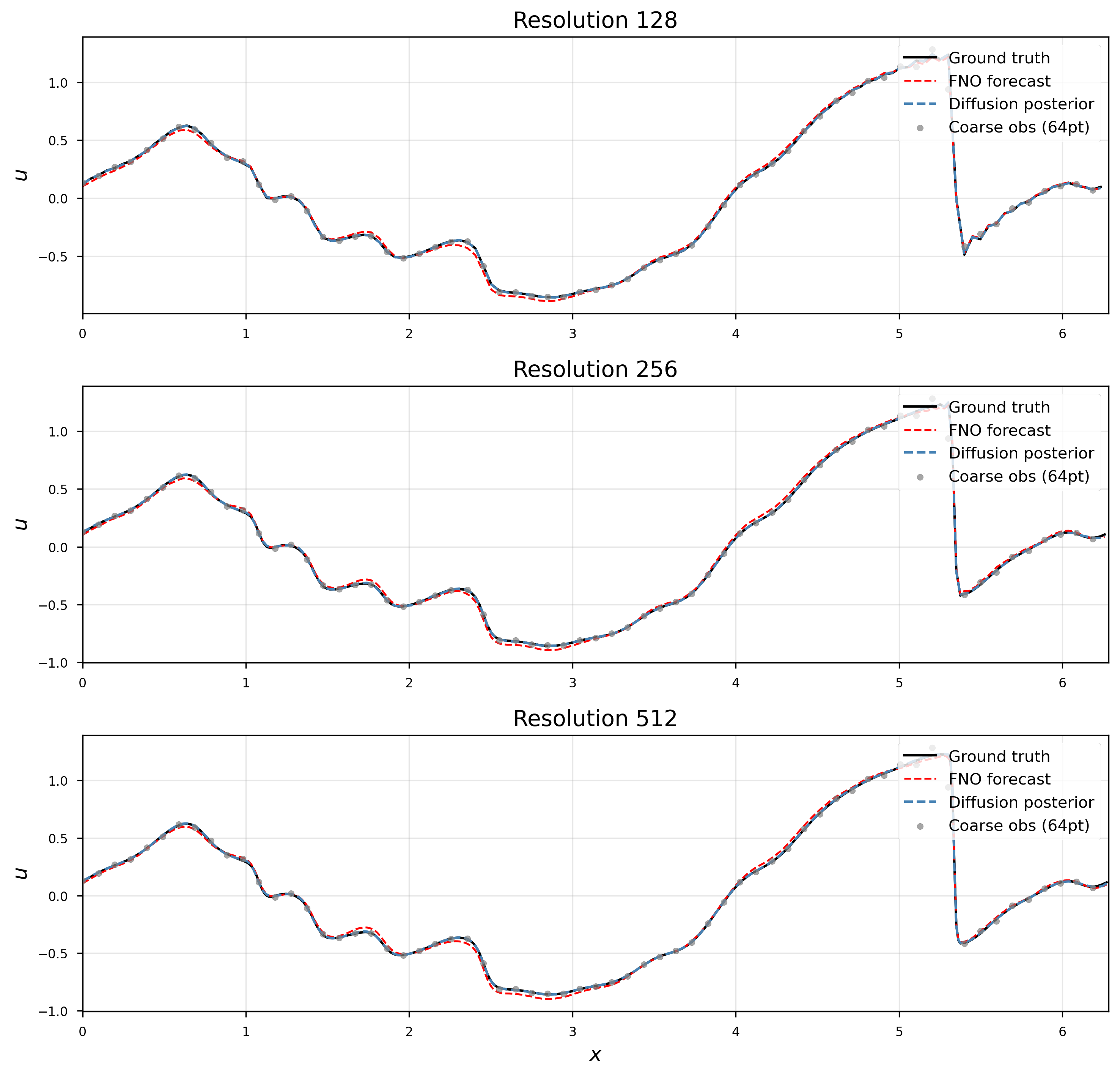}
  \caption{Forecast-prior diagnostic on the 1D Burgers benchmark. At a representative time step, the one-step FNO forecast already captures the global solution profile across resolutions, but deviates near the shock and steep-gradient regions. The diffusion posterior corrects these local errors and aligns more closely with the ground truth, showing that the corrector acts primarily as a targeted refinement of the forecast prior.}
  \label{fig:app_1d_prior_vs_posterior}
\end{figure}

In the 1D Burgers case, shown in Fig.~\ref{fig:app_1d_prior_vs_posterior}, the FNO prior captures the overall shock profile and large-scale solution shape at all refinement levels, but exhibits noticeable deviations near steep gradients and around the shock location. The diffusion posterior consistently reduces these discrepancies and brings the reconstruction into closer agreement with the ground truth, especially in the vicinity of sharp transitions. This indicates that the corrector is primarily compensating for forecast bias and sharpening local structure, rather than replacing the forecast entirely.

\begin{figure}[htbp]
  \centering
  \includegraphics[width=\textwidth]{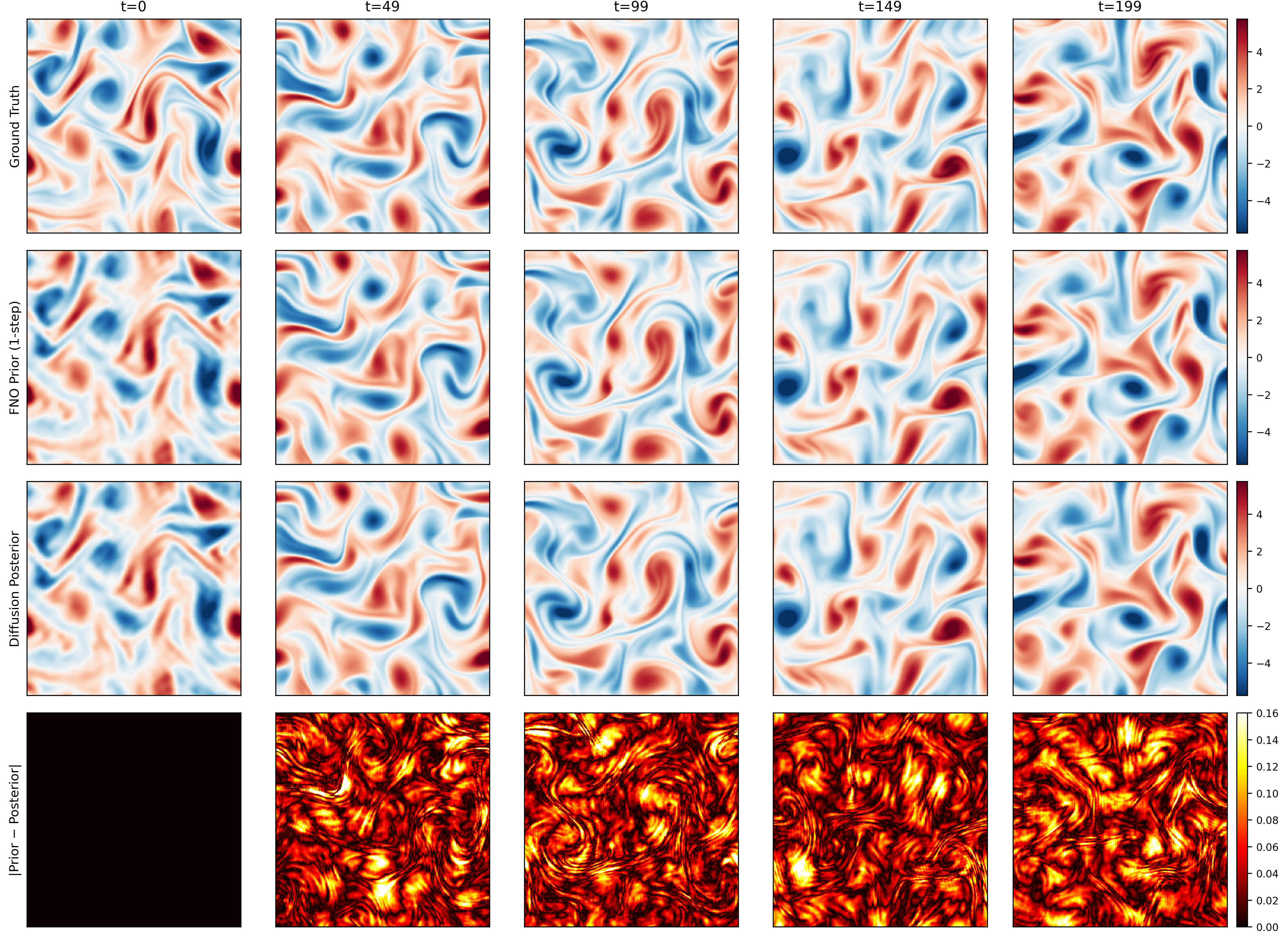}
  \caption{Forecast-prior diagnostic on the 2D Kraichnan benchmark at resolution \(256\times256\). Rows show the ground truth, the one-step FNO prior, the diffusion posterior, and the magnitude of the posterior correction relative to the prior. The forecast already captures the large-scale flow organization, while the diffusion corrector applies localized adjustments concentrated around filaments and vortex interfaces, indicating residual analysis rather than full re-synthesis.}
  \label{fig:app_2d_prior_vs_posterior}
\end{figure}

A similar pattern appears in the 2D Kraichnan case, shown in Fig.~\ref{fig:app_2d_prior_vs_posterior}. The one-step FNO prior already reproduces much of the large-scale vortex and filament organization, confirming that the forecaster provides a meaningful dynamical prior. The posterior then applies comparatively small but spatially coherent corrections, concentrated along thin filaments, shear layers, and vortex interfaces. The prior--posterior difference fields make this especially clear: the analysis is localized and flow-aware, not a wholesale re-generation of the state.

These observations support the forecast--analysis decomposition used throughout the method. The forecast prior supplies the global temporal structure of the evolving state, while the diffusion corrector focuses on reducing residual forecast error and restoring fine-scale detail consistent with the current coarse observation. This separation of roles helps explain why the iterative refinement procedure remains both accurate and stable over long rollouts.

\section{Sensitivity to Diffusion Sampling Settings}
\label{app:sampler_sensitivity}

The proposed method relies on iterative diffusion sampling at inference time. Its practical utility therefore depends not only on reconstruction quality, but also on how sensitive that quality is to sampler hyperparameters. We examine two aspects of the sampling process: the stochasticity parameter \(\eta\) in DDIM sampling and the number of reverse diffusion steps.

\subsection{Effect of Stochasticity}

\begin{figure}[htbp]
  \centering
  \includegraphics[width=\textwidth]{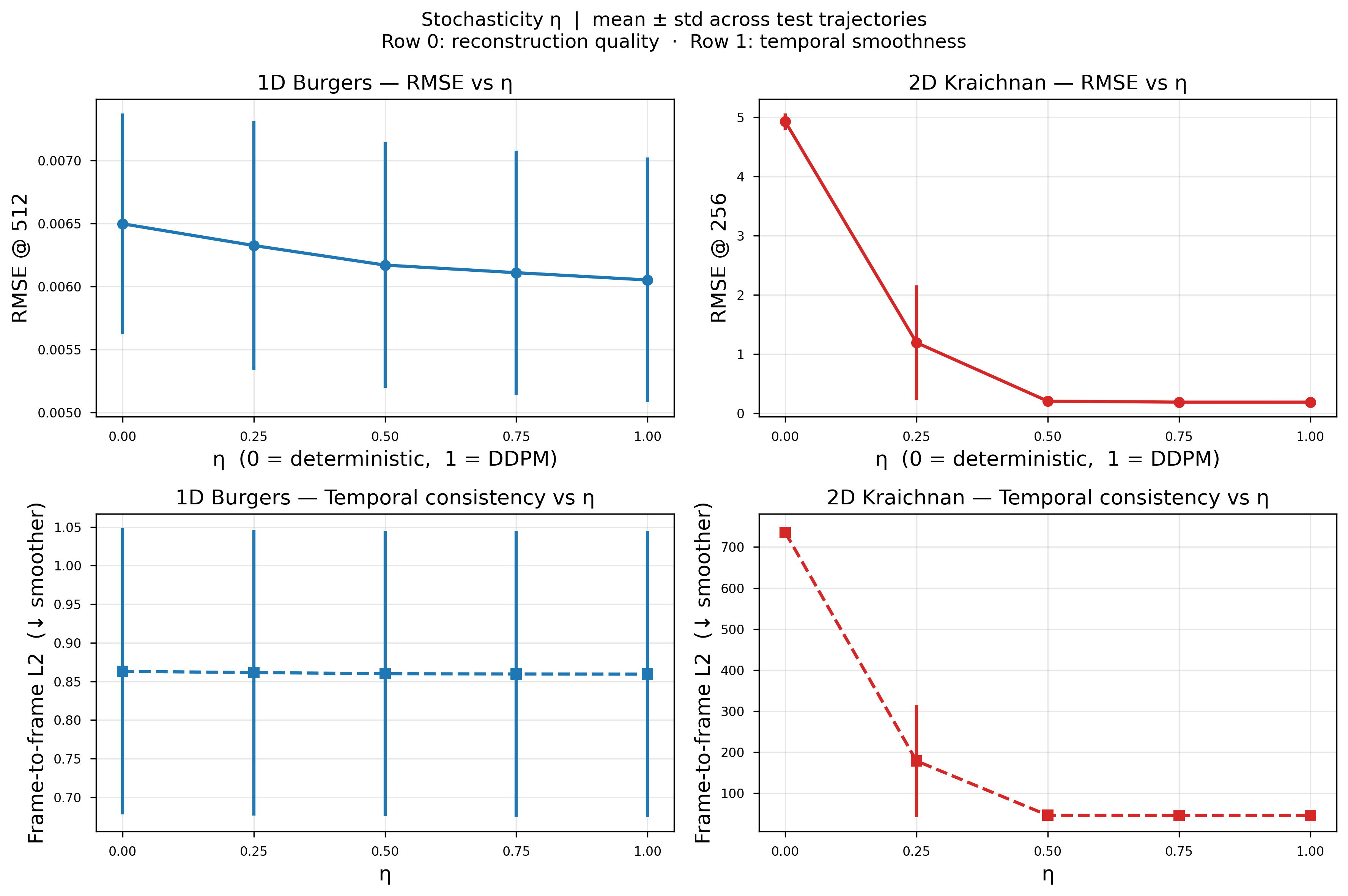}
  \caption{Sensitivity of iterative refinement to the DDIM stochasticity parameter \(\eta\), reported as mean \(\pm\) standard deviation across test trajectories. Top row: reconstruction RMSE. Bottom row: temporal smoothness measured by frame-to-frame \(\ell_2\) displacement. In the 1D Burgers case, performance changes only mildly with \(\eta\). In the 2D Kraichnan case, deterministic or weakly stochastic sampling performs poorly, while moderate stochasticity \((\eta \ge 0.5)\) yields both lower RMSE and substantially smoother temporal behavior.}
  \label{fig:app_eta_factor}
\end{figure}

Figure~\ref{fig:app_eta_factor} shows that the role of sampling stochasticity differs substantially between the two testbeds. In the 1D Burgers case, reconstruction quality is only weakly sensitive to \(\eta\): RMSE decreases slightly as \(\eta\) increases from deterministic sampling \((\eta=0)\) to DDPM-like stochastic sampling \((\eta=1)\), while the temporal smoothness metric remains essentially unchanged. This suggests that, in the simpler 1D setting, the conditional posterior is already relatively concentrated and the method is not strongly affected by the amount of injected reverse-process noise.

In contrast, the 2D Kraichnan benchmark is highly sensitive to \(\eta\) at low values. Deterministic or nearly deterministic sampling leads to severe degradation in both RMSE and temporal smoothness, whereas performance improves substantially once moderate stochasticity is introduced. For \(\eta \ge 0.5\), both metrics stabilize and remain nearly flat. This indicates that, in the more challenging turbulent setting, stochastic sampling is not merely a cosmetic choice but an important ingredient for robust posterior reconstruction. A likely interpretation is that moderate stochasticity helps the sampler avoid poor deterministic denoising trajectories in a more complex and multimodal conditional landscape.

\subsection{Effect of DDIM Step Count}

\begin{figure}[htbp]
  \centering
  \includegraphics[width=\textwidth]{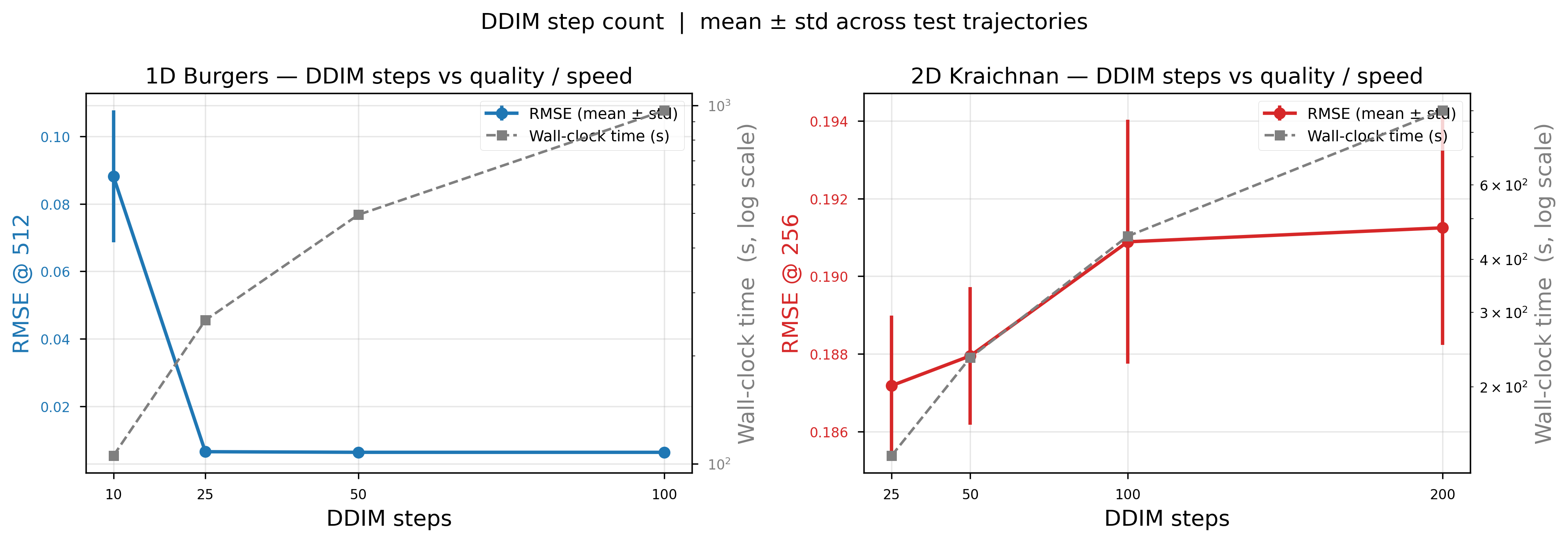}
  \caption{Sensitivity of iterative refinement to the number of DDIM reverse steps, shown as mean \(\pm\) standard deviation across test trajectories together with wall-clock inference time. In the 1D Burgers benchmark, RMSE improves sharply from \(10\) to \(25\) steps and then plateaus, indicating diminishing returns beyond moderate step counts. In the 2D Kraichnan benchmark, RMSE is relatively insensitive to the number of steps over the tested range, while computational cost increases steadily, highlighting a clear quality--cost trade-off.}
  \label{fig:app_ddim_steps}
\end{figure}

Figure~\ref{fig:app_ddim_steps} shows the trade-off between reconstruction accuracy and inference cost as the number of DDIM steps is varied. In the 1D Burgers case, increasing the step count from \(10\) to \(25\) yields a large improvement in RMSE, but further increases to \(50\) and \(100\) produce little additional gain while continuing to increase wall-clock time. Thus, for Burgers, the practical operating point lies near the onset of this plateau.

For the 2D Kraichnan case, the method is comparatively insensitive to the number of DDIM steps over the tested range. The mean RMSE changes only slightly from \(25\) to \(200\) steps, and the error bars overlap substantially, while inference cost increases monotonically. This suggests that, once a moderate number of reverse steps is used, iterative refinement is already operating in a regime where additional denoising iterations provide diminishing returns.

\subsection{Implications}

Taken together, these results show that the proposed method is more sensitive to \emph{how} sampling is performed than to \emph{how long} it is run. In particular, the stochasticity parameter \(\eta\) is critical in the harder 2D setting, whereas the number of reverse steps mainly controls the quality--cost trade-off and exhibits a clear plateau beyond moderate values. This behavior is encouraging from a deployment perspective: the method does not require extremely long diffusion chains, but it does benefit from a carefully chosen stochastic sampling regime, especially in strongly multiscale turbulent problems.

\section{Hyperparameters and architecture details}
\label{app: hyperparams}

\begin{table}[htbp]
\centering
\caption{PDE parameters and dataset configuration for the two testbeds.}
\label{tab:hyperparams_data}
\small
\begin{tabular}{l l l}
\toprule
\textbf{Parameter} & \textbf{1D Burgers} & \textbf{2D Kraichnan} \\
\midrule
\multicolumn{3}{l}{\textit{PDE \& Domain}} \\[2pt]
Domain                        & $[0, 2\pi]$, periodic          & $[0, 2\pi]^2$, periodic \\
Solver resolution             & $N = 512$                     & $256 \times 256$ \\
Solver timestep               & $\Delta t = 10^{-4}$           & $\Delta t = 5 \times 10^{-3}$ \\
Viscosity $\nu$               & $5 \times 10^{-3}$             & $1 \times 10^{-3}$ \\
Large-scale damping           & Linear friction $\alpha = 0.1$ & Ekman drag $\mu = 0.05$ \\
Stochastic forcing            & OU, $K_f{=}16$ modes, $\sigma_k \propto k^{-1}$ & OU band, $k_f{=}4$, width $= 1.5$ \\
OU correlation time $\tau$    & $0.5$                          & $0.5$ \\
Output cadence $\delta t$     & $0.05$                         & $0.05$ \\
\midrule
\multicolumn{3}{l}{\textit{Resolution Hierarchy}} \\[2pt]
Observation resolution        & 64-pt                          & $32 \times 32$ \\
Intermediate levels           & 128, 256                       & $64{\times}64$,\ $128{\times}128$ \\
Target resolution             & 512-pt                         & $256 \times 256$ \\
Refinement stages $R$         & 3                              & 3 \\
\midrule
\multicolumn{3}{l}{\textit{Dataset}} \\[2pt]
Total trajectories            & 50                             & 50 \\
Train / val / test            & 40 / 5 / 5                     & 44 / 5 / 3 \\
Snapshots per trajectory      & 800 (after 200 spinup)         & 200 (after 4000-step burn-in) \\
\bottomrule
\end{tabular}
\end{table}

\begin{table}[htbp]
\centering
\caption{Model architecture details. The diffusion corrector $G$ is shared across
         all resolution stages; a shared FNO forecaster $F_r$ is trained per stage.}
\label{tab:hyperparams_arch}
\small
\begin{tabular}{l l l}
\toprule
\textbf{Parameter} & \textbf{1D Burgers} & \textbf{2D Kraichnan} \\
\midrule
\multicolumn{3}{l}{\textit{FNO Forecaster}} \\[2pt]
Fourier layers                & 3                              & 4 \\
Channel width                 & 32                             & 32 \\
Mode truncation $k_\text{max}$& $N_r / 4$                      & $N_r / 4$ per dim \\
Activation                    & GELU                           & GELU \\
\midrule
\multicolumn{3}{l}{\textit{Diffusion Corrector U-Net (shared across all stages)}} \\[2pt]
Input channels                & \multicolumn{2}{l}{3\quad $[\mathbf{x}_\text{noisy},\, \mathbf{u}_\text{forecast},\, \mathbf{u}_\text{coarse}^\uparrow]$} \\
Base channels                 & 64                             & 48 \\
Channel multipliers           & $[1, 2, 4]$                    & $[1, 2, 4]$ \\
Residual blocks per level     & 1                              & 1 \\
Self-attention                & None                           & Bottleneck only ($32{\times}32$) \\
Normalization                 & GroupNorm ($G{=}8$)            & GroupNorm ($G{=}8$) \\
Padding mode                  & Zero                           & Circular \\
Downsampling                  & Conv stride-2                  & Conv stride-2 \\
Upsampling                    & Nearest $+$ Conv               & Nearest $+$ Conv \\
\midrule
\multicolumn{3}{l}{\textit{FiLM Conditioning (injected at every residual block)}} \\[2pt]
Conditioning signals          & \multicolumn{2}{l}{Diffusion step $\tau$ and resolution index $r \in \{0,1,2\}$} \\
Embedding                     & \multicolumn{2}{l}{Sinusoidal $\to$ MLP $\to$ 128-d; summed before injection} \\
Physical time context         & \multicolumn{2}{l}{None (OU forcing is stationary)} \\
\midrule
\multicolumn{3}{l}{\textit{Noise Schedule}} \\[2pt]
Schedule                      & Cosine                         & Cosine \\
Diffusion steps $T$           & 1000                           & 1000 \\
Inference                     & DDIM, 25 steps, $\eta{=}0$    & DDPM, 100 steps, $\eta{=}1$ \\
\bottomrule
\end{tabular}
\end{table}

\begin{table}[htbp]
\centering
\caption{Training hyperparameters. All experiments use a single NVIDIA RTX 4090 (24\,GB).}
\label{tab:hyperparams_training}
\small
\begin{tabular}{l l l}
\toprule
\textbf{Parameter} & \textbf{1D Burgers} & \textbf{2D Kraichnan} \\
\midrule
\multicolumn{3}{l}{\textit{FNO Forecaster Training}} \\[2pt]
Epochs                        & 100                            & 80 \\
Batch size                    & 64                             & 32 \\
Optimizer                     & AdamW                          & AdamW \\
Learning rate                 & $10^{-3}$                      & $10^{-3}$ \\
Weight decay                  & $10^{-4}$                      & $10^{-4}$ \\
LR scheduler                  & Cosine annealing               & Cosine annealing \\
Loss                          & MSE                            & MSE \\
\midrule
\multicolumn{3}{l}{\textit{Diffusion Corrector Training}} \\[2pt]
Training steps                & 150{,}000                      & 400{,}000 \\
Batch size                    & 16                             & 16 \\
Optimizer                     & AdamW                          & AdamW \\
Learning rate                 & $2 \times 10^{-4}$             & $2 \times 10^{-4}$ \\
Weight decay                  & $10^{-5}$                      & $10^{-5}$ \\
LR warmup steps               & 2{,}000                        & 5{,}000 \\
LR schedule (post-warmup)     & Cosine decay                   & Cosine decay \\
EMA decay                     & 0.9999                         & 0.9999 \\
\midrule
\multicolumn{3}{l}{\textit{Input Noise Augmentation — per-stage $[\sigma_\text{fc},\, \sigma_\text{co}]$}} \\[2pt]
Stage 0 (obs $\to$ level 1)   & $[0.027,\; 0.000]$             & $[0.050,\; 0.020]$ \\
Stage 1                       & $[0.027,\; 0.003]$             & $[0.100,\; 0.200]$ \\
Stage 2 ($\to$ target)        & $[0.027,\; 0.004]$             & $[0.100,\; 0.190]$ \\
Augmentation probability      & 1.0                            & 1.0 \\
\midrule
\multicolumn{3}{l}{\textit{General}} \\[2pt]
Random seed                   & \multicolumn{2}{l}{42} \\
Hardware                      & \multicolumn{2}{l}{NVIDIA RTX 4090 (24\,GB)} \\
\bottomrule
\end{tabular}
\end{table}

\begin{table}[htbp]
\centering
\caption{Learnable parameter counts for the diffusion corrector and competing baselines.
         For the Iterative Refinement (IR) method only the shared U-Net corrector $G$
         is listed; FNO forecasters are additional but are also required by ablation
         variants and are not part of the generative corrector comparison.
         Spectral Upsample has no learnable parameters (analytical zero-padding).}
\label{tab:params}
\small
\begin{tabular}{l r r}
\toprule
\textbf{Model / Component} & \textbf{1D Burgers} & \textbf{2D Kraichnan} \\
\midrule
Spectral Upsample           & 0                   & 0 \\
EDSR~\cite{lim2017edsr}     & 445{,}121           & 1{,}662{,}977 \\
One-Shot Diffusion SR U-Net & 2{,}280{,}641       & 6{,}970{,}689 \\
\midrule
IR Diffusion Corrector $G$ (ours) & \textbf{2{,}412{,}353} & \textbf{4{,}171{,}537} \\
\bottomrule
\end{tabular}
\end{table}





\end{document}